\documentclass[sn-mathphys-num]{sn-jnl}

\usepackage{graphicx}
\usepackage{multirow}
\usepackage{amsmath,amssymb,amsfonts}
\usepackage{amsthm}
\usepackage{mathrsfs}
\usepackage[title]{appendix}
\usepackage{xcolor}
\usepackage{textcomp}
\usepackage{booktabs}
\usepackage{algorithm}
\usepackage{algpseudocode}
\usepackage{subcaption}

\usepackage{longtable}
\usepackage{caption}
\usepackage{tabularx}
\usepackage{makecell}
\theoremstyle{thmstyleone}

\theoremstyle{thmstyletwo}

\theoremstyle{thmstylethree}

\usepackage{tikz}
\usetikzlibrary{shapes.geometric, positioning, arrows.meta}


\begin{document}

  \title{When Does Oversampling Help? A Decision Framework for Imbalanced Data Streams Under Label Delay and Concept Drift}

\author*[1,2]{\fnm{Ahmed} \sur{Awad}}\email{ahmed.awad@buid.ac.ae}

\author[1]{\fnm{Anirudh} \sur{Sengar}}\email{23001546@ug.buid.ac.ae}

\affil*[1]{\orgdiv{Faculty of Engineering \& IT}, \orgname{The British University in Dubai}, \orgaddress{\city{Dubai}, \country{UAE}}}

\affil[2]{\orgdiv{Faculty of Computers and Artificial Intelligence}, \orgname{Cairo University}, \orgaddress{\city{Giza}, \country{Egypt}}}

\abstract{
For imbalanced data streams with delayed labels, is it worth adding a streaming oversampler, or does continuous re-evaluation of earlier predictions already recover the cost of the delay? We address this through an extensive experimental comparison of streaming oversampling and continuous re-evaluation under joint class imbalance, concept drift, and label delay. Streaming oversamplers are typically evaluated under the test-then-train assumption of immediate labels; their behaviour under these conditions, assessed through re-evaluation, has not been systematically studied. We close this gap with a decision framework that predicts when oversampling adds value over re-evaluation, organised around three axes: drift density, minority-class learnability, and delay relative to the refinement window, and operationalised as a competition between a recovery channel (re-evaluation) and a minority-coverage channel (oversampling). Evaluating $15$ classifiers on $6$ synthetic and $6$ real streams, we find: re-evaluation recovers delay-induced loss on drift-active streams but is silent on stationary ones; oversampling's marginal contribution over re-evaluation is statistically detectable in only one real-data configuration; and cost--performance Pareto recommendations from synthetic streams do not transfer to real data, where fronts partition by drift regime. Method selection should therefore be conditioned on the stream's profile and re-estimated as it evolves, not inherited from synthetic benchmarks.
}


\maketitle

\section{Introduction}
\label{sec:intro}

Streaming classification operates under three simultaneous pressures: instances arrive continuously, the class distribution is often heavily skewed toward a non-informative majority, and the underlying concept drifts over time~\cite{gama2014survey,gomes2017adaptive,wang2015imbalanced}. Streaming oversamplers (C-SMOTE~\cite{csmote2018}, VFC-SMOTE~\cite{vfcsmote2020}, OOB/UOB~\cite{wang2015imbalanced}, RebalanceStream~\cite{rebalancestream2016}) address the imbalance dimension and are evaluated under the test-then-train protocol~\cite{gama2013evaluating}, which assumes labels arrive immediately after prediction. This assumption is rarely satisfied in deployment: fraud labels emerge with chargeback latency~\cite{dal2014learned}, clinical labels with follow-up latency, and conversion labels with user-engagement latency~\cite{li2010contextual}. Grzenda et al.~\cite{DBLP:journals/datamine/GrzendaGB20} proposed \emph{continuous re-evaluation} as the appropriate protocol for delayed labels: while an instance waits for its label, later-arriving labels train the model, and a refined prediction is recomputed at re-evaluation time. Whether oversampling retains its benefit once labels are delayed under imbalance and drift, or whether continuous re-evaluation alone recovers the loss, has not been established.

Two strands of literature have not been crossed. The continuous re-evaluation study~\cite{DBLP:journals/datamine/GrzendaGB20} evaluated bare streaming classifiers on balanced and mildly imbalanced streams; it did not examine oversamplers or severely imbalanced streams. The streaming oversampling literature reports performance under the synchronous-label assumption and does not characterise behaviour under delay. The interaction is non-trivial: synthetic minority instances generated under a stale concept may reinforce an outdated boundary, harming performance precisely where imbalance most demands it.

We close this gap with a systematic experimental comparison. We evaluate $15$ streaming classifiers ($6$ bare base learners and $9$ oversampler variants) on $6$ controlled synthetic streams with abrupt and gradual drifts, and on $6$ real datasets spanning stationary, dense-drift, and high-impact-drift profiles (Electricity~\cite{harries1999splice}, KDDCup~\cite{kddcup1999}, PAKDD~\cite{pakdd2009}, Forest Covertype~\cite{blackard1999covertype}, INSECTS-Abrupt-imbalanced~\cite{souza2020insects}, Credit Card Fraud~\cite{dalpozzolo2015creditcard}). Following \cite{DBLP:journals/datamine/GrzendaGB20}, we report both the arrival and refined predictions obtained $K = 100$ labels later. On synthetic streams we parametrise delay through the delay-to-drift ratio $\rho = \Delta / W$ at three regimes ($\rho \in \{0.25, 1.0, 2.0\}$); on real streams, where drift structure is unknown, we use percentage-of-stream delays ($\delta \in \{0.1\%, 0.5\%, 1.0\%\}$).

The contributions in this paper are:
\begin{itemize}
    \item A \emph{systematic experimental comparison} of $15$ streaming classifiers under delayed feedback, on $6$ controlled synthetic streams and $6$ real datasets with heterogeneous drift profiles, evaluated through continuous re-evaluation with a $K=100$ refinement window. To our knowledge, this is the first such evaluation of streaming oversamplers across both controlled and uncontrolled drift settings.
    \item A \emph{transferability test} of cost--performance Pareto recommendations from synthetic to real streams: under heterogeneous real-data drift, recommendations derived from controlled synthetic streams do not transfer, and per-dataset Pareto fronts partition by drift regime instead.
    \item A \emph{three-axis decision framework} that unifies these results, drift density, minority-class learnability, and delay relative to the refinement window, selecting between oversampling and re-evaluation via a two-channel competition between the bare learner's per-label \emph{recovery channel} and the oversampler's \emph{minority-coverage channel}.
    \item \emph{Deployment guidance} that conditions oversampler choice on the stream's profile rather than on synthetic Pareto status.
\end{itemize}

The remainder of the paper is organised as follows. Section~\ref{sec:related} reviews streaming imbalanced learning, concept drift, and evaluation under asynchronous labels. Section~\ref{sec:problem} formalises the label-delay setting and introduces the $\rho$ formalism. Section~\ref{sec:experimental} describes the experimental setup. Section~\ref{sec:results} presents results. Section~\ref{sec:discussion} interprets the results and presents the decision framework and deployment guidance. Section~\ref{sec:conclusion} concludes with limitations and future work.

\section{Related Work}
\label{sec:related}

\subsection{Streaming Learning with Class Imbalance}
Class imbalance in data streams has been extensively studied \cite{aguiar2024survey,gomes2017adaptive,wang2015imbalanced}. Early approaches adapted batch-mode techniques such as cost-sensitive learning \cite{elkan2001foundations} and ensemble methods \cite{wang2013mining} to the streaming setting. The intrinsic difficulty of an imbalanced problem is understood to be a joint function of imbalance ratio and feature-conditional separability rather than ratio alone \cite{lopez2013insight,aguiar2024survey}, motivating evaluation across multiple regimes rather than a single imbalance level.

SMOTE-based oversampling methods emerged as a promising direction. VFC-SMOTE \cite{vfcsmote2020} maintains a sliding window of minority instances and generates synthetic examples via linear interpolation when the minority class is underrepresented. C-SMOTE \cite{csmote2018} extends this with chunk-based processing and adaptive synthetic generation rates. Online Bagging variants (OOB, UOB) \cite{wang2015imbalanced} use Poisson sampling to implicitly oversample minority instances without explicit synthetic generation. RebalanceStream \cite{rebalancestream2016} combines ensemble learning with online SMOTE.

All existing evaluations of these methods assume synchronous labels, an assumption we systematically relax in this work.

\subsection{Concept Drift}
Concept drift refers to changes in the underlying data distribution $P(\mathcal{X}, \mathcal{Y})$ over time \cite{gama2014survey,lu2018learning}. Drift can manifest as changes in prior probability $P(\mathcal{Y})$, class-conditional density $P(\mathcal{X}|\mathcal{Y})$, or posterior probability $P(\mathcal{Y}|\mathcal{X})$ \cite{webb2016characterizing}. Drift may be sudden (instantaneous), gradual (linear transition over a window $W$), incremental (slow continuous change), or recurring (oscillating between concepts) \cite{zliobaite2010learning}.

Adaptive learners for drifting streams employ mechanisms such as ensemble diversity maintenance \cite{bifet2009adaptive,gomes2017adaptive}, explicit drift detection \cite{gama2004learning,DBLP:conf/sdm/BifetG07}, online bagging \cite{oza2005online}, and forgetting mechanisms \cite{klinkenberg2004learning}. However, these methods do not account for label delay---they assume labels are available synchronously for drift detection and model updates.

\subsection{Evaluation Under Asynchronous Labels}
Streaming-classifier evaluation is traditionally framed as prequential (test-then-train) \cite{gama2013evaluating}: each instance is first predicted by the current model, then its label, assumed immediately available, is used to update the model and score the prediction. The framework rests on the assumption that labels arrive synchronously with features, an assumption rarely satisfied in production systems, where the time between feature arrival and label release can be substantial.

A small body of supervised work relaxes this assumption. \cite{mesterharm2005line} studies label delay in the mistake-bound model and \cite{joulani2013online} analyses delayed feedback in online convex optimisation; both treat delay as an analytical constant without examining its interaction with non-stationary streams. \cite{zliobaite2015evaluation} discusses evaluation challenges under delayed labels but focuses on metric computation rather than learning algorithm design.

The closest prior work is \cite{DBLP:journals/datamine/GrzendaGB20}, which introduces continuous re-evaluation under delay: while an instance waits for its label, later-arriving labels train the model, and a refined prediction can be produced from the updated model state. Their analysis establishes re-evaluation as a learner-agnostic recovery mechanism on otherwise stationary streams. The interaction of re-evaluation with concept drift and class imbalance, particularly how the delay-to-drift ratio governs the trade-off and how oversampling either amplifies or interferes with the recovery channel, remains open and motivates this work.

\section{Problem Formulation}
\label{sec:problem}

We consider a potentially infinite data stream $\mathcal{S} = \langle(\mathbf{x}_1, y_1), (\mathbf{x}_2, y_2), \ldots\rangle$ where $\mathbf{x}_i \in \mathcal{X}$ is a feature vector and $y_i \in \{0, 1\}$ is a binary class label. The stream is generated by a time-varying distribution $P_t(\mathcal{X}, \mathcal{Y})$. The subscript $i$ identifies a stream instance, whose components are the feature vector $\mathbf{x}_i$ and the true label $y_i$; thus, we can correlate label $y_i$ to the feature vector $\mathbf{x}_i$ of instance $i$. To track timing, we define $t_{\text{arrival}}(i)$ and $t_{\text{label}}(i)$, both in $\mathbb{N}$, as the monotonic arrival times of the feature vector and of the true label of stream instance $i$, respectively.


The problem addressed in this paper considers three practical challenges in machine learning on data streams: 1) Imbalanced data: a problem inherited from traditional machine learning that is amplified in a streaming setup, 2) Concept drift: An original challenge for streaming machine learning that is well-defined with well-established techniques for detection and adaptation of the streaming learners, 3) The label delay: the lag (latency) between the arrival of the features and the true label, i.e., $(t_{\text{label}}(i) - t_{\text{arrival}}(i) = \Delta) > 0$. We briefly describe each challenge and then argue that their interplay reveals gaps in the literature and practice.

\subsection{Imbalanced Data Streams}

Data imbalance is well-studied in traditional machine learning. In binary classification, the ratio of labels $0$ to $1$ deviates substantially from $1:1$. Imbalance ratio alone, however, does not fully characterise how hard the minority class is to learn; feature-conditional separability, the extent to which minority instances occupy distinct regions of feature space rather than overlap with the majority, jointly determines minority-class learnability~\cite{lopez2013insight, aguiar2024survey}. A moderately imbalanced stream with poor minority separability can be harder to learn than a severely imbalanced one with cleanly separated classes. In offline learning, the ratio can be learned from the training dataset, and an oversampling technique (e.g., \cite{chawla2002smote}) can be applied. On the contrary, in a streaming fashion, this ratio can vary over time. Therefore, it has to be continuously monitored, and synthetic samples have to be injected into the stream to help the streaming learner better identify the rare label(class). Unlike traditional machine learning, oversampling incurs continuous overhead that can increase prediction latency and increase memory consumption in a streaming machine learning pipeline.

\subsection{Data Stream with Concept Drift}

The process of classification can be described by the prior probabilities of the targets $P(\mathcal{Y})$, and the target conditional probability density function $P(\mathcal{X}|\mathcal{Y})$. The classification decision is made based on the posterior probabilities of the targets, which can be obtained from:
     \begin{equation}\label{eq:classif}
         P(\mathcal{Y}|\mathcal{X}) = \dfrac{P(\mathcal{Y}) \cdot P(\mathcal{X}|\mathcal{Y})}{P(\mathcal{X})}
     \end{equation}

     Since $P(\mathcal{Y})$ and $P(\mathcal{X}|\mathcal{Y})$ uniquely determines the joint distribution $P(\mathcal{X}, \mathcal{Y})$, concepts can be defined as the joint distribution $P(\mathcal{X}, \mathcal{Y})$~\cite{gama2014survey}. In practice, concepts can change over time, resulting in concept drift. Based on the definition of concept, concept drift between data at point of time $t_k$ and data at point of time $t_j$ can formally be defined as a difference in the distributions of the data in these time points:
     \begin{equation}
         P_{t_k}(\mathcal{X}, \mathcal{Y}) \neq P_{t_j}(\mathcal{X}, \mathcal{Y})
     \end{equation}

    However, it can be observed from (\ref{eq:classif}) that a concept drift can be caused by either a change in the distribution of the covariates $P(\mathcal{X})$ (and thus in the target conditional probability distribution $P(\mathcal{X}| \mathcal{Y})$), a change in the prior probability distribution $P(\mathcal{Y})$ or change in the posterior probability distribution $P(\mathcal{Y}|\mathcal{X})$.

    Changes in the covariate distribution $P(\mathcal{X})$ that are not accompanied by changes in the posterior probability distribution $P(\mathcal{Y}|\mathcal{X})$ are referred to as \textbf{\textit{``virtual concept drift"}}~\cite{gama2014survey}, $P_{t_k}(\mathcal{X}) \neq P_{t_j}(\mathcal{X}) \wedge
    P_{t_k}(\mathcal{Y}|\mathcal{X}) = P_{t_j}(\mathcal{Y}|\mathcal{X})$. These changes do not affect the decision boundary of the model. Thus, handling of this type of drift may not be required. It may still prove useful in practice to detect this drift for exploratory analysis and to better understand how the data change over time.

    Changes in the posterior probability distribution $P(\mathcal{Y}|\mathcal{X})$ are referred to as \textbf{\textit{``real concept drift"}}~\cite{gama2014survey}. These changes affect the model's decision boundary, and when they occur, the model's predictions may be compromised. Thus, it is very important to detect these types of drift and handle them by retraining the model.
 
\subsection{Label Delay}

In the streaming machine learning literature, test-then-train, a.k.a. prequential evaluation, is commonly accepted. However, it relies on the strong assumption of synchronous label arrival, i.e., $\Delta = 0$ for all instances. We relax the synchronous assumption. At time $t$, the learner receives instance $\mathbf{x}_i$ and makes prediction $\hat{y}_i^{(t)}$, but the true label $y_i$ does not become available until time $t_{\text{arrival}}(i) + \Delta$, where $\Delta \geq 0$ is the \emph{label delay} measured in stream timesteps.

The learner maintains a \emph{pending queue} of instances awaiting label release:
\[
Q_t = \{(\mathbf{x}_i, \hat{y}_i^{(t_{\text{arrival}}(i))}, t_{\text{arrival}}(i), t_{\text{label}}(i)) \mid t_{\text{arrival}}(i) \leq t, \; t_{\text{label}}(i) > t\}
\]

At each timestep $t$, all instances with $t_{\text{label}}(i) \leq t$ are dequeued, their true labels $y_i$ are observed, and the learner trains on them. Moreover, under label delay the prequential evaluation regime scores instance $i$ using $\hat{y}_i^{(t_{\text{arrival}}(i))}$, the prediction made by the model state at the arrival of $\mathbf{x}_i$, before any label that arrives during $i$'s waiting period has been observed. As the streaming instance is queued until its true label arrival, reevaluation may help reach the correct prediction. Reevaluation is based on the fact that between $t_{\text{arrival}}(i)$ and $t_{\text{label}}(i)$ of $\mathbf{x}_i$, several labels $\{y_j \mid \mathbf{x}_j \in Q_t \wedge j \neq i\}$ might have arrived and trained the model. Let $f^{(t)}$ denote the model state at time $t$. Using the model state at the arrival of such a label $y_j$, we can produce a new prediction $\hat{y}_i^{(t)} = f^{(t)}(\mathbf{x}_i)$ for the waiting instance $\mathbf{x}_i$. That is, we can get a new prediction for the same feature vector $\mathbf{x}_i$ of instance $i$ at a later time. The set of valid reevaluation predictions for instance $i$ is
\[
\{\hat{y}_i^{(t)} = f^{(t)}(\mathbf{x}_i) \mid t_{\text{arrival}}(i) < t \leq t_{\text{label}}(i), \; f^{(t)} \neq f^{(t_{\text{arrival}}(i))}\},
\]
where the condition $f^{(t)} \neq f^{(t_{\text{arrival}}(i))}$ requires that the model has been updated since $i$'s arrival, since reevaluating against an unchanged model leaves the prediction unchanged. We refer to $\hat{y}_i^{(t_{\text{arrival}}(i))}$ as the \emph{arrival} prediction and to a prediction taken at a later valid reevaluation time as a \emph{refined} prediction. This reevaluation technique is proposed in~\cite{DBLP:journals/datamine/GrzendaGB20}.

\subsection{Challenges' Interaction}

Handling concept drift and handling stream imbalance have been addressed in the literature individually~\cite{gama2014survey},~\cite{csmote2018} and in combination~\cite{vfcsmote2020}. Similarly, re-evaluation under label delay and concept drift has been studied in~\cite{DBLP:journals/datamine/GrzendaGB20}. One can say that concept drift is common to all streaming learning scenarios; it is the shared axis along which streaming learners are already designed to adapt. What has not been studied is the pair of challenges layered on top of it: class imbalance and label delay, considered jointly.

The effect of the interaction of the three challenges has not been studied. As will be shown later, such interaction is non-trivial. The reason is that the two remedies work against each other in the context of delay. An oversampler injects synthetic minority instances based on the minority distribution it has observed so far; under label delay, that observed distribution is stale, because the labels that would reveal a drift have not yet been released. Synthetic instances generated from a stale minority concept can therefore reinforce an outdated decision boundary, so oversampling under delay is not merely less helpful; it can be actively harmful, and precisely in the high-imbalance regime where it is most needed. Whether this happens and how bad should depend on how the delay compares with the timescale of the drift: a delay much shorter than the drift window leaves the oversampler working with a still-valid concept, whereas a delay that exceeds it does not. This is what motivates studying the effect as a function of the delay-to-drift ratio rather than the raw delay.

To reason about which remedy is responsible for which gain, we name two analytical channels through which arrival-time performance can be improved under label delay. The \emph{recovery channel} refers to re-evaluation: available to every streaming learner, whether or not it oversamples, it refines the arrival-time prediction, for instance $i$ using model states updated by labels released during $i$'s waiting period; the number of such refinements scales with the delay. The \emph{minority-coverage channel} refers to oversampling: it injects synthetic minority instances that the bare learner would not otherwise see, raising the minority signal in the training stream. The two channels are not mutually exclusive; an oversampling pipeline uses both, but they are conceptually separate, and any observed performance gain can in principle come from either or both.

These considerations raise three questions. First, if we assume an oversampling technique is used, we do not know how label delay affects overall pipeline performance under prequential evaluation. Second, if both oversampling and re-evaluation are employed and we observe performance improvement, is it the effect of the oversampler only, the re-evaluation only, or both? This question is sharpened by the fact that re-evaluation is available to every streaming learner, with or without oversampling, so any oversampling-specific benefit must be isolated from a benefit that the bare learner already obtains. Finally, is the potential improvement in model performance significant to justify the oversampling and/or re-evaluation overhead? In a stream, oversampling is not a one-time preprocessing cost but a continuous one, as noted above; a method that improves predictive performance while replaying many times more instances than another is not obviously preferable, and the trade-off has to be made explicit rather than assumed.

In this work, we answer these questions by conducting benchmarking experiments on a large set of streaming classifiers trained on both synthetic and real data streams. The synthetic data streams exhibit concept drift and varying degrees of class imbalance. In real data streams, concept drift may occur, but the streams also exhibit class imbalance. For both stream types, we simulate label delays. For synthetic streams, we define delay magnitudes, $\rho$, as a function of the known drift window length; this allows us to test the interaction mechanistically, since the drift timescale is controlled. For real streams, where the drift structure is unknown, and $\rho$ cannot be computed, we define the delay as a function of the overall stream length to test whether the findings generalise beyond the controlled setting. Exact formulas and values are discussed in more detail in Section~\ref{sec:experimental}, where we provide our answers for the following research questions:

\textbf{RQ1.} Does the delay-to-drift ratio $\rho$ govern how label delay affects the performance of streaming oversampling methods under concept drift?

\textbf{RQ2.} Does continuous re-evaluation recover the performance that label delay costs streaming oversamplers?

\textbf{RQ3.} When re-evaluation improves performance, is the gain attributable to re-evaluation alone or to the combination of oversampling and re-evaluation?

\textbf{RQ4.} Which streaming oversampling methods deliver performance gains that justify their processing overhead?

\section{Experimental Design}
\label{sec:experimental}

\subsection{Datasets}

\paragraph{Synthetic Data Streams} To study the interaction between class imbalance, concept drift, and label delay under controlled conditions, we use synthetic streams in which the drift mechanism, drift timing, and class imbalance are all known by construction. We restrict attention to real concept drift ($P(\mathcal{Y}|\mathcal{X})$), the only form of drift that alters the decision boundary and the form that the evaluated oversampling methods are designed to address. Drift in the covariate distribution $P(\mathcal{X})$ or the class prior $P(\mathcal{Y})$ alone does not require model adaptation and is therefore outside the scope of this study.

\textbf{Stream generators.} We use two stream generators from the MOA framework~\cite{bifet2010moa}, both producing two-class problems with a single drift event placed at the midpoint of the stream:
\begin{itemize}
    \item $SEA$ (abrupt drift): the underlying concept switches instantaneously at the drift point ($W = 1$).
    \item $SEA_G$ (gradual drift): the underlying concept transitions linearly over a window of $W = 10{,}000$ instances centred on the drift point.
\end{itemize}
The two generators differ only in the temporal profile of the drift; the pre-drift and post-drift concepts are identical between them. This pairing allows us to attribute any differences between the two settings specifically to the drift timescale, holding all other factors constant.

\textbf{Imbalance ratios.} For each generator, we configure three class-imbalance ratios (minority: majority): $1\!:\!9$, $2\!:\!8$, and $3\!:\!7$. The imbalance ratio is held constant throughout the stream and is preserved across the drift event. The most-skewed setting ($1\!:\!9$) is comparable to the imbalance level routinely reported in streaming oversampling evaluations~\cite{vfcsmote2020}; the milder settings ($2\!:\!8$, $3\!:\!7$) test whether observed effects persist as the imbalance pressure weakens.

\textbf{Scope.} The combination of two generators and three imbalance ratios yields six synthetic streams, denoted by their generator and minority:majority ratio: \texttt{sea\_1,9}, \texttt{sea\_2,8}, \texttt{sea\_3,7}, \texttt{seaG\_1,9}, \texttt{seaG\_2,8}, and \texttt{seaG\_3,7}. Each stream contains $100{,}000$ instances, with the drift event located at instance $50{,}000$. Across the six streams, the drift type ($P(\mathcal{Y}|\mathcal{X})$), drift location, and stream length are fixed; only the drift timescale (abrupt vs.\ gradual) and the imbalance ratio vary.


 \paragraph{Real-World Datasets} 

To test whether the synthetic findings generalise to data of practical interest,
we evaluate on six publicly available real-world streams: Electricity~\cite{harries1999splice}
($45{,}312$ instances, $8$ features, $\approx 42\%$ minority),
KDDCup99~\cite{kddcup1999} ($494{,}021$; $41$; $\approx 20\%$),
PAKDD~\cite{pakdd2009} ($50{,}000$; $30$; $\approx 20\%$),
Forest Covertype~\cite{blackard1999covertype} ($581{,}012$; $54$; $\approx 3\%$),
INSECTS-Abrupt-imbalanced~\cite{souza2020insects} ($355{,}275$; $33$; $\approx 28\%$),
and Credit Card Fraud~\cite{dalpozzolo2015creditcard} ($284{,}807$; $30$; $\approx 0.17\%$),
where the reported sizes and class distributions are those after binarisation.\footnote{%
Covertype and INSECTS are multi-class in raw form and were binarised (positive = minority);
full binarisation details are in the repository. Throughout, class~$1$ is the positive
(minority) class and class~$0$ the negative (majority) class.}
The first three match the evaluation set used by Bernardo and Della Valle in the original
VFC-SMOTE paper~\cite{vfcsmote2020}, enabling direct comparison with their reported results
under synchronous labels. The remaining three extend the evaluation toward stronger imbalance
(Covertype $\approx 3\%$, Credit Card Fraud $\approx 0.17\%$ minority) and include a stream
(INSECTS) explicitly designed as a class-imbalanced streaming benchmark~\cite{souza2020insects}.
We do not formally certify the presence of concept drift in these streams; rather, we follow
the convention of the streaming-learning literature, in which these datasets are routinely
treated as drift-bearing on the basis of their temporal ordering and the documented
non-stationarity of their underlying phenomena (electricity demand, network traffic, credit
applications, fraud patterns, insect activity, forest composition over time). This study
assumes drift in line with that convention rather than claiming it.

\subsection{Label Delay Parametrisation}
\label{sec:exp:delay}
Label delay enters through the pending-queue mechanism of Section~\ref{sec:problem}: a vector arriving at time $t$ has its label released at $t + \Delta$, and labels arriving in between train the model. The delay $\Delta$ is the design parameter under study; it is set differently on the synthetic and real tracks because each exposes different information about the drift.

\paragraph{Synthetic streams: the delay-to-drift ratio $\rho$} Here the drift window $W$ is known by construction, so we parametrise delay relative to it as $\rho = \Delta/W$ rather than in absolute terms, and evaluate three regimes: \emph{safe} ($\rho = 0.25$), where the label is released before the post-drift concept replaces the one that generated the prediction; \emph{boundary} ($\rho = 1.0$), where labels reflect a mixture of pre- and post-drift concepts; and \emph{unsafe} ($\rho = 2.0$), where the generating concept has been entirely supplanted before its label arrives. For SEAg streams ($W = 10{,}000$) these correspond to $\Delta \in \{2{,}500,\,10{,}000,\,20{,}000\}$. SEA streams drift instantaneously ($W = 1$), so any positive $\Delta$ gives $\rho > 1$; we treat the same absolute delays as unsafe by construction.

\paragraph{Real datasets: the delay percentage $\delta\%$} Here $W$ is unknown and likely heterogeneous, so the $\rho$ formalism does not apply. We instead set delay as a constant fraction of stream length, $\Delta = \mathrm{round}(\delta\%/100 \cdot N)$ for stream length $N$, with $\delta\% \in \{0.1\%,\,0.5\%,\,1.0\%\}$ applied identically across the six datasets. A percentage keeps the delay commensurate with each stream's own timescale: $1{,}000$ instances is large on Electricity ($\approx 2.2\%$) but small on Forest Covertype ($\approx 0.17\%$), so no single absolute $\Delta$ would treat both fairly.

\paragraph{The no-delay reference} A fourth setting, $\Delta = 0$, is used only in the prequential (no re-evaluation) experiments, providing a synchronous-label reference against which the cost of delay is measured. Re-evaluation excludes it: with $\Delta = 0$ the queue never accumulates and no refinement is possible. The prequential track thus runs at $\delta\% \in \{0\%,\,0.1\%,\,0.5\%,\,1.0\%\}$ (real) and $\rho \in \{0,\,0.25,\,1.0,\,2.0\}$ (synthetic), the re-evaluation track at the three non-zero values in each case.

\subsection{Streaming Learners}
\label{sec:exp:learners}

We evaluate $15$ streaming classifiers, comprising six bare base learners and nine oversampling variants. The pool is constructed as the union of three references. First, the bare base learners include the methods evaluated by Grzenda et al.~\cite{DBLP:journals/datamine/GrzendaGB20} in their continuous re-evaluation study, ensuring that the re-evaluation behaviour we observe is directly comparable with the protocol's original validation. Second, the oversampler set covers the state-of-the-art streaming oversampling methods evaluated in the original VFC-SMOTE paper~\cite{vfcsmote2020}, ensuring that any conclusions about oversampling under label delay are drawn against the literature's standard comparison set. Third, we include all five base-learner configurations of VFC-SMOTE used in~\cite{vfcsmote2020}, so the effect of the base-learner choice can be analysed in isolation from the synthetic-instance generation mechanism.

\paragraph{Bare base learners (6)} Adaptive Random Forest (\textbf{ARF})~\cite{gomes2017adaptive}, Hoeffding Adaptive Tree (\textbf{HAT})~\cite{bifet2009adaptive}, $k$-Nearest Neighbours (\textbf{KNN}), Naive Bayes (\textbf{NV}), Temporally Augmented Classifier with ARF base (\textbf{SWT}), and Online Bagging without resampling (\textbf{OzaBag})~\cite{oza2005online}. ARF and OzaBag are ensemble methods without any imbalance-aware mechanisms and serve as the reference for the bagging-based oversamplers below. HAT, KNN, and NV are single-model learners drawn from the bare-learner set in~\cite{DBLP:journals/datamine/GrzendaGB20}. SWT augments the feature space with recent class labels to capture temporal context.

\paragraph{Oversampling methods (9)} Online Bagging with oversampling (\textbf{OOB}) and undersampling (\textbf{UOB})~\cite{wang2015imbalanced}, which adjust the Poisson sampling parameter of OzaBag to favour the minority class. Continuous SMOTE (\textbf{C-SMOTE})~\cite{csmote2018}, which applies SMOTE oversampling at chunk boundaries. RebalanceStream (\textbf{RB})~\cite{rebalancestream2016}, which combines ensemble learning with online SMOTE. VFC-SMOTE (\textbf{VFCSMOTE})~\cite{vfcsmote2020}, configured in five base-learner variants (\textbf{VFCSMOTE\_ARF}, \textbf{VFCSMOTE\_HAT}, \textbf{VFCSMOTE\_KNN}, \textbf{VFCSMOTE\_NV}, \\\textbf{VFCSMOTE\_SWT}), which maintains a sliding window of minority instances and generates synthetic instances by SMOTE interpolation when the minority class is underrepresented.

\paragraph{Method--base pairings} For the analysis of oversampling's isolated effect (Section~\ref{sec:exp:stats}, RQ3), we pair each oversampler with the bare learner it most directly extends. The five VFCSMOTE variants are paired with their corresponding bare base learner (VFCSMOTE\_ARF with ARF, and so on). OOB and UOB are paired with OzaBag, the bagging ensemble they extend through Poisson reweighting. C-SMOTE is paired with ARF, its default base learner. RB is paired with SWT, the closest available analogue to its temporal-windowing component. 


\paragraph{Hyperparameters} All learners run with their MOA default configurations. For the VFC-SMOTE family, we follow the hyperparameter values reported in the original paper~\cite{vfcsmote2020}: minimum minority window size $\texttt{minSizeAllowed} = 100$, and synthetic generation threshold $\texttt{percCorrectlyClassified} = 1.0$. No tuning is performed on any dataset, on the grounds that hyperparameter optimisation under label delay is itself an open problem and would confound the effect we seek to isolate.


\subsection{Evaluation Protocol}
\label{sec:exp:protocol}

Each combination of (stream or dataset) $\times$ learner $\times$ delay setting is evaluated using two independent protocols: a \emph{prequential} run and a \emph{re-evaluation} run. The two protocols share the same delay mechanism (the pending queue of Section~\ref{sec:problem}) and the same training schedule (labels train the model when released); they differ only in when predictions are recorded.

\paragraph{Prequential protocol} For every arriving instance $\mathbf{x}_i$ at time $t_{\mathrm{arrival}}(i)$, the model issues a single prediction, $\hat{y}_i^{(t_{\mathrm{arrival}}(i))}$, before the label is released. This prediction is the \emph{arrival} prediction. It corresponds exactly to the test-then-train metric of standard streaming evaluation~\cite{gama2013evaluating}, lifted to the delayed-label setting as formalised in Section~\ref{sec:problem}.

\paragraph{Re-evaluation protocol} In addition to the arrival prediction, the model re-issues a prediction for every \emph{still-pending} instance every $K$ released labels. The re-issued prediction at time $t$ is $\hat{y}_i^{(t)} = f^{(t)}(\mathbf{x}_i)$, where $f^{(t)}$ is the model state at time $t$, updated by all labels released between $t_{\mathrm{arrival}}(i)$ and $t$. The final \emph{refined} prediction for instance $i$ is the prediction issued at the last re-evaluation event before $t_{\mathrm{release}}(i)$. We fix $K = 100$, following~\cite{DBLP:journals/datamine/GrzendaGB20}: every released label triggers a training update, but only every hundredth event triggers a sweep over the pending queue. The sampling is a constant-factor reduction in the number of refined predictions logged; it does not affect the model state or the training schedule.

\paragraph{Refinable subset} Re-evaluation is meaningful only for instances whose pending interval contains at least one model update. The set of such instances is the \emph{refinable subset}. Instances for which the model has not changed between arrival and release contribute no refinement signal and are excluded from re-evaluation analyses; their arrival prediction is the only prediction available, and including them would dilute the refined-vs-arrival comparison with zero-effect cells. For each refinable instance, we retain only the last refined prediction logged before label release, as that is the prediction the learner would commit to at release.

\subsection{Metrics}
\label{sec:exp:metrics}
\paragraph{Predictive performance} The primary metric is the geometric mean of per-class recalls, $G\text{-mean} = \sqrt{R_0 \cdot R_1}$, where $R_0$ and $R_1$ are majority and minority recall. G-mean is standard for imbalanced streaming evaluation~\cite{wang2015imbalanced,vfcsmote2020}: it collapses whenever either class is consistently misclassified, so high accuracy bought by sacrificing minority recall is penalised. We additionally report per-class recall ($R_0$, $R_1$), per-class F1 ($F1_0$, $F1_1$), and accuracy as descriptive metrics that characterise where methods differ in G-mean. All metrics are computed over the full evaluated stream, with the refinable-subset rule (Section~\ref{sec:exp:protocol}) as the only filter. For each cell we compute two values, $G_{\mathrm{arrival}}$ over arrival predictions and $G_{\mathrm{refined}}$ over refined predictions of the same subset; their difference $\Delta_G = G_{\mathrm{refined}} - G_{\mathrm{arrival}}$ is the per-stream observation driving the re-evaluation analysis.

\paragraph{Computational overhead} For each oversampler we track $n_{\mathrm{replayed}}$, the cumulative count of instances replayed to the base learner: synthetic instances generated by SMOTE interpolation for VFC-SMOTE, C-SMOTE, and RebalanceStream, and resampled instances drawn through the modified Poisson scheme for OOB and UOB. Bare base learners and OzaBag perform no minority-targeted resampling, so $n_{\mathrm{replayed}} = 0$. Normalising by stream length $N$ gives the overhead ratio $n_{\mathrm{replayed}}/N$, used to place methods on a common cost axis in Section~\ref{sec:exp:stats}.

\subsection{Research Questions Operationalised}
\label{sec:exp:rqs}
The four research questions of Section~\ref{sec:problem} are operationalised against the preceding protocol as follows.

\textbf{RQ1.} \emph{Does the delay-to-drift ratio $\rho$ govern how label delay affects streaming oversamplers under concept drift?} Addressed against the prequential protocol, applying the delay magnitudes of Section~\ref{sec:exp:delay} to both tracks.

\textbf{RQ2.} \emph{Does continuous re-evaluation recover the performance that label delay costs streaming oversamplers?} Per (learner, delay) cell we compare $G_{\mathrm{refined}}$ against $G_{\mathrm{arrival}}$ across streams; the within-cell observation is $\Delta_G$, and the question is whether $\Delta_G > 0$ holds systematically.

\textbf{RQ3.} \emph{When re-evaluation improves performance, is the gain attributable to re-evaluation alone or to its combination with oversampling?} For each oversampler--base pair (Section~\ref{sec:exp:learners}) we compute the per-stream difference-in-differences $\mathrm{DiD} = \Delta_G^{\mathrm{oversampler}} - \Delta_G^{\mathrm{base}}$. A non-zero DiD indicates re-evaluation benefits the oversampler beyond the bare learner; a near-zero DiD indicates the improvement is attributable to re-evaluation alone. RQ3 is conditional on re-evaluation having improved performance, so cells with no detectable RQ2 improvement are excluded.

This rule applies at different granularities on the two tracks. The six synthetic streams share generator and drift type: if $\Delta_G^{\mathrm{oversampler}}$ is uniformly positive across them, no cell is removed, and we aggregate over $\rho$ to obtain one observation per (oversampler, base) pair at $n=6$. The six real datasets differ substantially in drift profile (Section~\ref{sec:results:real}), so $\Delta_G^{\mathrm{oversampler}}$ need not be uniformly positive; aggregating would conflate genuine recovery on drift-active streams with null recovery on stationary ones. We therefore apply the rule at the finer (dataset, delay) granularity: a cell enters the DiD test only if $\Delta_G^{\mathrm{oversampler}} > 0$. The resulting per-pair $n$ varies, but as long as it stays well above $6$, significance at $\alpha=0.05$ remains reachable.

\textbf{RQ4.} \emph{Which oversamplers deliver performance that justifies their overhead?} The overhead ratio $n_{\mathrm{replayed}}/N$ and the absolute refined G-mean $G_{\mathrm{refined}}$ are aggregated per learner across streams and datasets. A method is \emph{Pareto-optimal} if no other has both a smaller overhead ratio and a larger or equal $G_{\mathrm{refined}}$. Bare learners incur no replay overhead, anchoring the zero-overhead end of the frontier as the baseline an oversampler must surpass in absolute performance to justify its cost.

\subsection{Statistical Methodology}
\label{sec:exp:stats}
\paragraph{Within-cell paired test} Each (learner, delay) cell yields six paired observations: one $\Delta_G$ per stream (synthetic) or dataset (real). We test whether the median $\Delta_G$ differs from zero with the \textbf{Wilcoxon signed-rank test} (two-sided, $\alpha = 0.05$), paired by stream or dataset to control for stream-specific baseline difficulty. With $n = 6$ and all signs in agreement, the smallest attainable two-sided $p$-value is $0.031$; smaller values are arithmetically unreachable regardless of effect magnitude, which is why $n = 6$ is the minimum sample size at which the test can declare significance at conventional thresholds.

\paragraph{Oversampling-isolated effect (DiD)} For RQ3 we apply the same test to the difference-in-differences $\mathrm{DiD} = \Delta_G^{\mathrm{oversampler}} - \Delta_G^{\mathrm{base}}$ across the six streams (or datasets), using the pairings of Section~\ref{sec:exp:learners}. A significant DiD evidences an oversampling effect beyond what re-evaluation provides to the bare learner.

\paragraph{Pareto-optimality} For RQ4 we identify the cost--performance Pareto front from per-learner mean overhead ratio and mean refined-G-mean gain. A learner is on the front if no other has both lower overhead and higher-or-equal gain, with at least one inequality strict. Dominated learners---paying more overhead for no better gain---are reported but flagged as not recommended.


\subsection{Implementation and Reproducibility}
\label{sec:exp:impl}
Experiments are implemented in Java on the MOA framework~\cite{bifet2010moa}, with separate runners for the prequential and re-evaluation protocols sharing identical data-preparation and delay logic. The full evaluation comprises $1{,}260$ runs ($6$ streams/datasets $\times$ $15$ learners $\times$ $4$ delay settings, across both tracks and both protocols). All code, raw and processed results, and analysis scripts are available at the repository.\footnote{\url{https://github.com/AhmedAwad/streaming-oversampling-under-delay-drift/}}


\section{Results}
\label{sec:results}
In this section, we reflect on the four research questions established in Section~\ref{sec:problem} and grounded on the experimental setup discussed in Section~\ref{sec:exp:rqs}. Sections~\ref{sec:results:synthetic} and~\ref{sec:results:real} will discuss the answers we reached to the research questions for the synthetic and real data sets, respectively.

\subsection{Synthetic Datasets}
\label{sec:results:synthetic}

\begin{figure}[t]
  \centering
  \begin{subfigure}[b]{0.32\textwidth}
    \includegraphics[width=\linewidth]{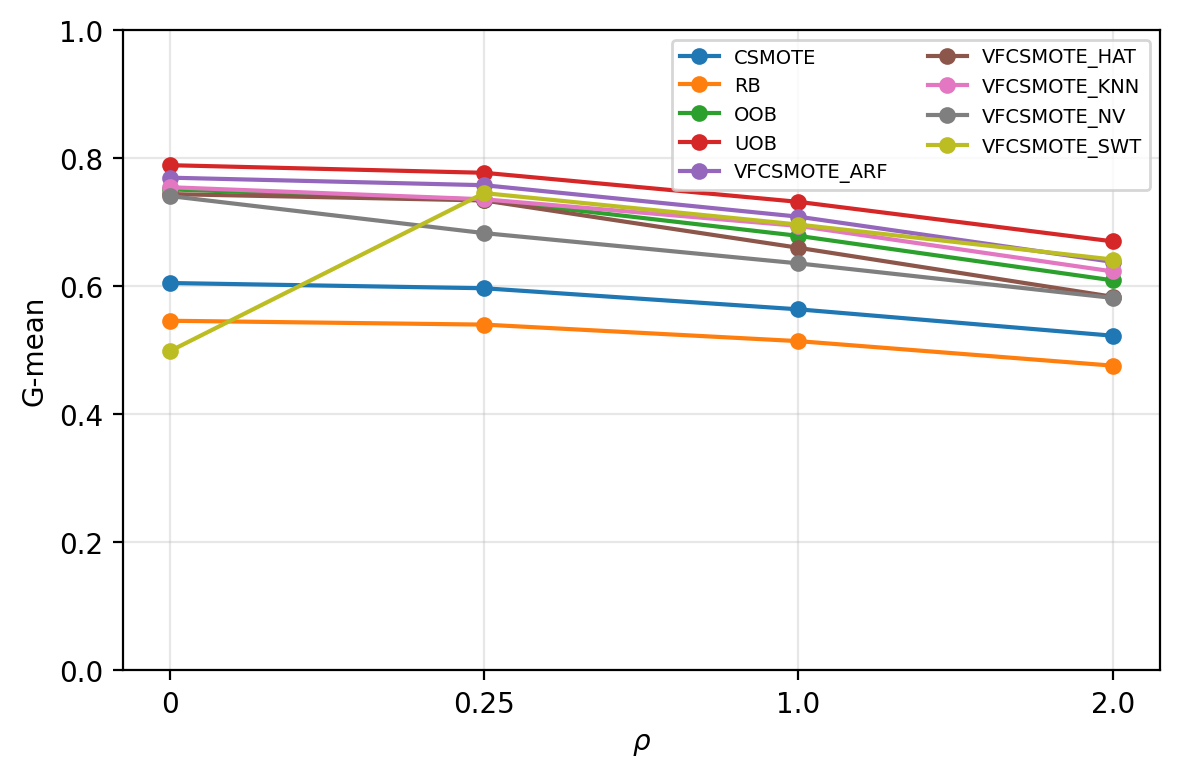}
    \caption{SEA 1:9}
    \label{fig:gmean-sea19}
  \end{subfigure}
  \hfill
  \begin{subfigure}[b]{0.32\textwidth}
    \includegraphics[width=\linewidth]{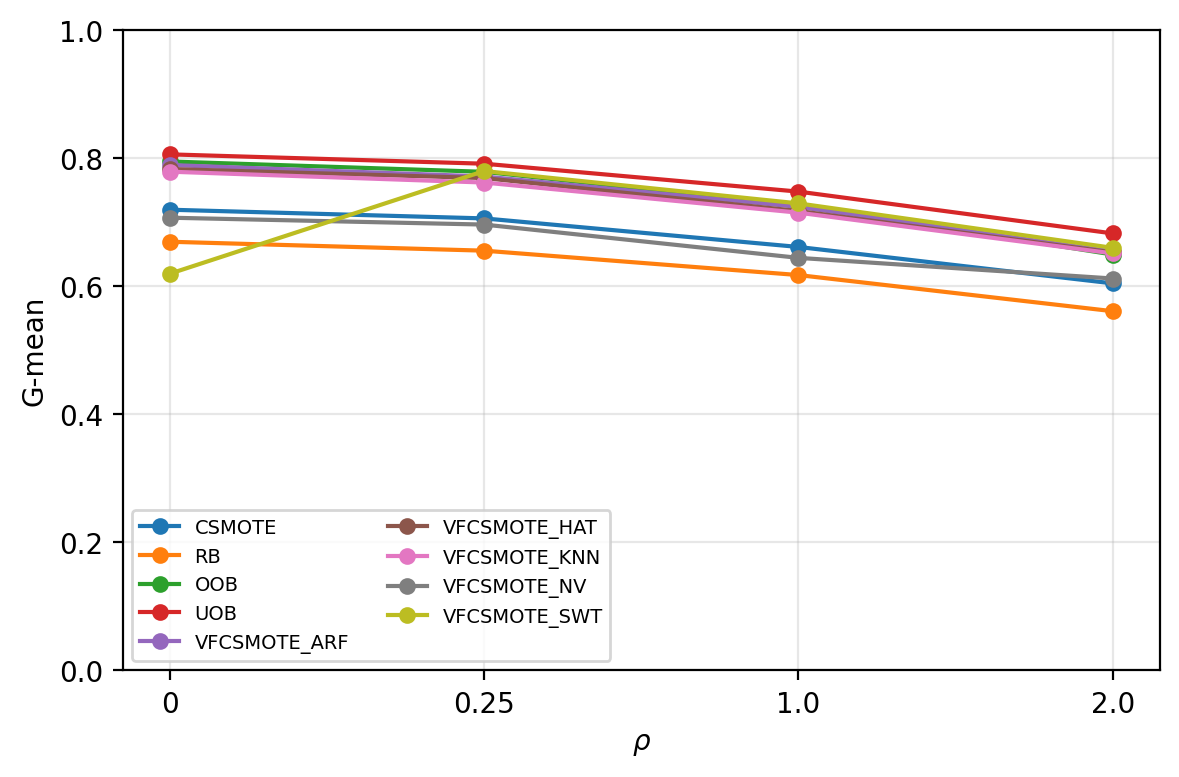}
    \caption{SEA 2:8}
    \label{fig:gmean-sea28}
  \end{subfigure}
  \hfill
  \begin{subfigure}[b]{0.32\textwidth}
    \includegraphics[width=\linewidth]{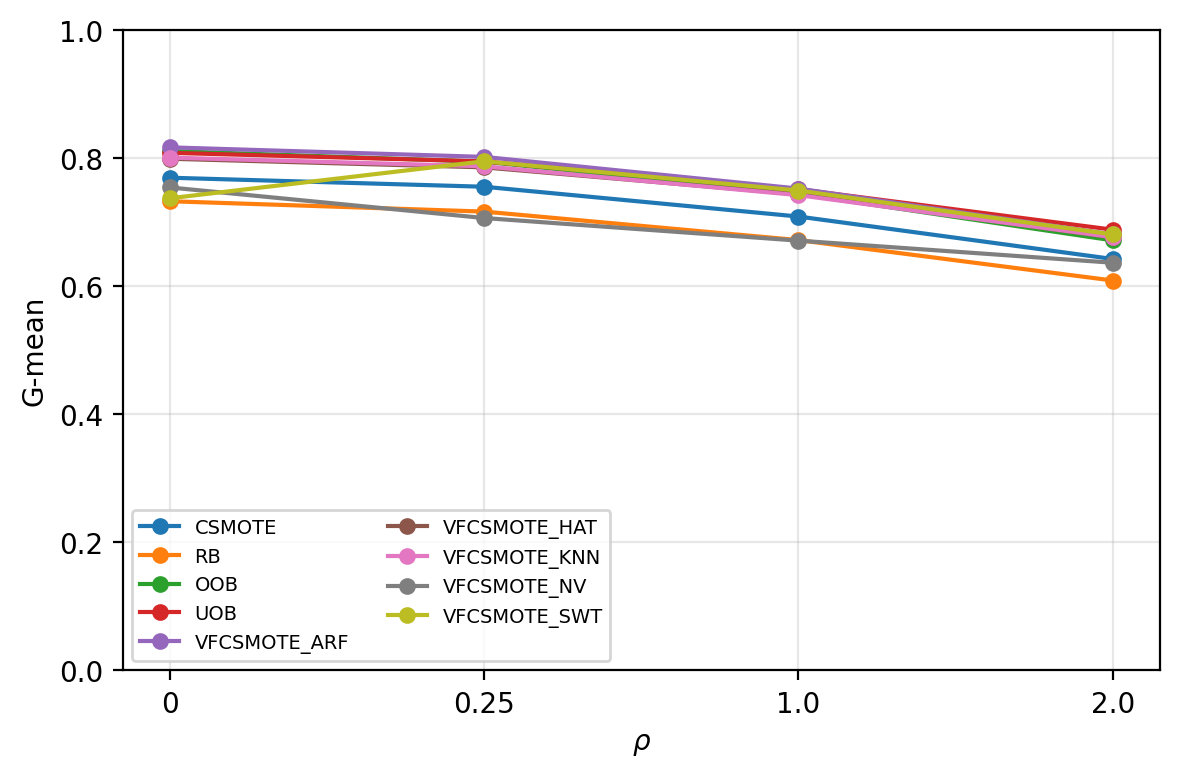}
    \caption{SEA 3:7}
    \label{fig:gmean-sea37}
  \end{subfigure}

  \medskip
  \begin{subfigure}[b]{0.32\textwidth}
    \includegraphics[width=\linewidth]{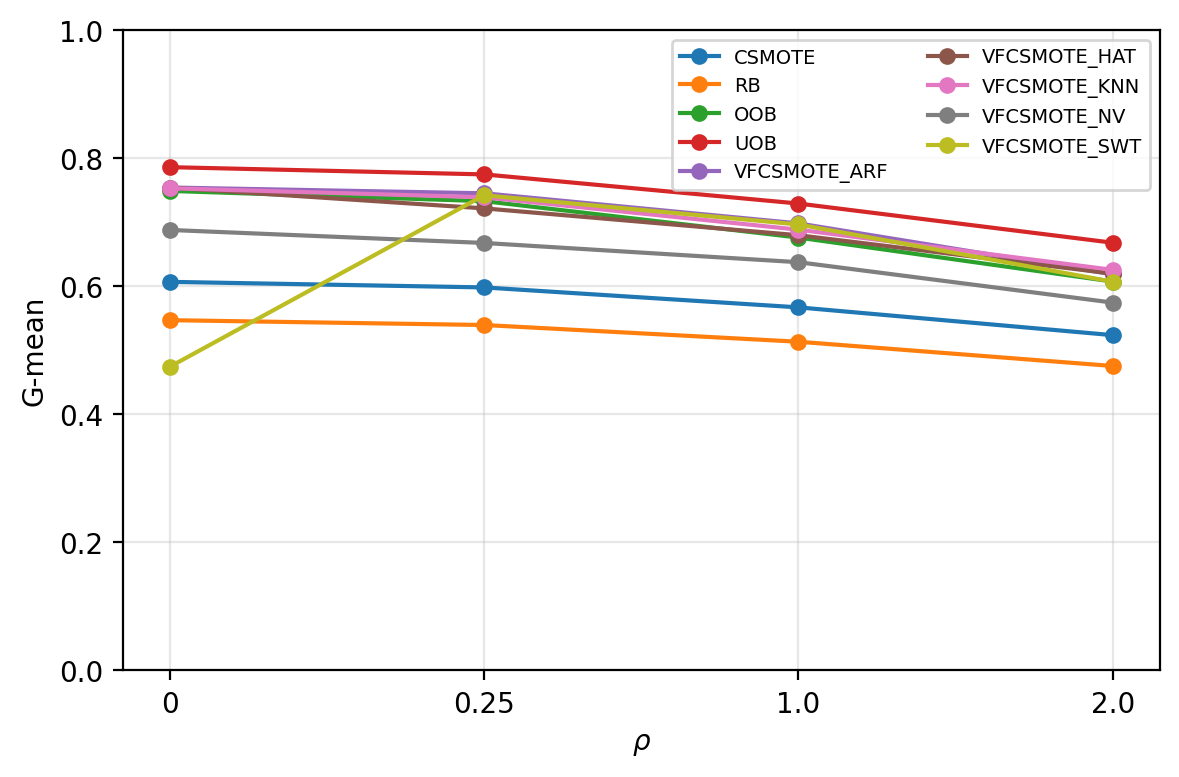}
    \caption{SEA\textsubscript{G} 1:9}
    \label{fig:gmean-seaG19}
  \end{subfigure}
  \hfill
  \begin{subfigure}[b]{0.32\textwidth}
    \includegraphics[width=\linewidth]{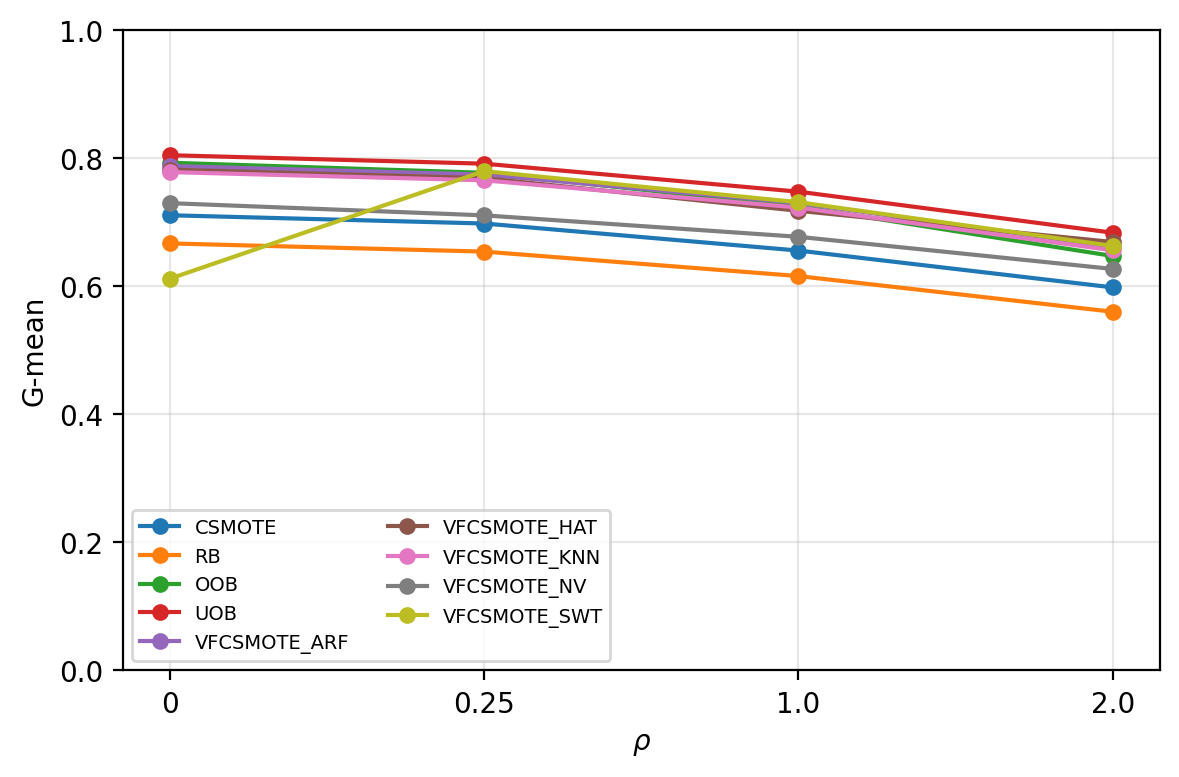}
    \caption{SEA\textsubscript{G} 2:8}
    \label{fig:gmean-seaG28}
  \end{subfigure}
  \hfill
  \begin{subfigure}[b]{0.32\textwidth}
    \includegraphics[width=\linewidth]{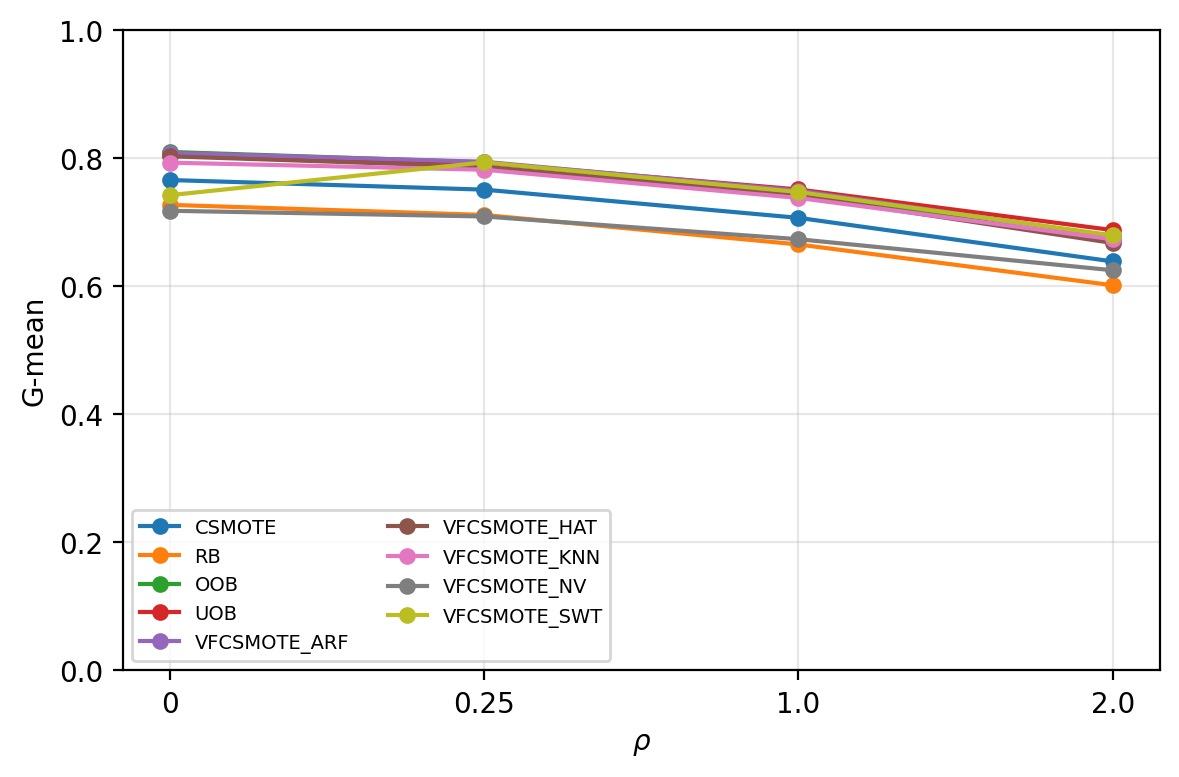}
    \caption{SEA\textsubscript{G} 3:7}
    \label{fig:gmean-seaG37}
  \end{subfigure}

  \caption{Arrival G-mean versus delay-to-drift ratio $\rho$ for all oversampling
           methods, across the six synthetic streams. Top: SEA (abrupt drift);
           bottom: SEA\textsubscript{G} (gradual drift).}
  \label{fig:synthetic-gmean}
\end{figure}

\begin{figure}[htb!]
  \centering
  \begin{subfigure}[b]{0.32\textwidth}
    \includegraphics[width=\linewidth]{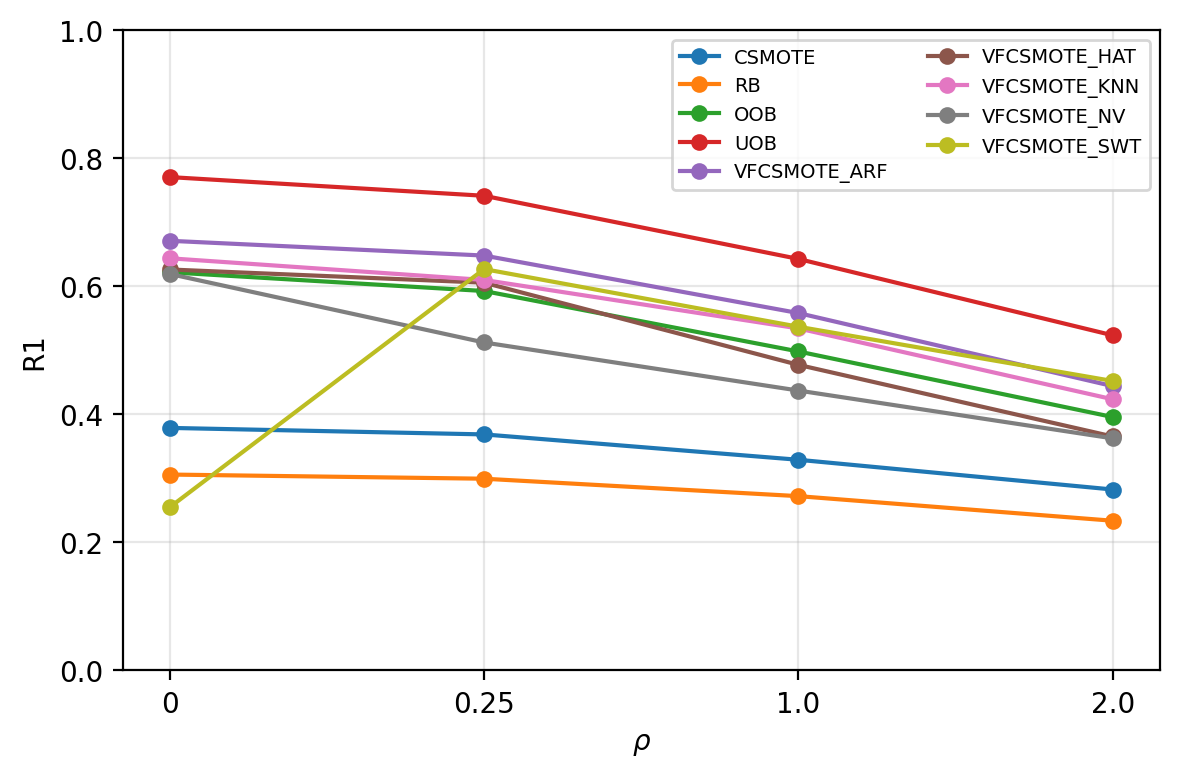}
    \caption{SEA 1:9}
    \label{fig:r1-sea19}
  \end{subfigure}
  \hfill
  \begin{subfigure}[b]{0.32\textwidth}
    \includegraphics[width=\linewidth]{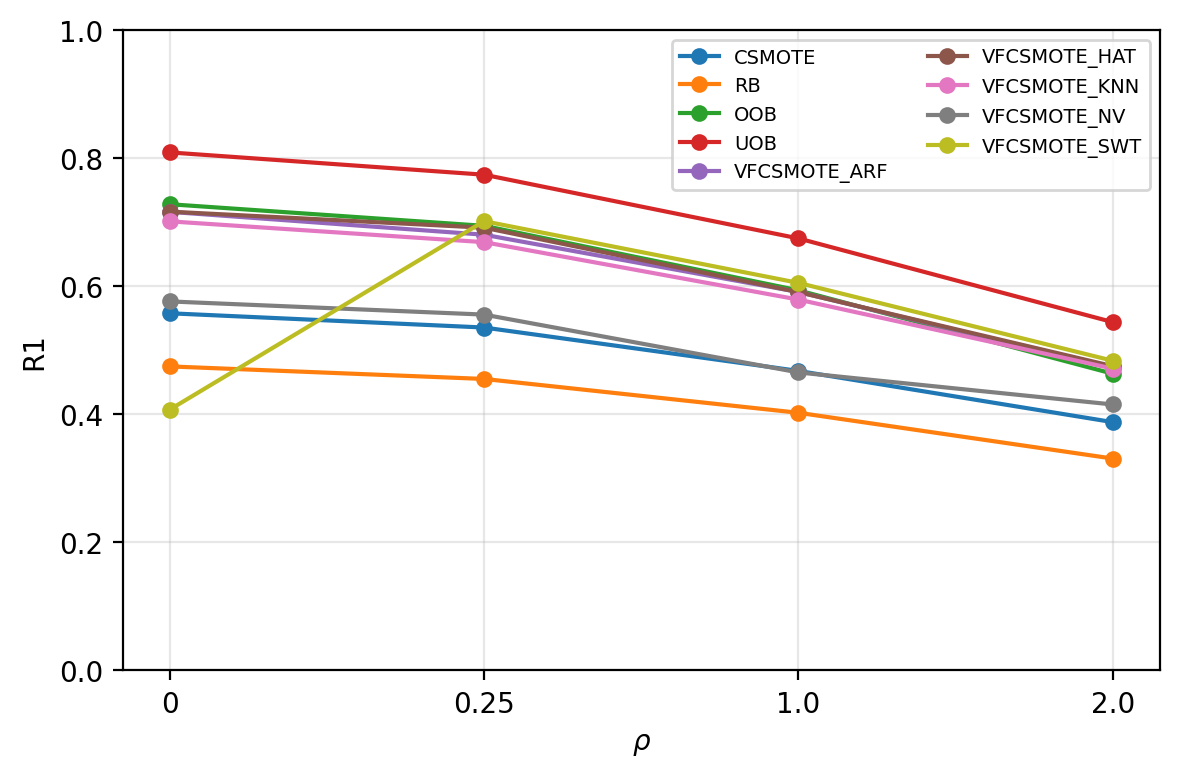}
    \caption{SEA 2:8}
    \label{fig:r1-sea28}
  \end{subfigure}
  \hfill
  \begin{subfigure}[b]{0.32\textwidth}
    \includegraphics[width=\linewidth]{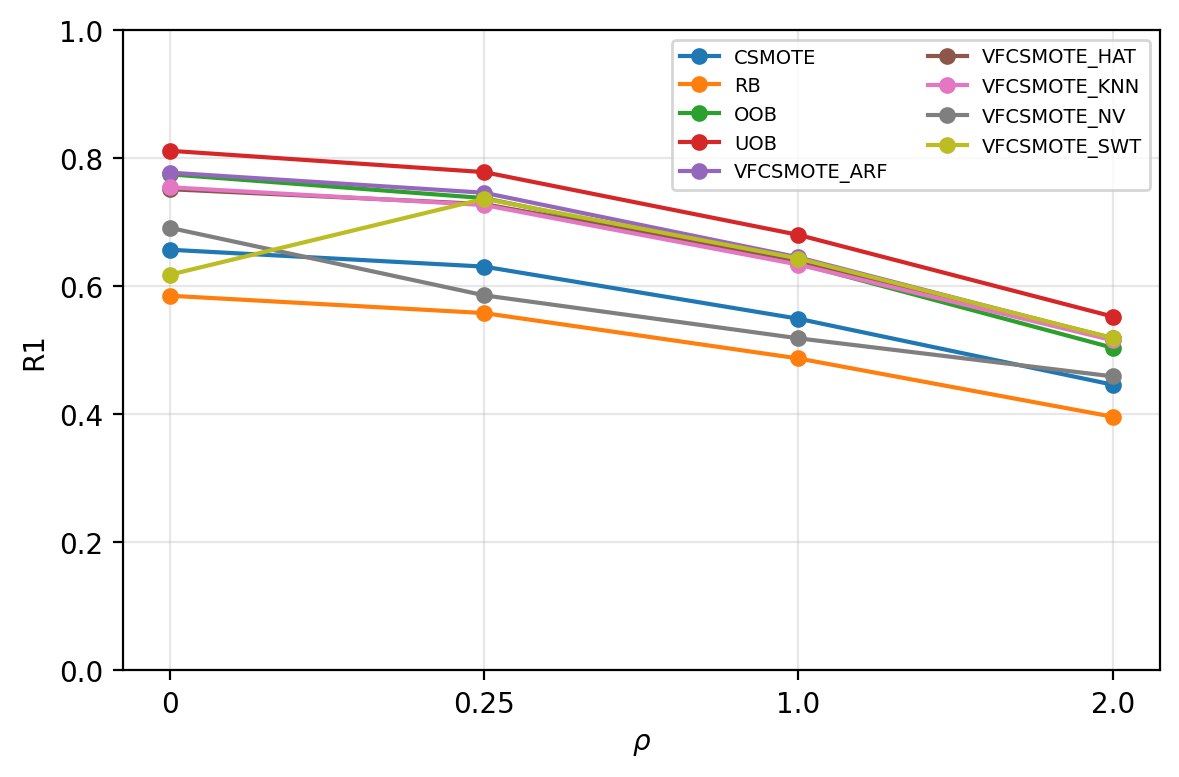}
    \caption{SEA 3:7}
    \label{fig:r1-sea37}
  \end{subfigure}

  \medskip
  \begin{subfigure}[b]{0.32\textwidth}
    \includegraphics[width=\linewidth]{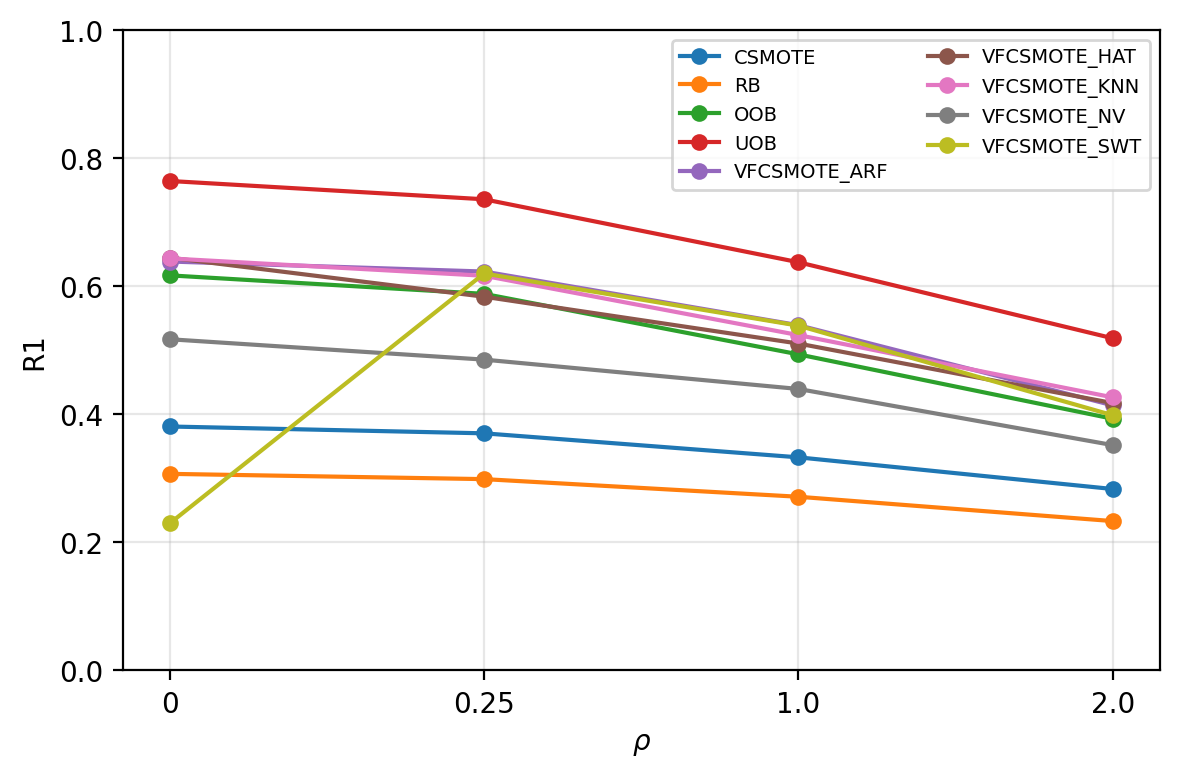}
    \caption{SEA\textsubscript{G} 1:9}
    \label{fig:r1-seaG19}
  \end{subfigure}
  \hfill
  \begin{subfigure}[b]{0.32\textwidth}
    \includegraphics[width=\linewidth]{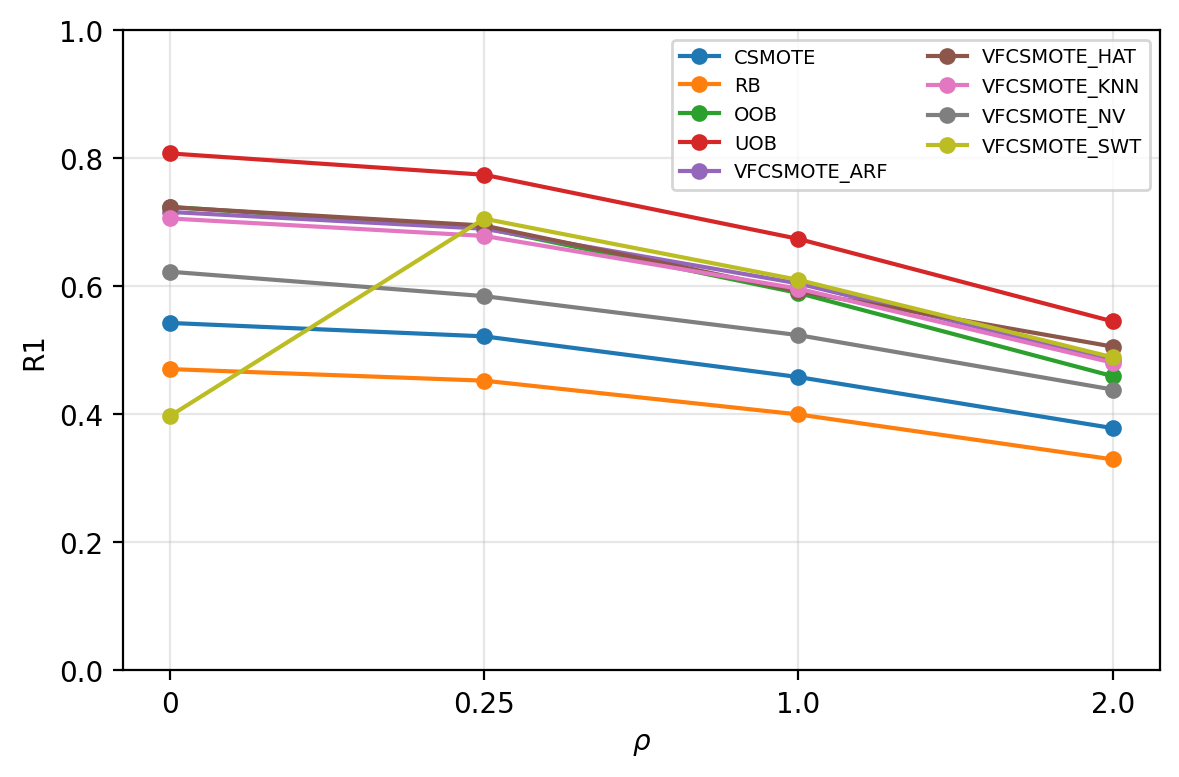}
    \caption{SEA\textsubscript{G} 2:8}
    \label{fig:r1-seaG28}
  \end{subfigure}
  \hfill
  \begin{subfigure}[b]{0.32\textwidth}
    \includegraphics[width=\linewidth]{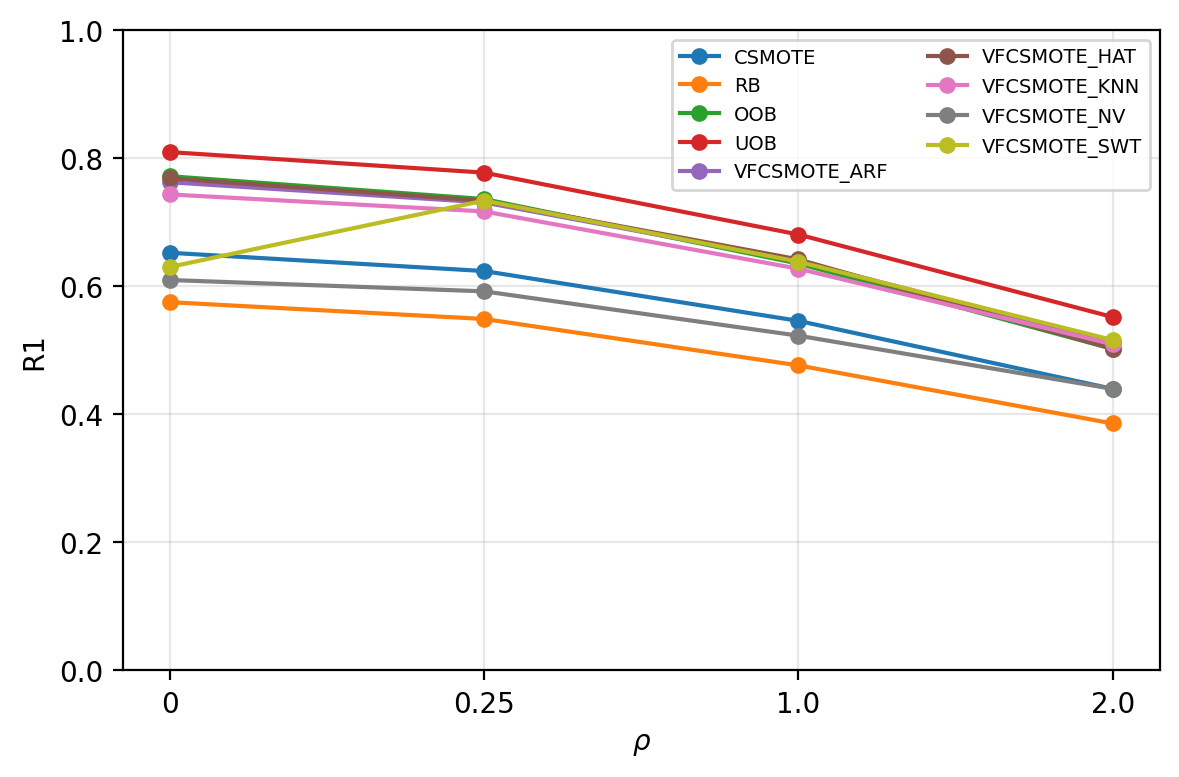}
    \caption{SEA\textsubscript{G} 3:7}
    \label{fig:r1-seaG37}
  \end{subfigure}

  \caption{Arrival R1 versus delay-to-drift ratio $\rho$ for all oversampling
           methods, across the six synthetic streams. Top: SEA (abrupt drift);
           bottom: SEA\textsubscript{G} (gradual drift).}
  \label{fig:synthetic-r1}
\end{figure}

\paragraph{RQ1}
Figures~\ref{fig:synthetic-gmean} and~\ref{fig:synthetic-r1} show the effect of label delays on the arrival G-mean and $R1$ across the six synthetic streams~\footnote{Figures for the other metrics are available in our GitHub repository}.
Across all six streams, arrival performance degrades as the delay-to-drift ratio $\rho$ grows.
One consistent exception is \texttt{VFCSMOTE\_SWT}, which shows a small transient gain at $\rho=0.25$ across all streams before degrading once $\rho \geq 1.0$. SWT~\cite{bifet2013pitfalls} augments the input attributes with the most recent \emph{true} class labels and assumes those labels arrive immediately; under a short delay, the appended labels are only mildly stale and remain autocorrelated with the current instance, so the augmentation still helps, whereas as the delay grows, the appended labels decorrelate and the augmented feature begins to mislead the learner. This is consistent with the delay caveat raised by the authors of SWT~\cite{bifet2013pitfalls}. Because SWT is the only learner in our suite that consumes the label itself as an input feature, its degradation is affected by this additional, stale-feature channel, so it is not directly comparable to learners that use the attribute space alone.

The overall rank of the oversamplers remains stable in most cases, with \texttt{UOB} ranking first. The imbalance ratio causes minor reranking among \texttt{VFCSMOTE\_ARF}, \texttt{OOB}, \texttt{VFCSMOTE\_HAT}, and \texttt{VFCSMOTE\_NV}, but \texttt{UOB} is first regardless of the imbalance ratio.

The imbalance ratio together with the delay regime $\rho$ appear to govern the loss in performance, whereas the drift type (abrupt vs.\ gradual) does not show a clear effect.

These RQ1 results concern \emph{arrival} performance only. The re-evaluation step examined under RQ2 recovers most of this loss, and the refined G-mean reported later in Table~\ref{tab:syntehtic:pareto} is consequently near-invariant to $\rho$; the arrival (RQ1) and refined results are therefore complementary, not contradictory.

\paragraph{RQ2}

\begin{figure}[t]
  \centering
  \begin{subfigure}[b]{0.32\textwidth}
    \includegraphics[width=\linewidth]{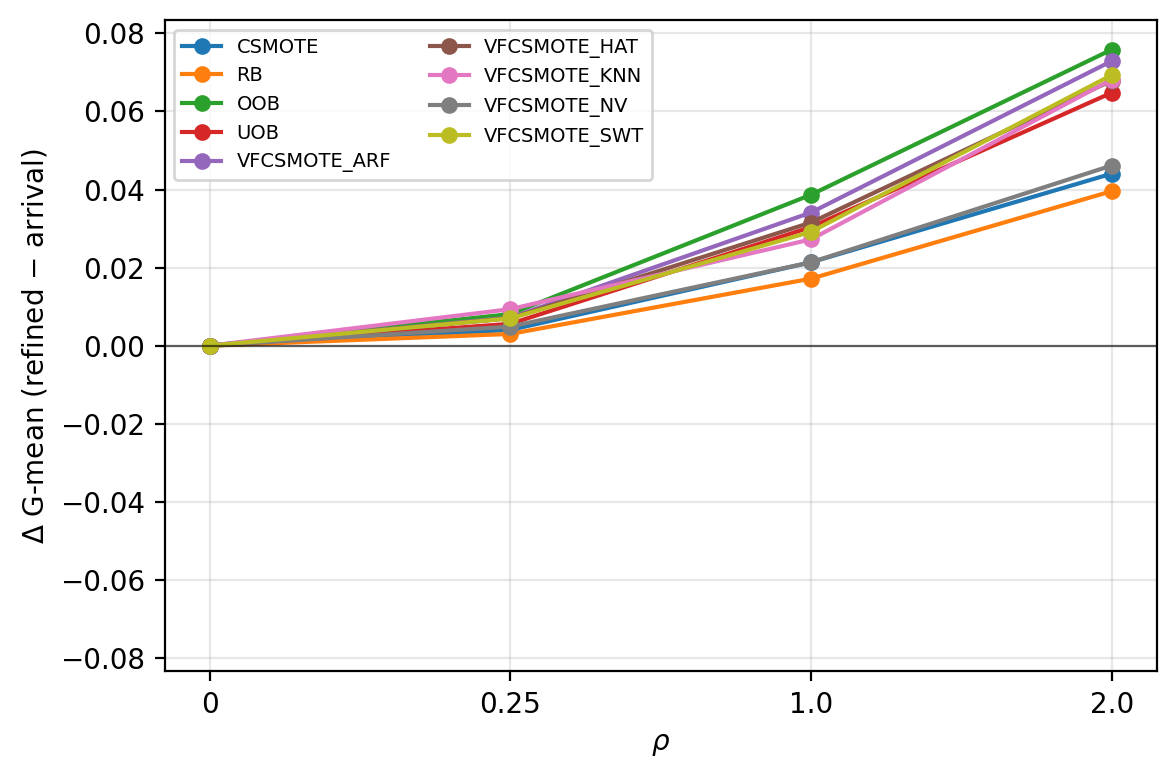}
    \caption{SEA 1:9}
    \label{fig:gmean-sea19:delta}
  \end{subfigure}
  \hfill
  \begin{subfigure}[b]{0.32\textwidth}
    \includegraphics[width=\linewidth]{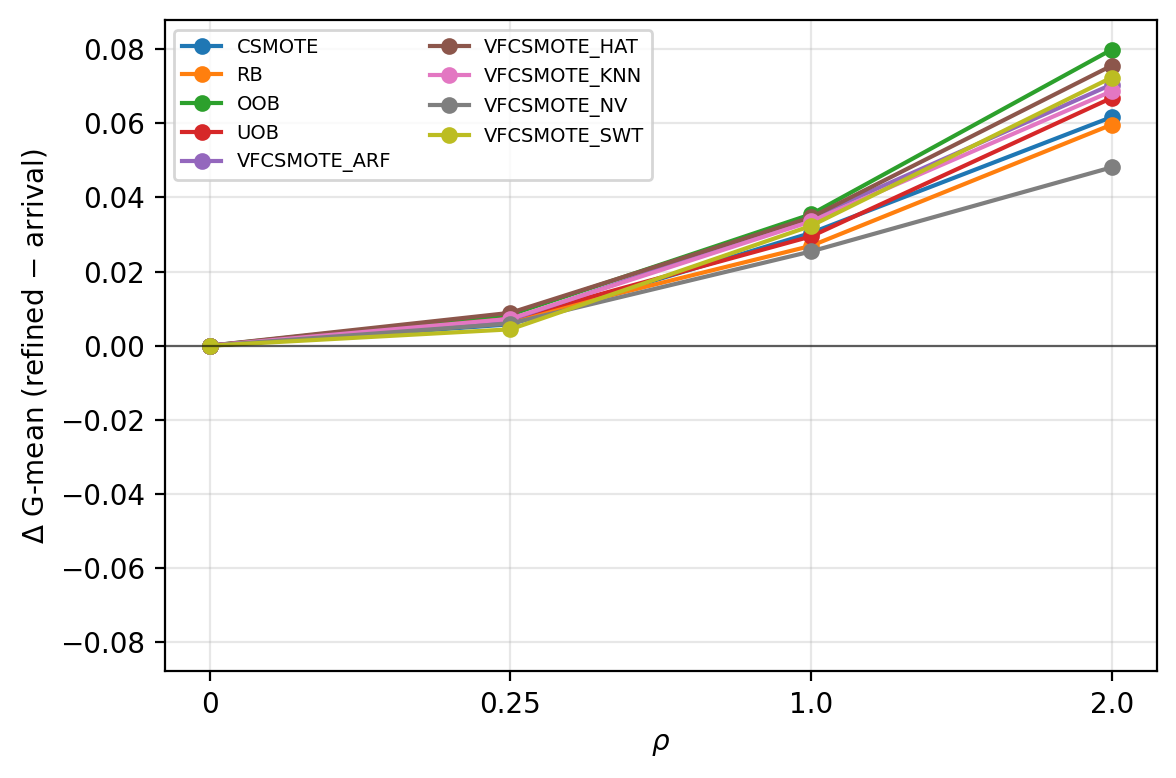}
    \caption{SEA 2:8}
    \label{fig:gmean-sea28:delta}
  \end{subfigure}
  \hfill
  \begin{subfigure}[b]{0.32\textwidth}
    \includegraphics[width=\linewidth]{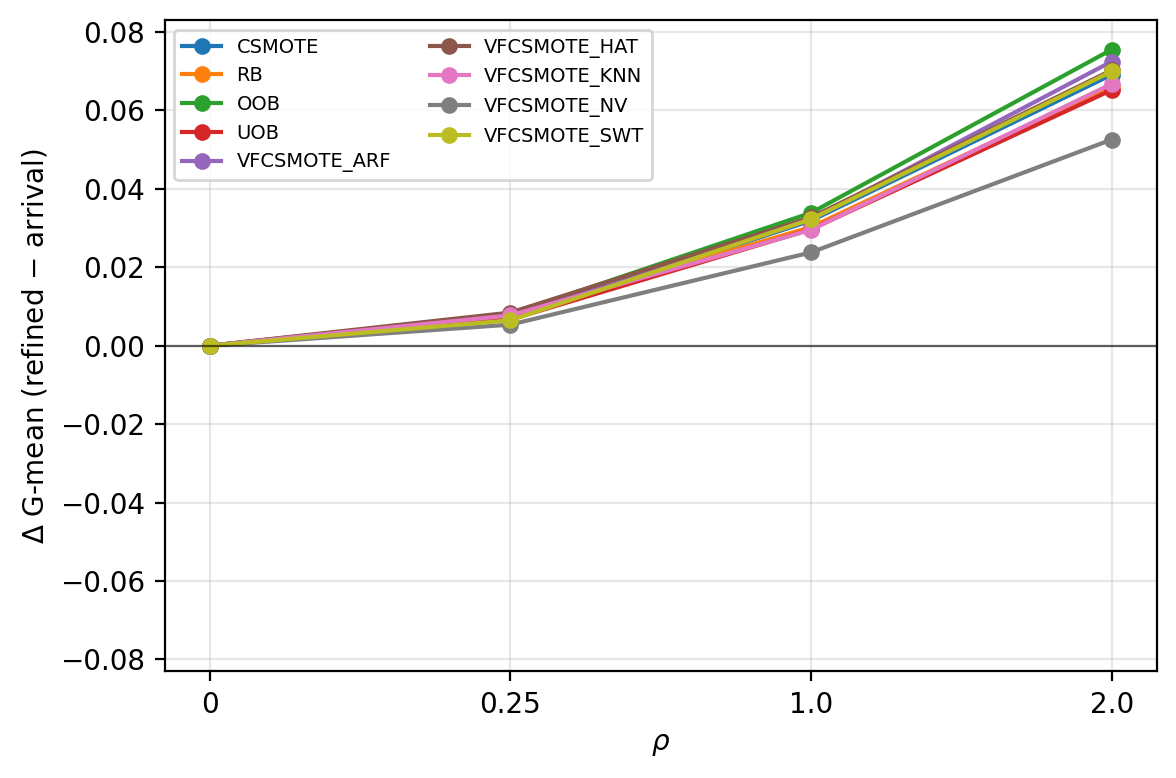}
    \caption{SEA 3:7}
    \label{fig:gmean-sea37:delta}
  \end{subfigure}

  \medskip
  \begin{subfigure}[b]{0.32\textwidth}
    \includegraphics[width=\linewidth]{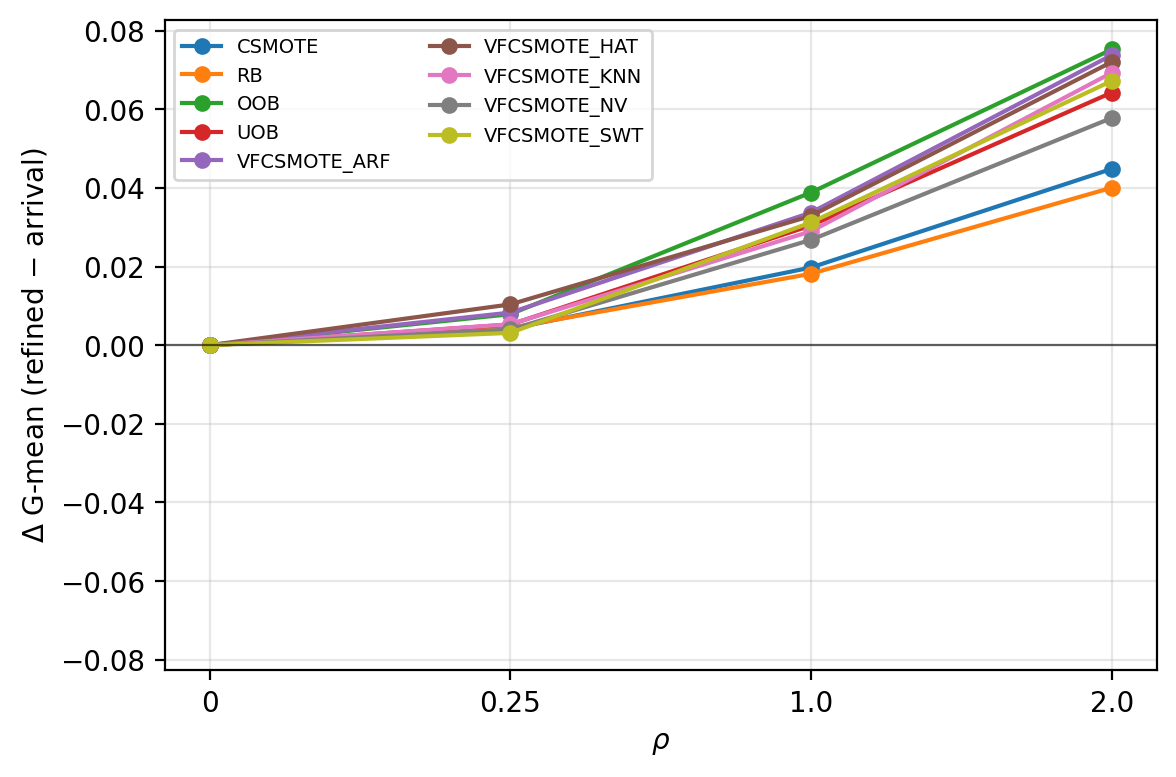}
    \caption{SEA\textsubscript{G} 1:9}
    \label{fig:gmean-seaG19:delta}
  \end{subfigure}
  \hfill
  \begin{subfigure}[b]{0.32\textwidth}
    \includegraphics[width=\linewidth]{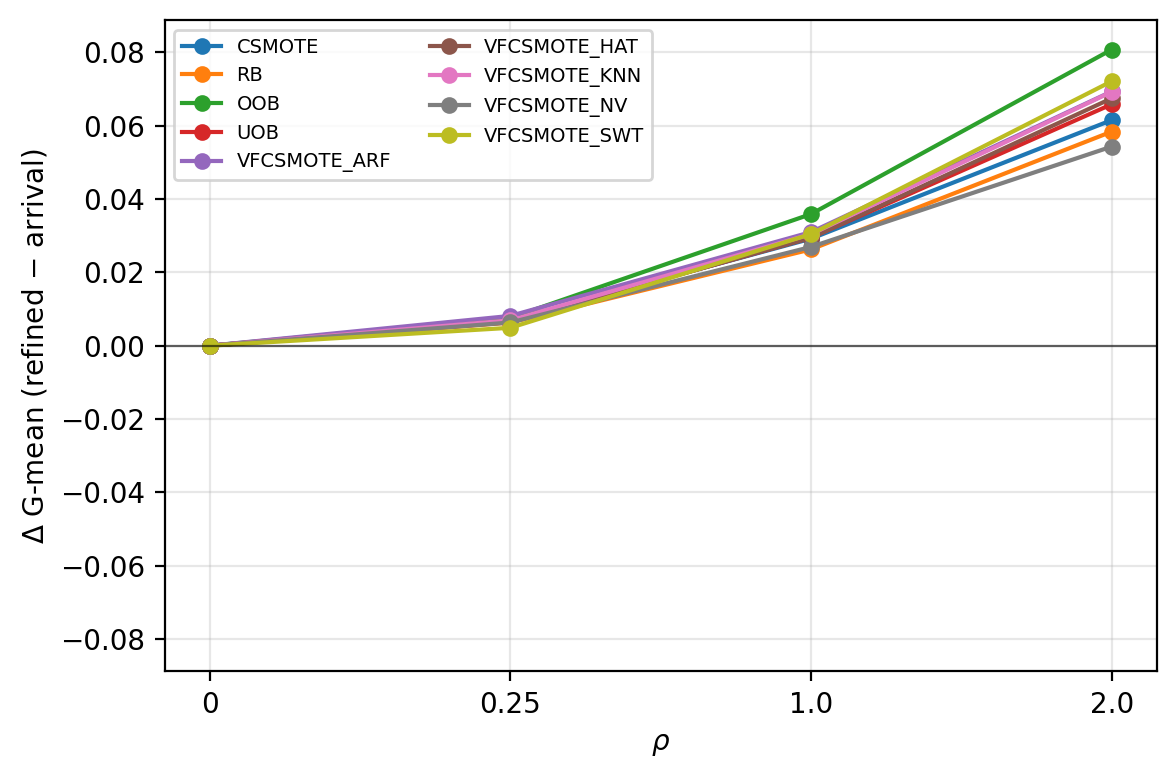}
    \caption{SEA\textsubscript{G} 2:8}
    \label{fig:gmean-seaG28:delta}
  \end{subfigure}
  \hfill
  \begin{subfigure}[b]{0.32\textwidth}
    \includegraphics[width=\linewidth]{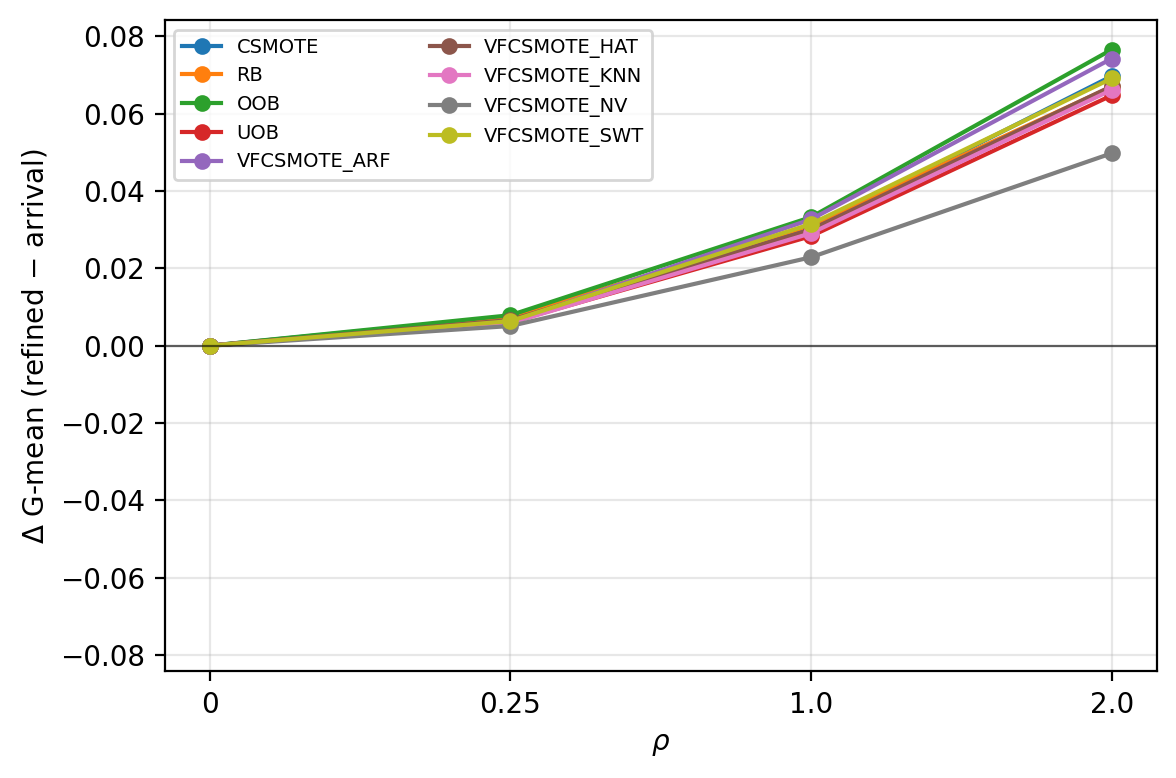}
    \caption{SEA\textsubscript{G} 3:7}
    \label{fig:gmean-seaG37:delta}
  \end{subfigure}

  \caption{$\Delta$ G-mean versus delay-to-drift ratio $\rho$ for all oversampling
           methods, across the six synthetic streams. Top: SEA (abrupt drift);
           bottom: SEA\textsubscript{G} (gradual drift).}
  \label{fig:synthetic-gmean:delta}
\end{figure}

\begin{figure}[htb!]
  \centering
  \begin{subfigure}[b]{0.32\textwidth}
    \includegraphics[width=\linewidth]{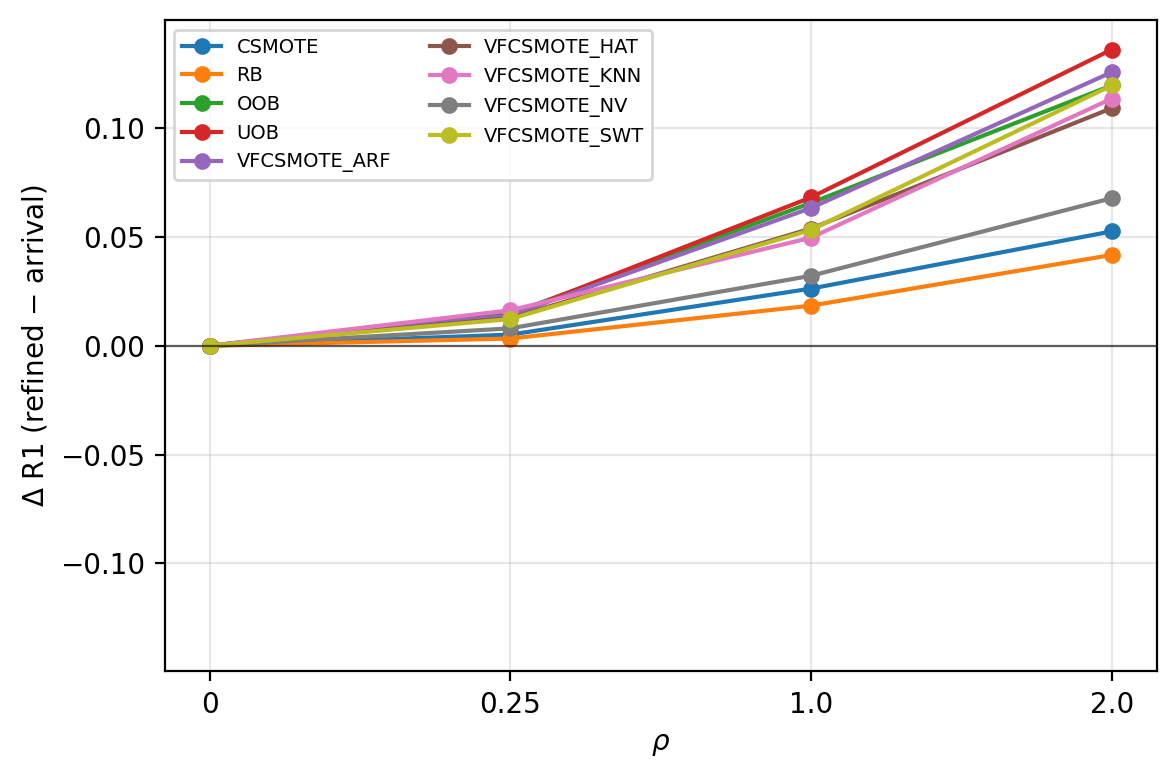}
    \caption{SEA 1:9}
    \label{fig:r1-sea19:delta}
  \end{subfigure}
  \hfill
  \begin{subfigure}[b]{0.32\textwidth}
    \includegraphics[width=\linewidth]{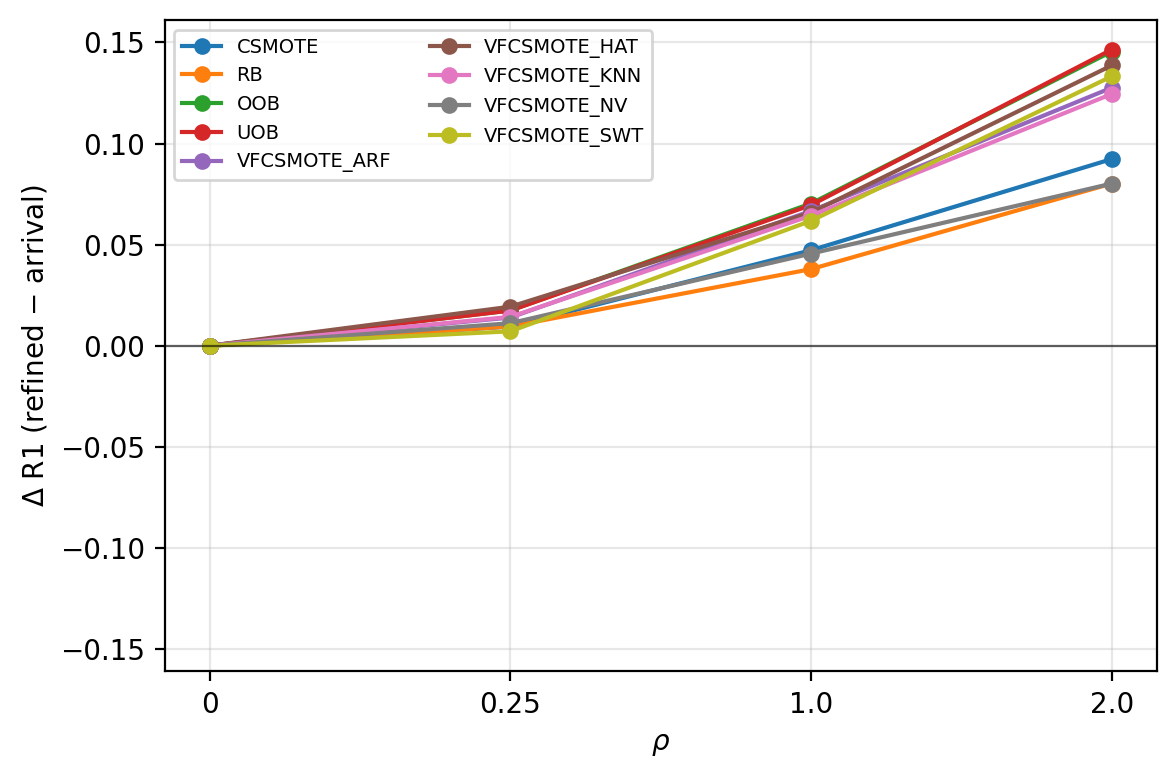}
    \caption{SEA 2:8}
    \label{fig:r1-sea28:delta}
  \end{subfigure}
  \hfill
  \begin{subfigure}[b]{0.32\textwidth}
    \includegraphics[width=\linewidth]{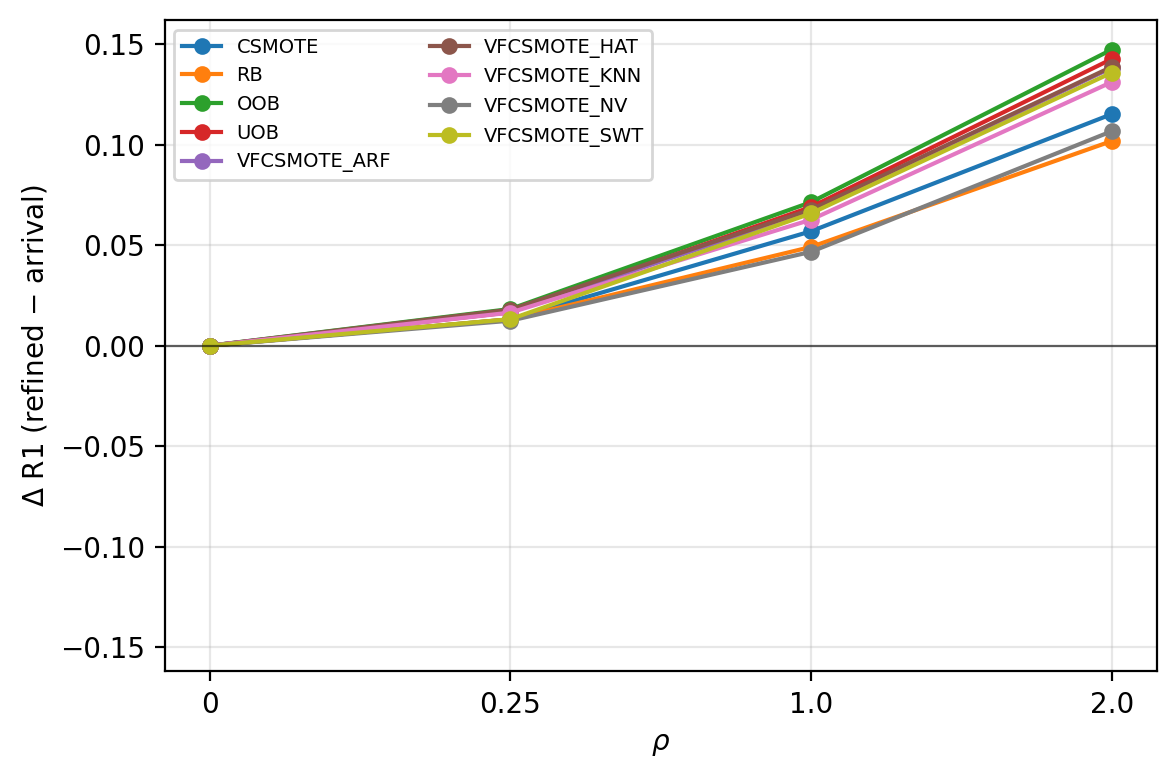}
    \caption{SEA 3:7}
    \label{fig:r1-sea37:delta}
  \end{subfigure}

  \medskip
  \begin{subfigure}[b]{0.32\textwidth}
    \includegraphics[width=\linewidth]{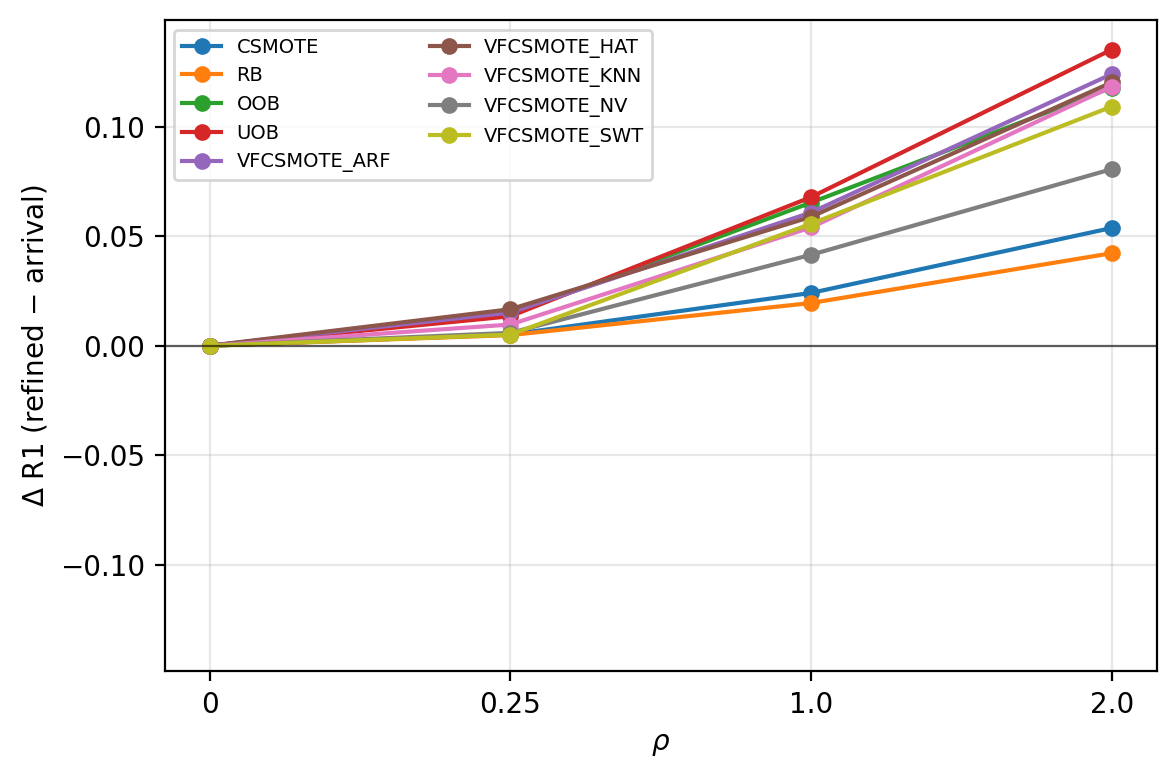}
    \caption{SEA\textsubscript{G} 1:9}
    \label{fig:r1-seaG19:delta}
  \end{subfigure}
  \hfill
  \begin{subfigure}[b]{0.32\textwidth}
    \includegraphics[width=\linewidth]{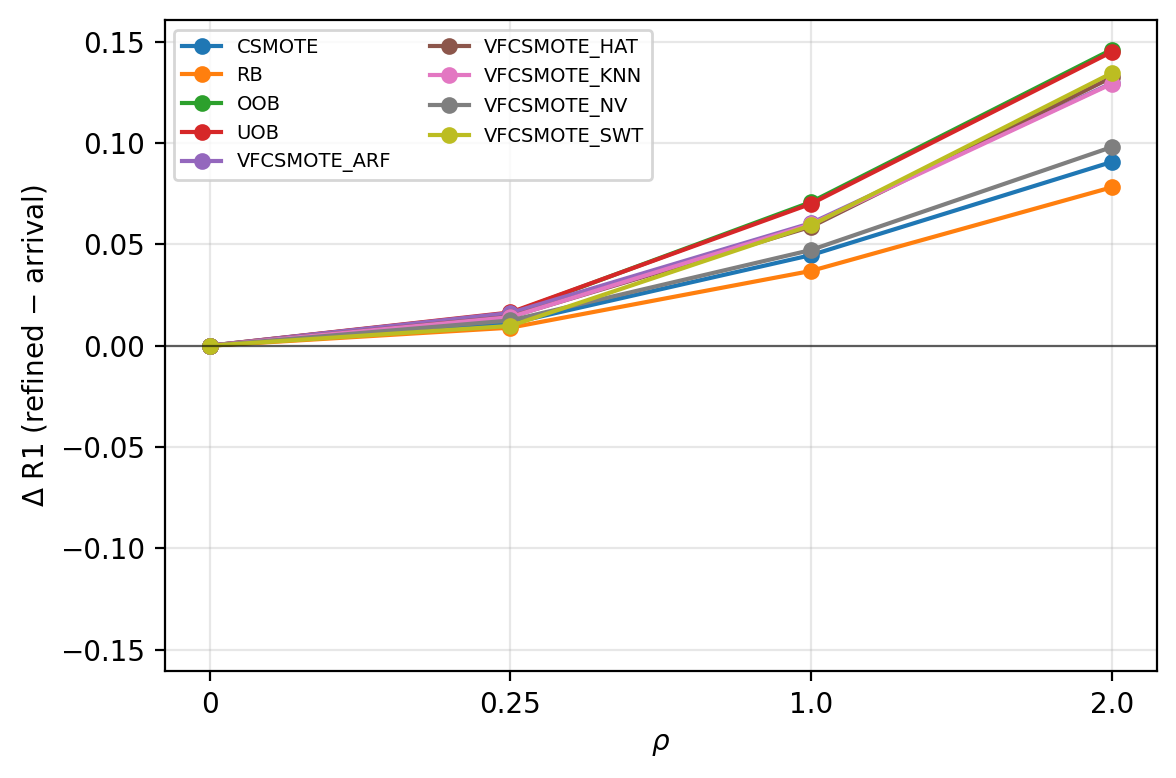}
    \caption{SEA\textsubscript{G} 2:8}
    \label{fig:r1-seaG28:delta}
  \end{subfigure}
  \hfill
  \begin{subfigure}[b]{0.32\textwidth}
    \includegraphics[width=\linewidth]{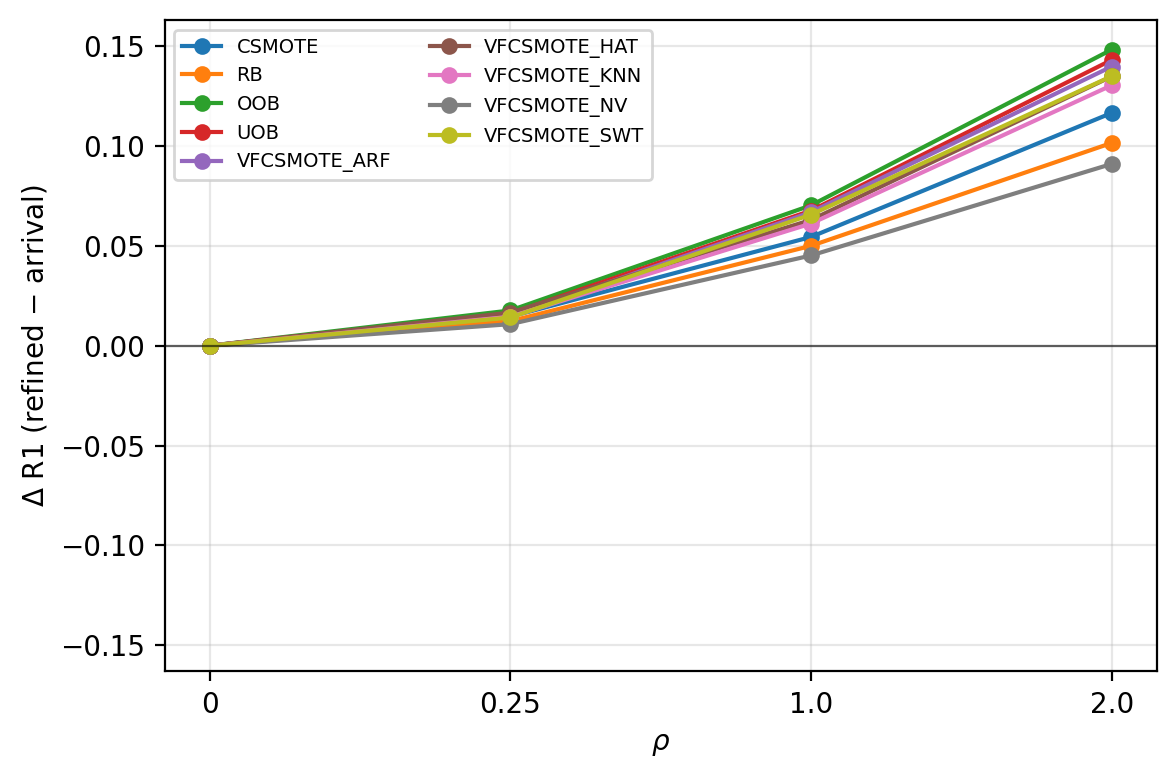}
    \caption{SEA\textsubscript{G} 3:7}
    \label{fig:r1-seaG37:delta}
  \end{subfigure}

  \caption{$\Delta$ $R1$ versus delay-to-drift ratio $\rho$ for all oversampling
           methods, across the six synthetic streams. Top: SEA (abrupt drift);
           bottom: SEA\textsubscript{G} (gradual drift).}
  \label{fig:synthetic-r1:delta}
\end{figure}

Figures~\ref{fig:synthetic-gmean:delta}~and~\ref{fig:synthetic-r1:delta} illustrate the impact of the re-evaluation protocol on G-mean and $R1$. Re-evaluation exhibits a monotonic positive effect across all methods: as $\rho$ increases, an instance waits longer between its arrival-time prediction and the release of its label, and during this interval, the labels of earlier instances arrive and update the online learner, so by the time the instance is re-predicted, the model has been trained on more data and yields a larger refinement gain. As shown in Figure~\ref{fig:synthetic-gmean:delta}, every oversampler achieves a positive gain on all six streams (unanimous direction), reaching the $n=6$ Wilcoxon signed-rank floor of $p = 0.031$. This indicates that re-evaluation is a recovery mechanism that operates across all oversampling strategies, although its magnitude varies by learner (from $+0.0996$ for \texttt{VFCSMOTE\_NV} to $+0.1417$ for \texttt{OOB} at $\rho=2.0$).

At this stage, it is not clear whether the performance improvement is due to oversampling, re-evaluation, or both.

\paragraph{RQ3}
To isolate the marginal contribution of the minority-coverage channel, we
evaluate the difference-in-differences ($\mathrm{DiD}$) (Section~\ref{sec:exp:stats})
for each oversampler--base pair at every delay regime. Only $6$ of the $27$ tested
cells reach significance, and all six sit at the $n=6$ floor of $p=0.031$: most of
the observed gain is therefore attributable to the recovery channel---re-evaluation
operating on the bare learner---rather than to oversampling. The marginal edge
concentrates under extreme delay ($\rho=2.0$), where \texttt{VFCSMOTE\_ARF},
\texttt{VFCSMOTE\_KNN}, and \texttt{UOB} significantly improve over their bare
counterparts; \texttt{OOB} is the only oversampler significant across the entire
spectrum ($p=0.031$ at every $\rho$). The remaining pairs---\texttt{CSMOTE},
\texttt{VFCSMOTE\_HAT}, \texttt{VFCSMOTE\_NV}, \texttt{VFCSMOTE\_SWT}, and
\texttt{RB}---show no detectable marginal contribution at any delay, consistent
with their gains being level effects of the minority-coverage channel preserved
from arrival time rather than interactions with the recovery channel. The
$\mathrm{DiD}$ test inherits the same $n=6$ floor as \textbf{RQ2} and differences
two noisy quantities, so a non-significant cell should be read as ``no detectable
marginal interaction'' rather than evidence of an exactly zero effect; we report
directional, sign-level outcomes without a family-wise correction, which the
$0.031$ floor would in any case preclude. Because these methods incur substantial
replay overhead, it is natural to ask which configurations yield the best
cost-to-benefit ratio.

\paragraph{RQ4}

While \textbf{RQ3} isolates oversampling's marginal contribution in differences, \textbf{RQ4} asks which complete configuration a practitioner should deploy, comparing absolute refined G-mean against replay overhead (Table~\ref{tab:syntehtic:pareto}).\footnote{For readability, Table~\ref{tab:syntehtic:pareto} lists only the Pareto-optimal configurations and, for each $\rho$, the nearest dominated configuration; the complete 15-method results at every $\rho$ are available in the project repository (Section~\ref{sec:exp:impl}).} The two lenses can disagree, and the clearest example is \texttt{CSMOTE}: it shows no significant $\mathrm{DiD}$ relative to \texttt{ARF} (\textbf{RQ3}) yet is Pareto-optimal at every $\rho$ (\textbf{RQ4}). This is exactly the marginal-versus-absolute distinction; the minority-coverage channel raises the absolute arrival-time level, which the recovery channel then preserves rather than amplifies. So, Pareto standing should be read as deployment guidance rather than a restatement of the attribution analysis.

Re-evaluation acts as the learner-agnostic recovery channel: every bare learner regains comparable performance under delay at zero replay overhead. Among the zero-replay options, \texttt{HAT} matches \texttt{ARF}'s refined G-mean ($0.656$) with a substantially lighter model, a single adaptive tree rather than a forest ensemble, so we take it as the preferred bare learner when computational economy matters. (Overhead here counts replayed instances only; the re-evaluation cost itself is common to every method, including the bare learners.)

The minority-coverage channel then buys incremental absolute performance at a replay cost: \texttt{CSMOTE} reaches $0.696$ at $\approx 100$k replays and \texttt{VFCSMOTE\_ARF} reaches $0.789$ at $\approx 130$k. \texttt{VFCSMOTE\_ARF} is the only VFCSMOTE-based variant on the frontier at every $\rho$; \texttt{VFCSMOTE\_SWT} drops off the frontier as delay increases, while \texttt{VFCSMOTE\_KNN}'s membership is unstable, on the front at $\rho=0.25$ and $\rho=2.0$ but dominated at $\rho=1.0$. The highest refined G-mean, $0.800$, is delivered by \texttt{UOB} at an average overhead of $\approx 335$k replayed instances.

\texttt{OOB} never reaches the frontier despite its strong marginal performance in \textbf{RQ3} (Figure~\ref{fig:synthetic-r1:delta}): \texttt{UOB} strictly dominates it at every $\rho$, with both a higher refined G-mean ($0.800$ vs\ $0.785$) and far lower overhead ($\approx 335$k vs $\approx 1.5$M replays). This is the sharpest case of the marginal-versus-absolute distinction: \texttt{OOB} interacts more effectively with the recovery channel, but \texttt{UOB} starts from a higher minority-coverage baseline, making it the better deployment choice.

\begin{table}[ht!]
\centering\small
\caption{Mean replayed and refined G-mean across synthetic streams. Only
         Pareto-optimal configurations (Y) and, per delay regime, the nearest dominated configuration (N) are shown.}
\label{tab:syntehtic:pareto}
\renewcommand{\arraystretch}{1.2}
\begin{tabular}{lccc}
\toprule
Learner & Overhead & G-mean & Pareto \\
\midrule
\multicolumn{4}{c}{$\rho=0.25$ (Pareto-optimal: 7/15)} \\
\midrule
ARF           & 0.0       & 0.656 & Y \\
HAT           & 0.0       & 0.656 & Y \\
CSMOTE        & 99798.0   & 0.696 & Y \\
VFCSMOTE\_KNN & 128797.3  & 0.777 & Y \\
VFCSMOTE\_SWT & 130031.0  & 0.785 & Y \\
VFCSMOTE\_HAT & 133100.7  & 0.769 & N \\
VFCSMOTE\_ARF & 133925.0  & 0.789 & Y \\
UOB           & 335489.8  & 0.800 & Y \\
\midrule
\multicolumn{4}{c}{$\rho=1.0$ (Pareto-optimal: 5/15)} \\
\midrule
ARF           & 0.0       & 0.656 & Y \\
HAT           & 0.0       & 0.656 & Y \\
CSMOTE        & 99798.0   & 0.696 & Y \\
VFCSMOTE\_ARF & 127200.3  & 0.789 & Y \\
VFCSMOTE\_KNN & 128781.0  & 0.777 & N \\
UOB           & 335489.8  & 0.800 & Y \\
\midrule
\multicolumn{4}{c}{$\rho=2.0$ (Pareto-optimal: 6/15)} \\
\midrule
ARF           & 0.0       & 0.656 & Y \\
HAT           & 0.0       & 0.656 & Y \\
CSMOTE        & 99798.0   & 0.696 & Y \\
VFCSMOTE\_KNN & 129014.7  & 0.776 & Y \\
VFCSMOTE\_ARF & 131504.7  & 0.788 & Y \\
VFCSMOTE\_SWT & 133763.7  & 0.784 & N \\
UOB           & 335489.8  & 0.800 & Y \\
\bottomrule
\end{tabular}
\end{table}

Finally, refined G-mean is essentially invariant to $\rho$ for every method (e.g.\ \texttt{ARF} and \texttt{HAT} at $0.656$, \texttt{UOB} at $0.800$, \texttt{CSMOTE} at $0.696$ across all three regimes): the final refinement is taken at most $K=100$ labels before label release, so it reflects a model trained on nearly all labels that arrived during the wait, regardless of the wait's length. The per-$\rho$ frontier counts ($7$, $5$, and $6$ optimal methods at $\rho=0.25$, $1.0$, and $2.0$) therefore reflect small overhead orderings and sub-$0.01$ G-mean differences near ties, not a delay-driven change in which methods are deployable. We accordingly do not interpret the per-$\rho$ membership changes as a substantive trend.

From the synthetic analysis, we carry three candidates forward to real-data scrutiny. \texttt{UOB} is the synthetic top-line: Pareto-optimal at every $\rho$, the highest refined G-mean ($0.800$), and a significant marginal contribution over \texttt{OzaBag} at $\rho=2.0$ (Figure~\ref{fig:synthetic-gmean:delta}). \texttt{VFCSMOTE\_ARF} is the most consistent synthesis-based pick: on the frontier at every $\rho$ and the only VFCSMOTE variant approaching significance against its bare counterpart below $\rho=2.0$ ($p=0.062$ at $\rho=1.0$). \texttt{VFCSMOTE\_KNN} is carried forward for a different reason, not for consistency but as the strongest synthetic \emph{case} for synthesis-based oversampling. Among Pareto-optimal VFCSMOTE pairings, it delivers the largest absolute lift over its bare counterpart ($+0.169$ at $\rho=0.25$, $+0.168$ at $\rho=2.0$, versus $+0.133$ for \texttt{VFCSMOTE\_ARF}); it pairs with the weakest bare learner that appears in any Pareto-optimal VFCSMOTE pair (\texttt{KNN}, refined G-mean $0.608$), so this lift represents the most dramatic transformation of a base classifier by oversampling among the frontier candidates; and it joins \texttt{VFCSMOTE\_ARF} as the only VFCSMOTE variants with significant marginal contribution in RQ3 ($p=0.031$ at $\rho=2.0$).
Its Pareto membership is not unbroken --- at $\rho=1.0$ it is dominated by
\texttt{VFCSMOTE\_ARF} by margins of $\approx 1.6$k replays and $\approx 0.012$ in G-mean --- so we treat its synthetic case as plausible rather than decisive. Whether either VFCSMOTE candidate survives validation on real streams is the central question of the following subsection.

\clearpage
\subsection{Real Datasets}
\label{sec:results:real}

\paragraph{RQ1}

\begin{figure}[htb!]
  \centering
  \begin{subfigure}[b]{0.32\textwidth}
    \includegraphics[width=\linewidth]{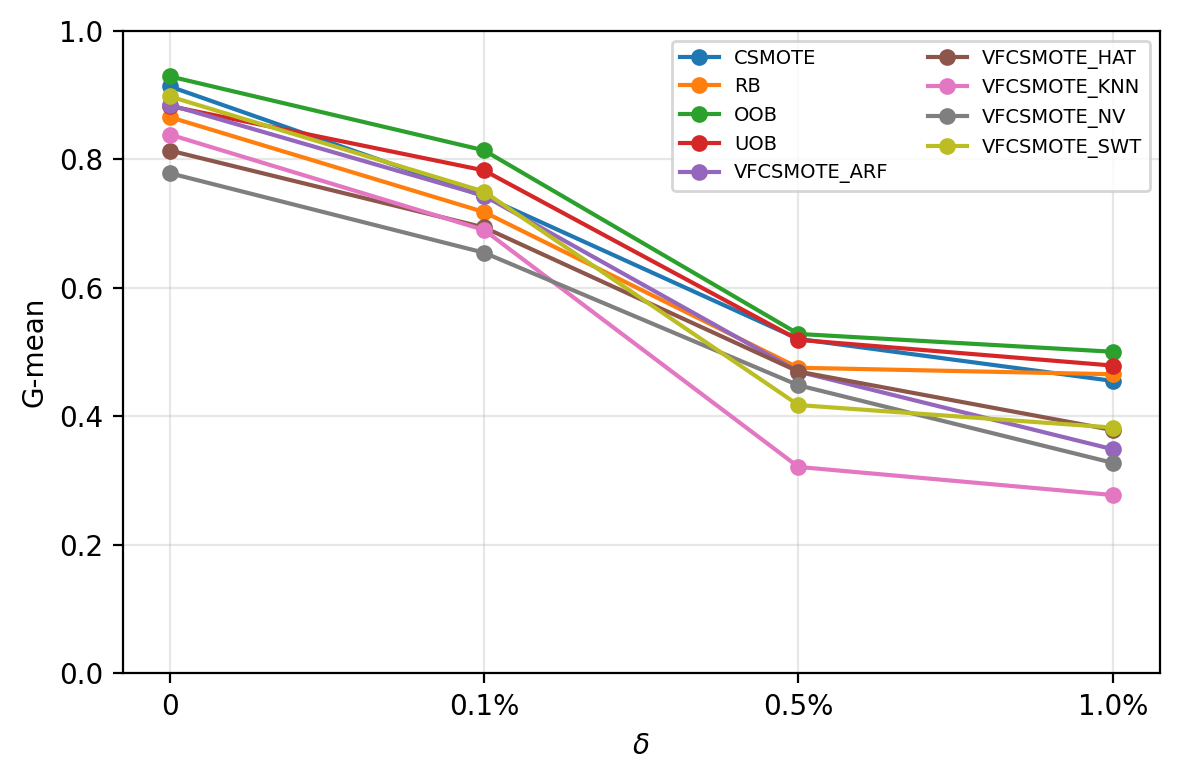}
    \caption{Cover type}
    \label{fig:gmean-covertype}
  \end{subfigure}
  \hfill
  \begin{subfigure}[b]{0.32\textwidth}
    \includegraphics[width=\linewidth]{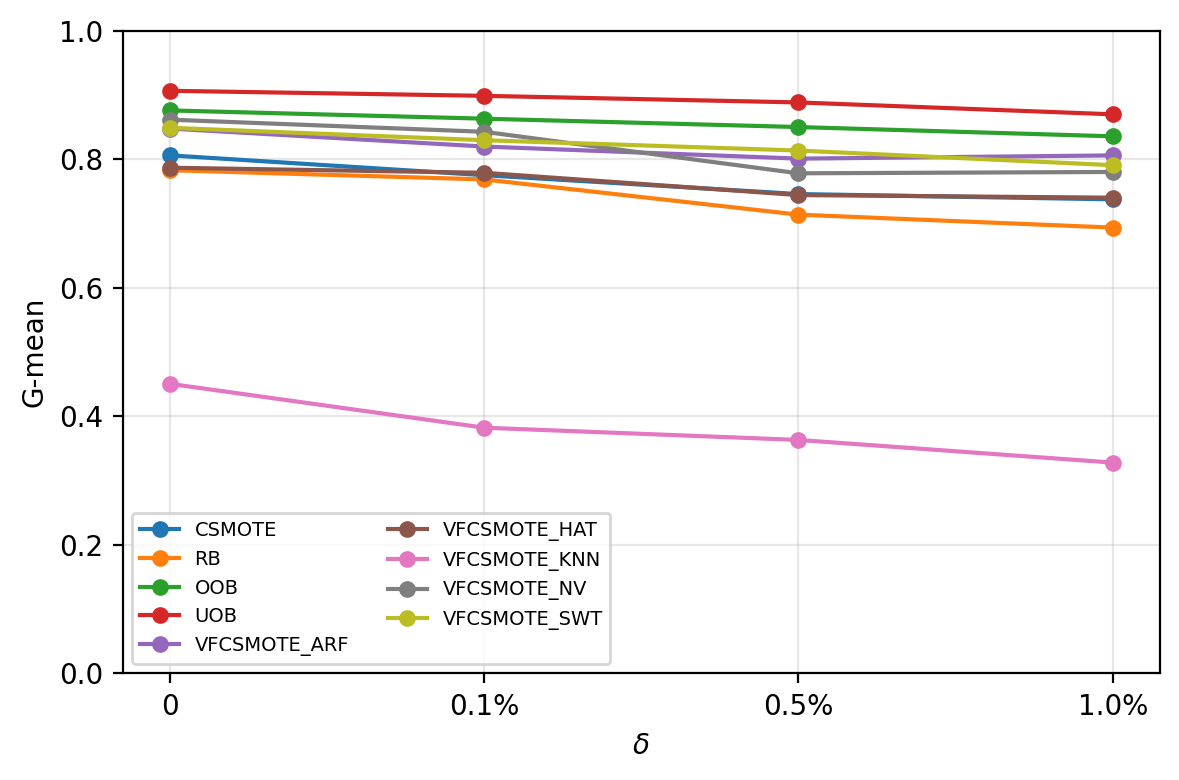}
    \caption{credit card}
    \label{fig:gmean-creditcard}
  \end{subfigure}
  \hfill
  \begin{subfigure}[b]{0.32\textwidth}
    \includegraphics[width=\linewidth]{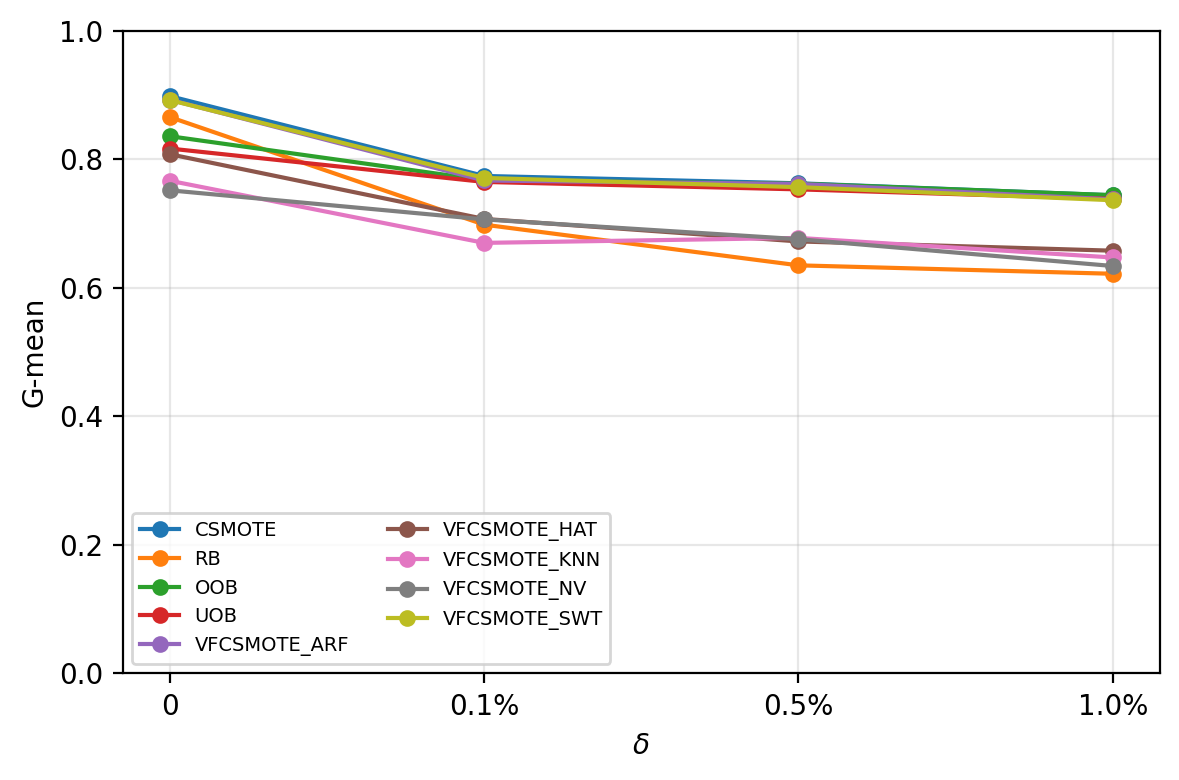}
    \caption{Electricity}
    \label{fig:gmean-electricity}
  \end{subfigure}

  \medskip
  \begin{subfigure}[b]{0.32\textwidth}
    \includegraphics[width=\linewidth]{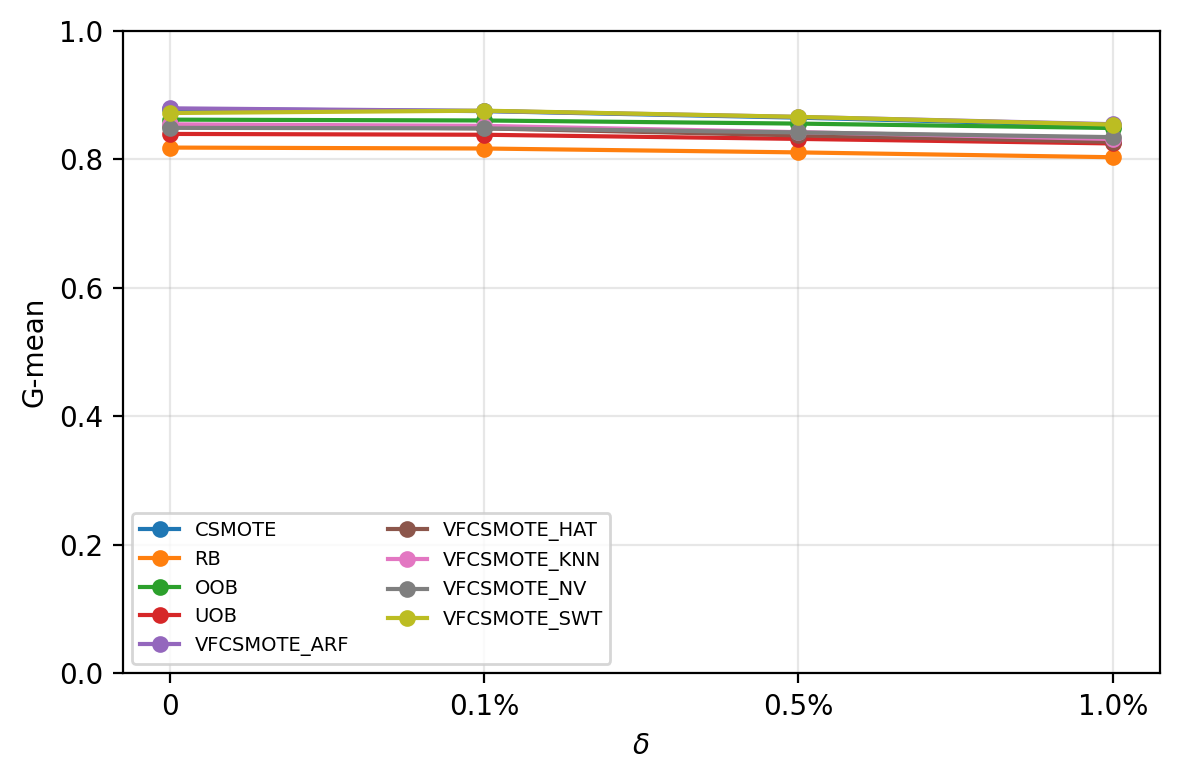}
    \caption{Insects}
    \label{fig:gmean-insects}
  \end{subfigure}
  \hfill
  \begin{subfigure}[b]{0.32\textwidth}
    \includegraphics[width=\linewidth]{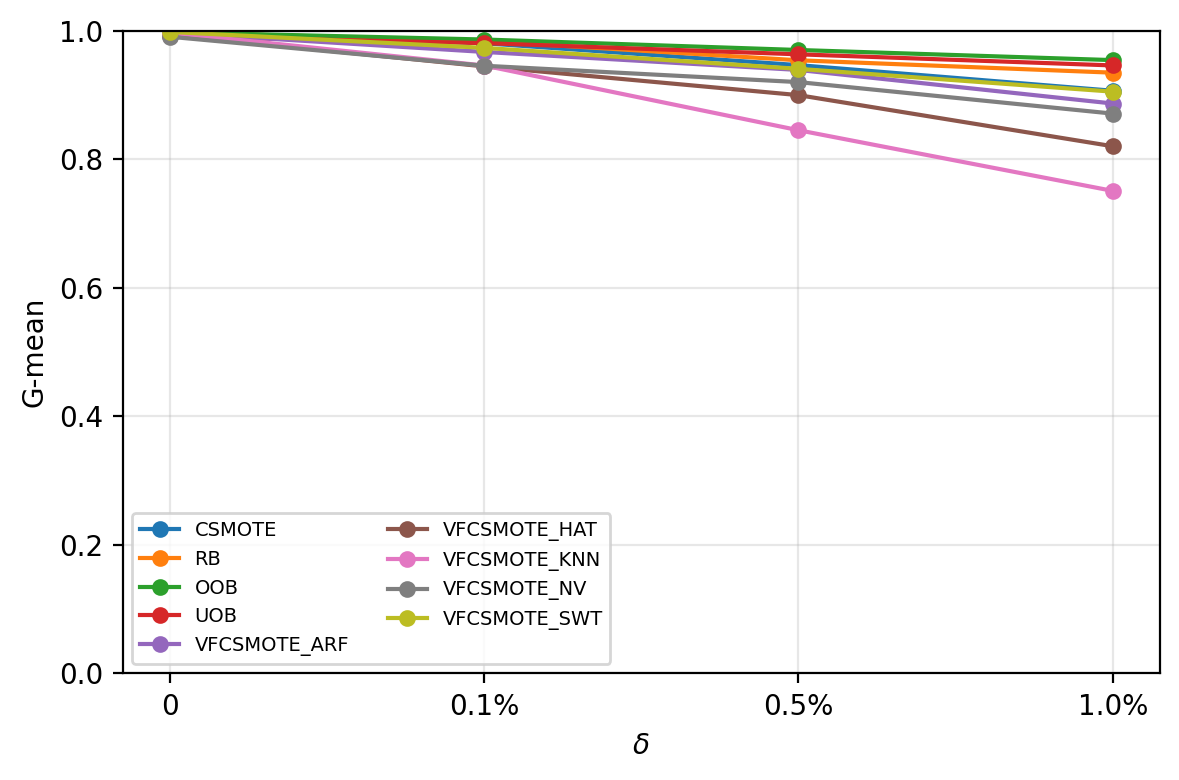}
    \caption{KDD Cup}
    \label{fig:gmean-KDDCup}
  \end{subfigure}
  \hfill
  \begin{subfigure}[b]{0.32\textwidth}
    \includegraphics[width=\linewidth]{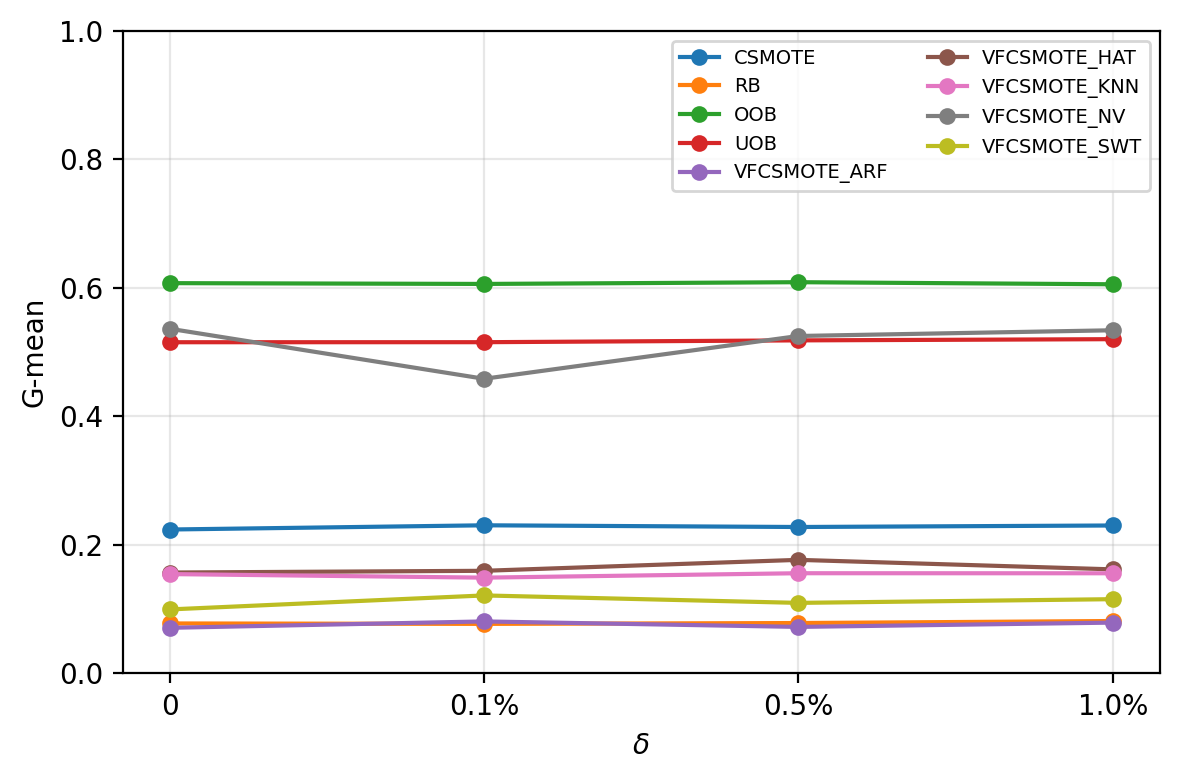}
    \caption{PAKDD}
    \label{fig:gmean-packdd}
  \end{subfigure}

  \caption{Arrival G-mean versus delay  $\delta$ for all oversampling
           methods across the six real data streams.}
  \label{fig:real-gmean}
\end{figure}

\begin{figure}[htb!]
  \centering
  \begin{subfigure}[b]{0.32\textwidth}
    \includegraphics[width=\linewidth]{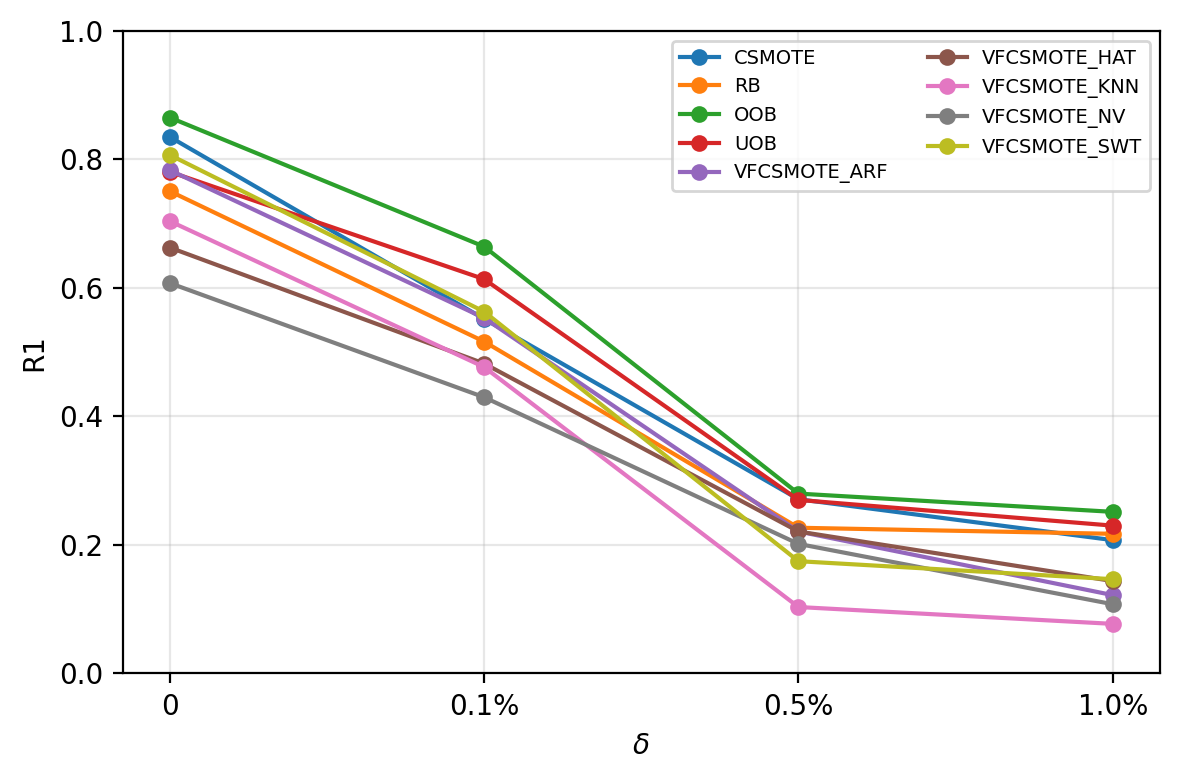}
    \caption{Cover type}
    \label{fig:r1-covertype}
  \end{subfigure}
  \hfill
  \begin{subfigure}[b]{0.32\textwidth}
    \includegraphics[width=\linewidth]{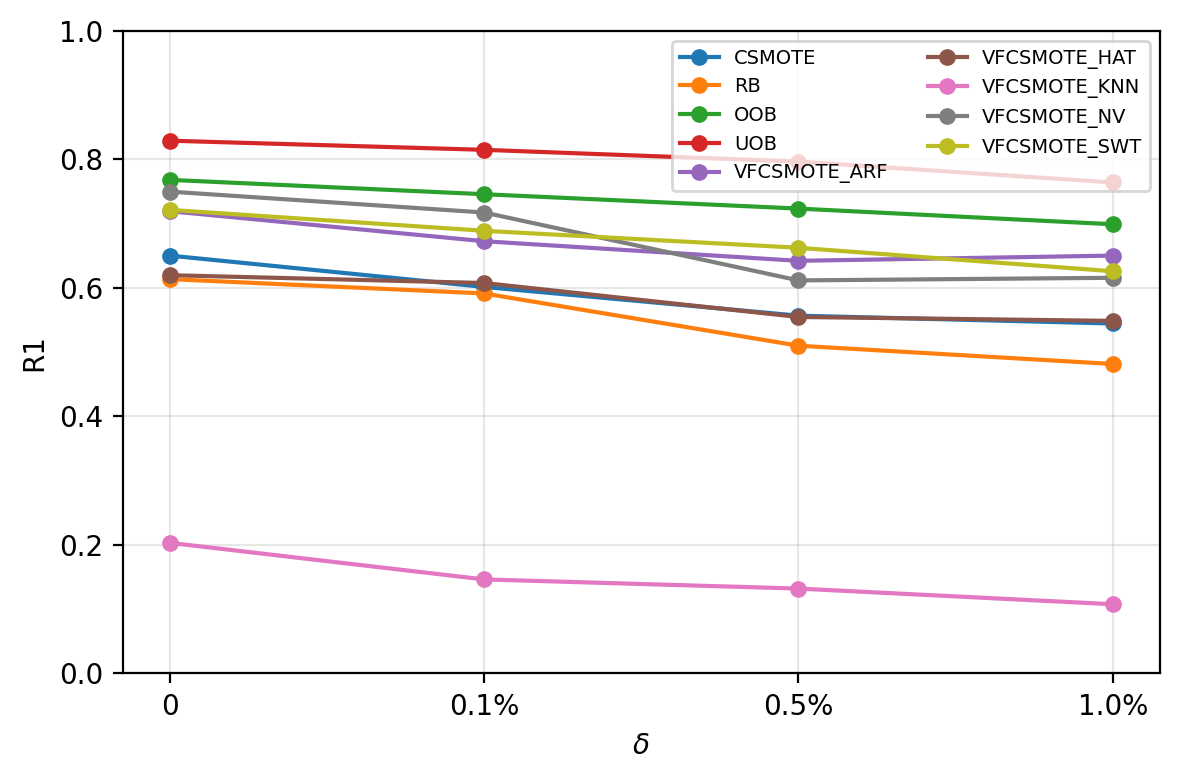}
    \caption{credit card}
    \label{fig:r1-creditcard}
  \end{subfigure}
  \hfill
  \begin{subfigure}[b]{0.32\textwidth}
    \includegraphics[width=\linewidth]{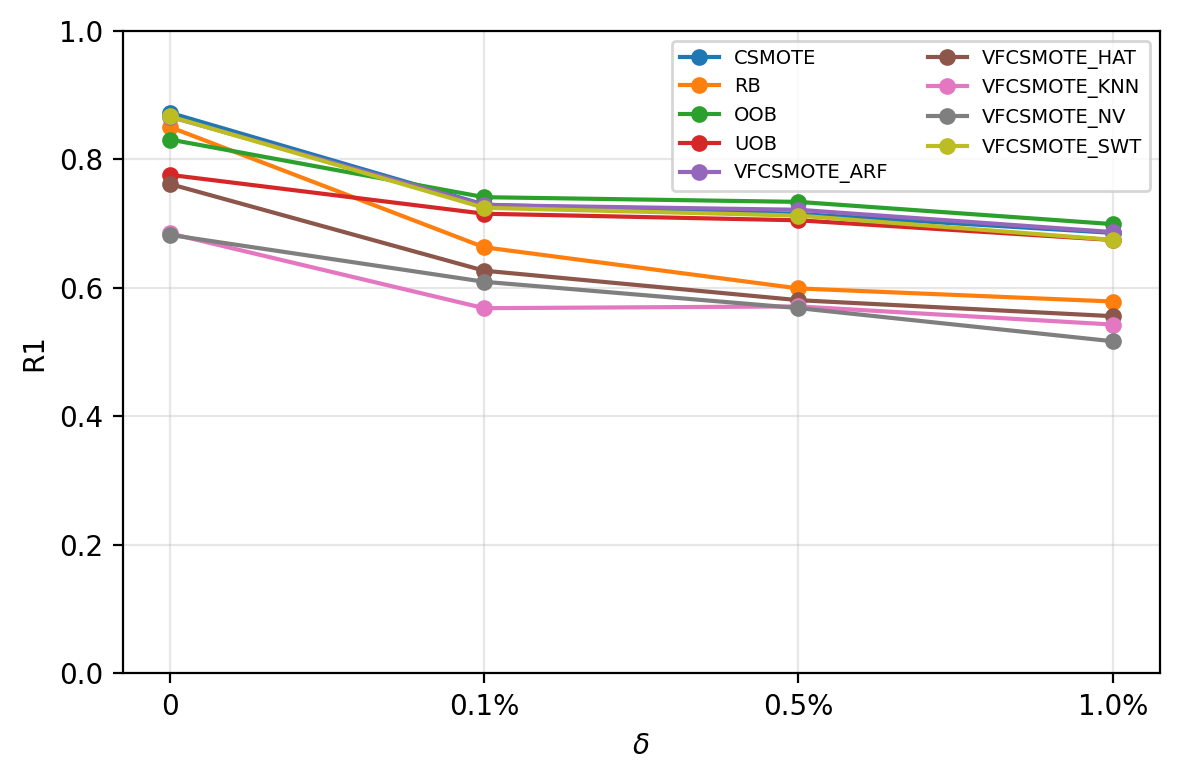}
    \caption{Electricity}
    \label{fig:r1-electricity}
  \end{subfigure}

  \medskip
  \begin{subfigure}[b]{0.32\textwidth}
    \includegraphics[width=\linewidth]{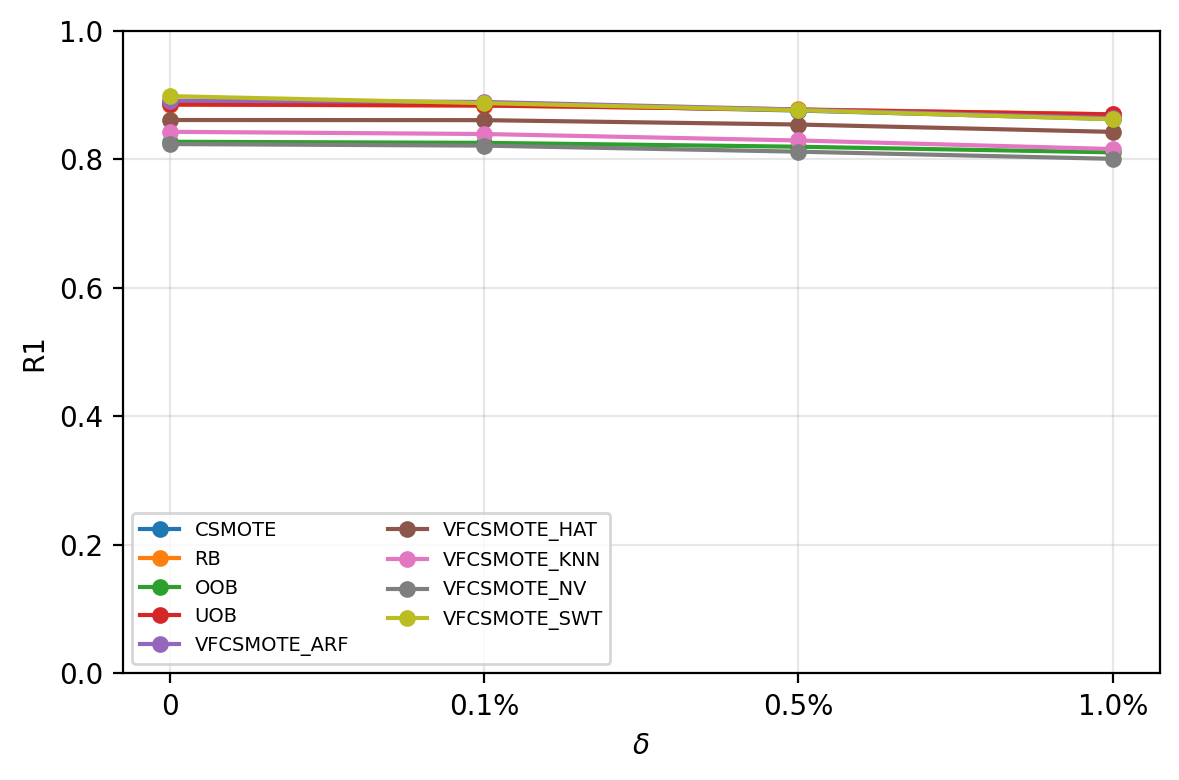}
    \caption{Insects}
    \label{fig:r1-insects}
  \end{subfigure}
  \hfill
  \begin{subfigure}[b]{0.32\textwidth}
    \includegraphics[width=\linewidth]{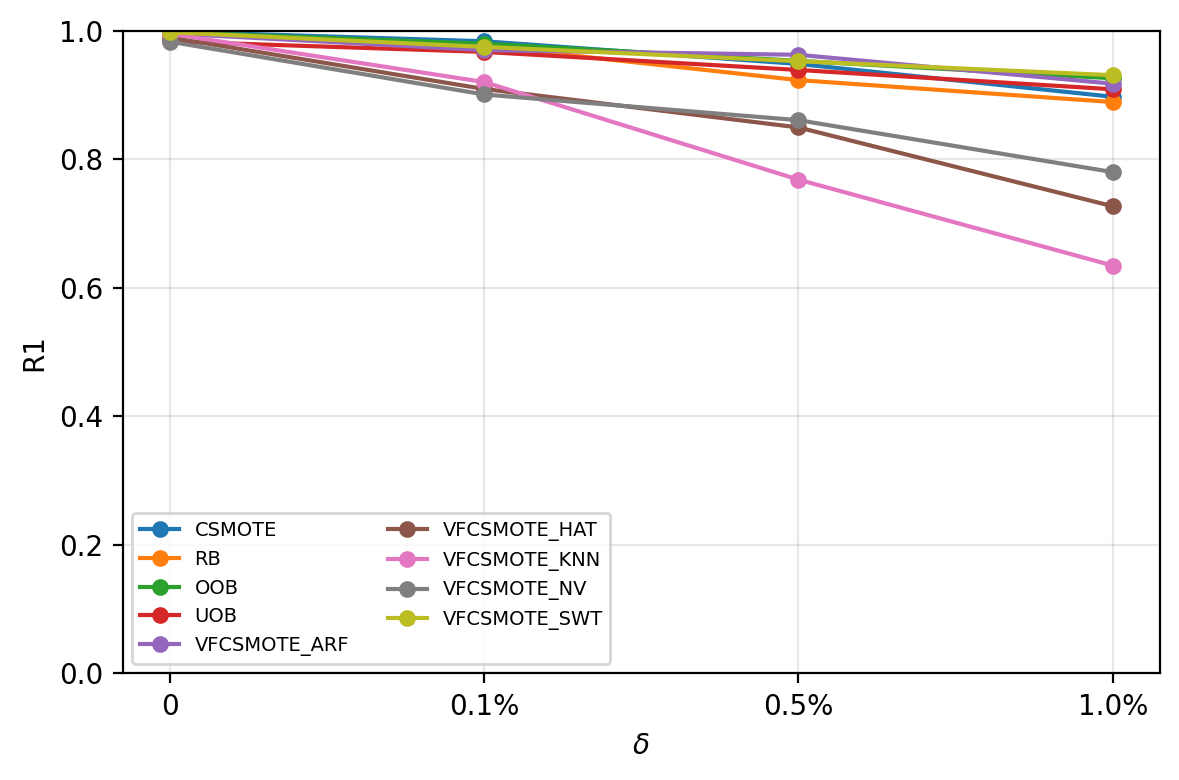}
    \caption{KDD Cup}
    \label{fig:r1-KDDCup}
  \end{subfigure}
  \hfill
  \begin{subfigure}[b]{0.32\textwidth}
    \includegraphics[width=\linewidth]{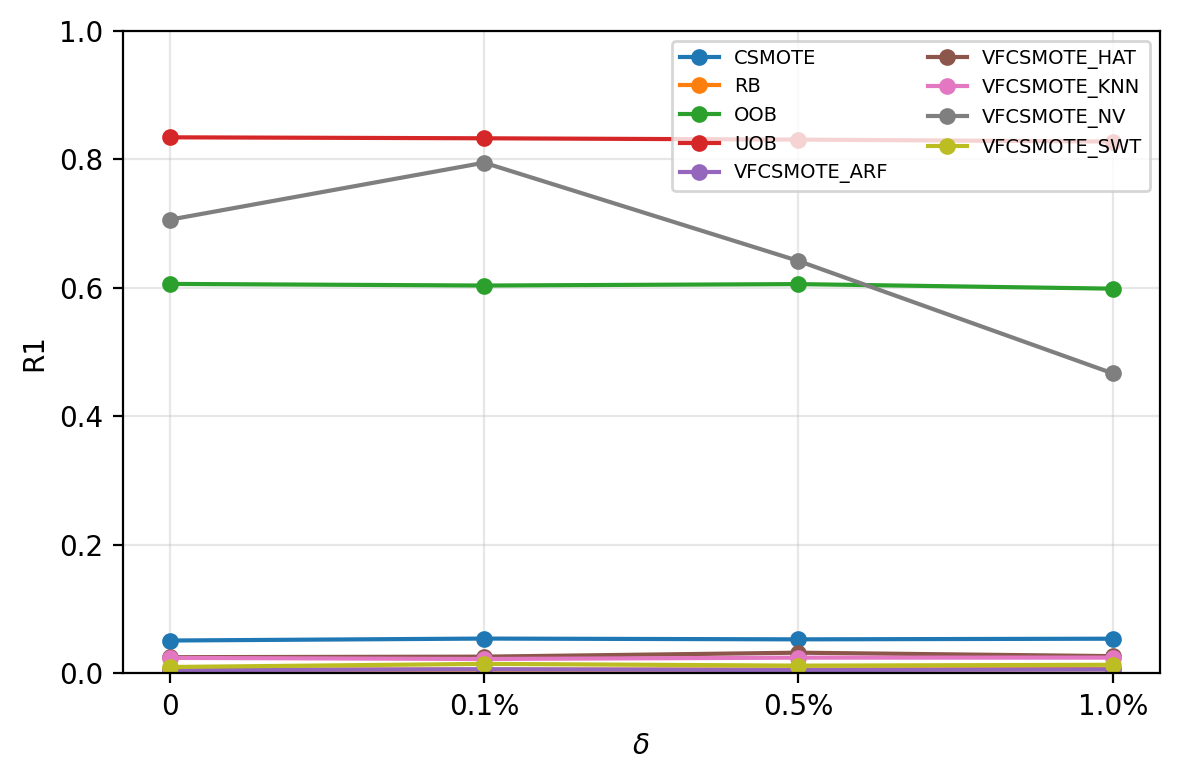}
    \caption{PAKDD}
    \label{fig:r1-packdd}
  \end{subfigure}

  \caption{Arrival $R1$ versus delay  $\delta$ for all oversampling
           methods across the six real data streams.}
  \label{fig:real-r1}
\end{figure}

Figures~\ref{fig:real-gmean} and~\ref{fig:real-r1} show the effect of delay on
arrival G-mean and $R1$ across the six real streams. Generally, whether a drift
exists, its nature, and its duration are not known beforehand. To characterise
each stream's intrinsic drift profile independently of the oversampling
methods, we ran a plain Hoeffding Tree (HT) probe with an external ADWIN detector at its default confidence of $0.002$, watching the probe's 0/1 loss stream. We deliberately use plain
HT rather than HAT, which contains an internal ADWIN that would reset on
detection and mask subsequent change points. The probe detects no drifts on
PAKDD and a single drift on CreditCard, both near stationary, with the CreditCard
count likely an imbalance artefact, against $45$ on
Covertype (sparse but high per-event impact), $64$ on KDD Cup (dense, low
per-event impact), $74$ on Insects-Abrupt-Imbalanced (dense, low boundary
impact), and $76$ on Electricity (dense, short-lag autocorrelated). We read
these counts alongside the per-dataset patterns in
Figures~\ref{fig:real-gmean} and~\ref{fig:real-r1}.

Two clusters emerge. PAKDD ($0$ detected drifts) and CreditCard ($1$ detected
drift, see caveat below) sit at the stationary end of the spectrum, and their
arrival G-mean curves are essentially flat across $\delta$ for every method.
Covertype, Electricity, and KDD Cup, in contrast, all degrade with delay, but
the magnitude tracks per-drift impact rather than count.
Covertype's $45$ detected drifts are sparse but severe in their boundary
effects, and the corresponding degradation is dramatic: several methods lose
more than half of their G-mean between $\delta=0$ and $\delta=1\%$.
KDD Cup's $64$ detected drifts are individually small, and most methods
retain $G \gtrsim 0.90$ at $\delta=1\%$; only \texttt{VFCSMOTE\_KNN} and
\texttt{VFCSMOTE\_HAT} degrade noticeably, a foreshadowing of the
synthetic--real transfer failure we analyse under RQ4.
Electricity ($76$ detected drifts) follows a third shape: a sharp drop between
$\delta=0$ and $\delta=0.1\%$ that then plateaus. This is consistent with the
short-lag autocorrelation of the Electricity stream documented by Bifet
et~al.~\cite{bifet2013pitfalls}, a $48$-instance cyclical demand structure
that even a small delay decorrelates, after which further delay adds little
incremental damage.

Insects-Abrupt-Imbalanced is the case that strengthens this reading rather
than complicating it. It has $74$ detected drifts, the second-highest
count in our suite, yet its arrival G-mean is essentially flat with
respect to $\delta$. This is the honest content of the count-versus-impact
distinction: ADWIN on a $0/1$ loss stream registers any shift in the
prediction-error process, including shifts that do not move the decision
boundary in a way that stale labels would amplify. Drift count, on its own,
is therefore an insufficient predictor of delay sensitivity; what matters is
the per-event contribution to boundary movement, which our diagnostic does
not directly measure but which Insects, Covertype, and KDD Cup together
demonstrate.


The $R1$ curves (Figure~\ref{fig:real-r1}) follow the same per-dataset
pattern but with sharper drops, as expected for the minority-class-sensitive
metric; we do not separately discuss them here.

\paragraph{RQ2}

\begin{figure}[t]
  \centering
  \begin{subfigure}[b]{0.32\textwidth}
    \includegraphics[width=\linewidth]{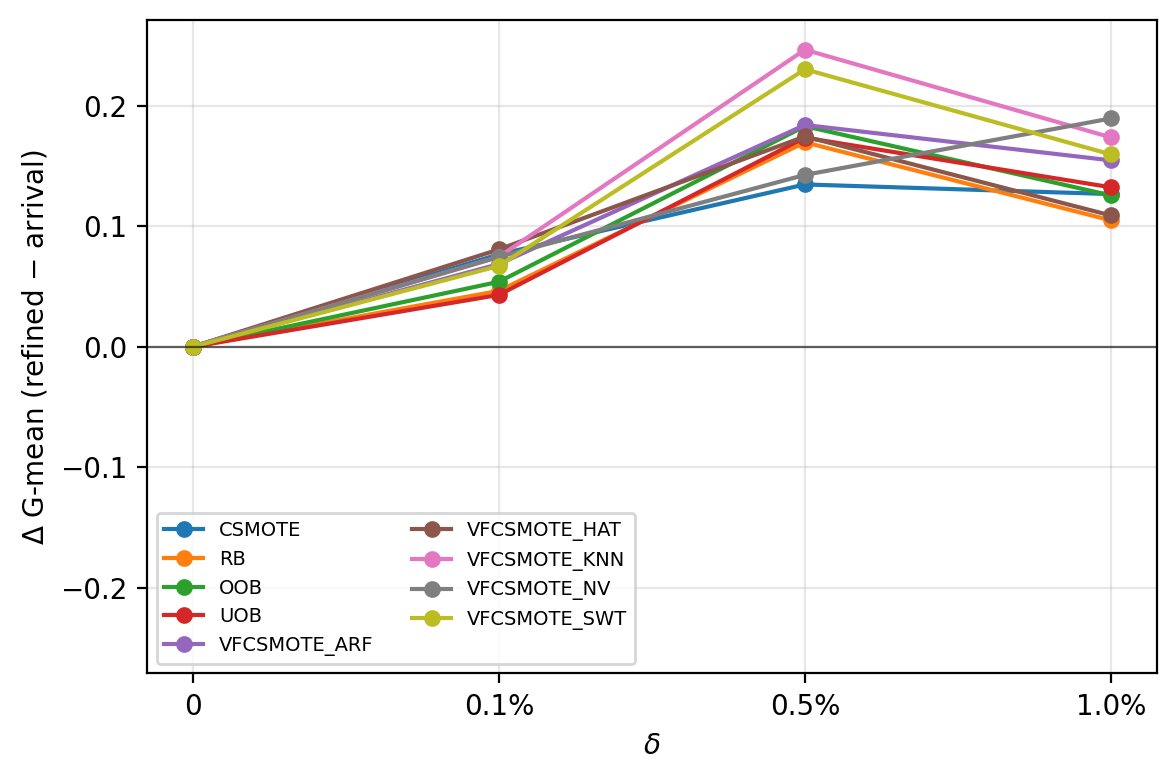}
    \caption{Cover type}
    \label{fig:gmean-covertype:delta}
  \end{subfigure}
  \hfill
  \begin{subfigure}[b]{0.32\textwidth}
    \includegraphics[width=\linewidth]{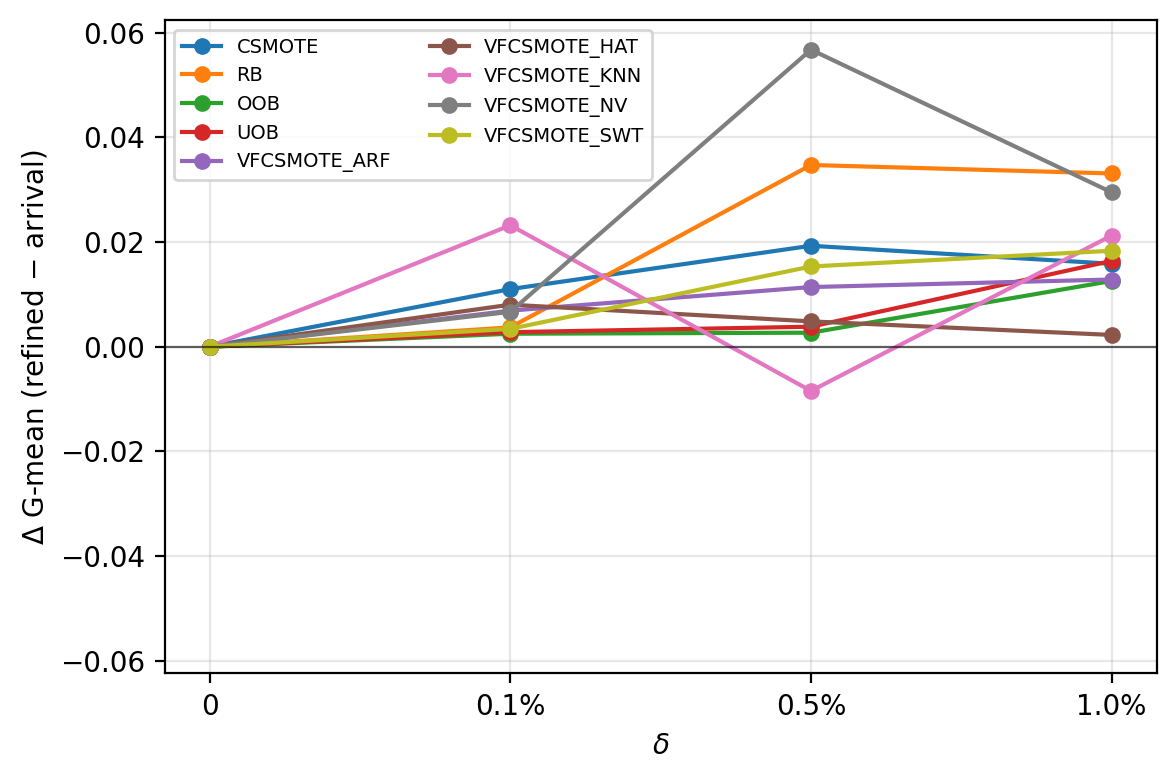}
    \caption{credit card}
    \label{fig:gmean-creditcard:delta}
  \end{subfigure}
  \hfill
  \begin{subfigure}[b]{0.32\textwidth}
    \includegraphics[width=\linewidth]{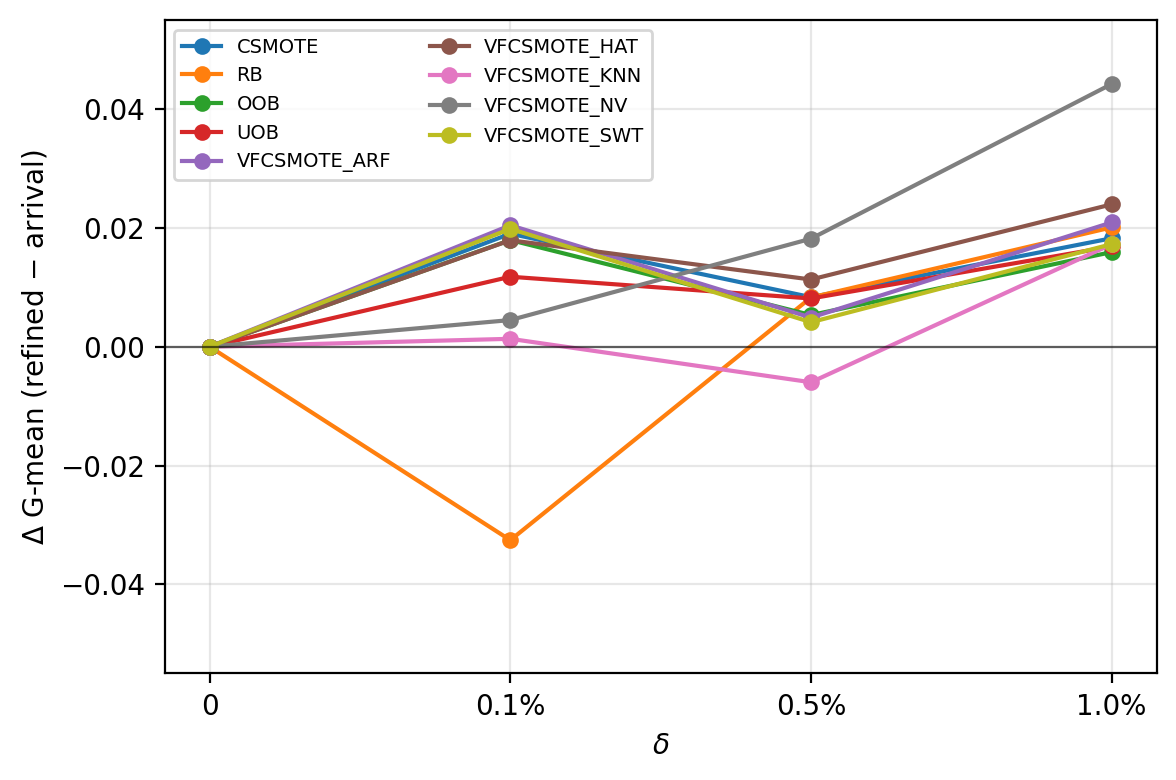}
    \caption{Electricity}
    \label{fig:gmean-electricity:delta}
  \end{subfigure}

  \medskip
  \begin{subfigure}[b]{0.32\textwidth}
    \includegraphics[width=\linewidth]{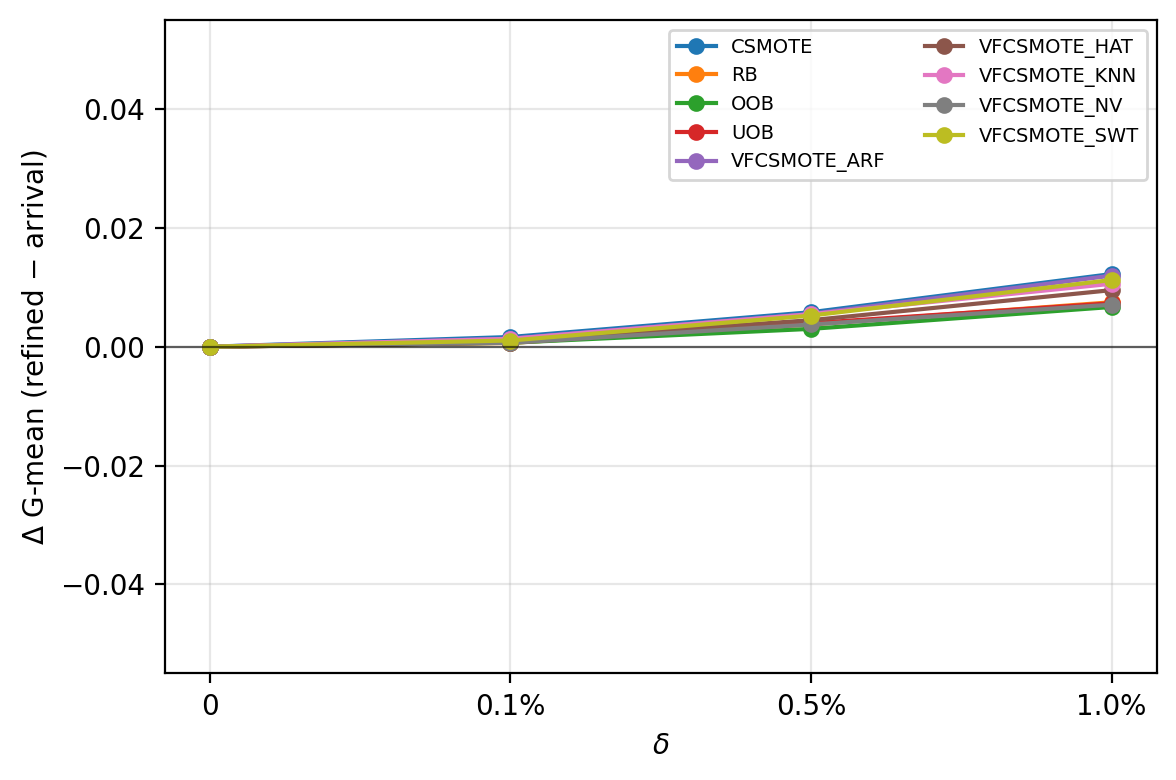}
    \caption{Insects}
    \label{fig:gmean-insects:delta}
  \end{subfigure}
  \hfill
  \begin{subfigure}[b]{0.32\textwidth}
    \includegraphics[width=\linewidth]{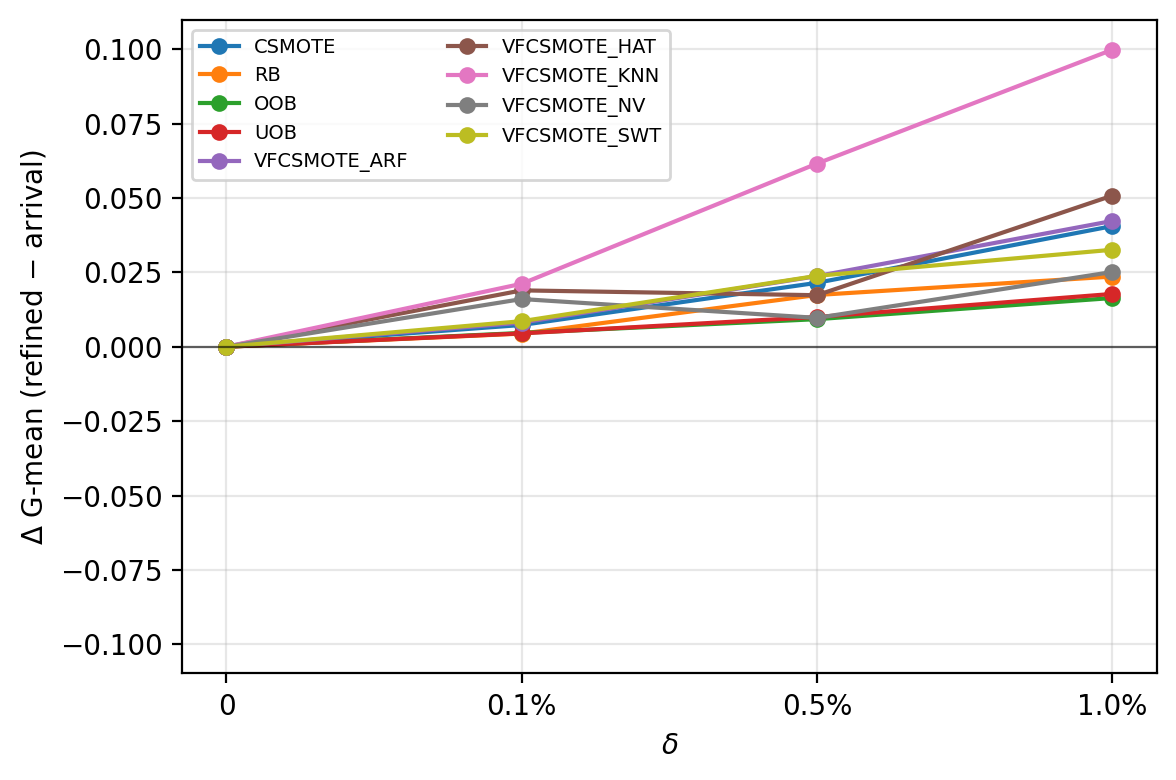}
    \caption{KDD Cup}
    \label{fig:gmean-KDDCup:delta}
  \end{subfigure}
  \hfill
  \begin{subfigure}[b]{0.32\textwidth}
    \includegraphics[width=\linewidth]{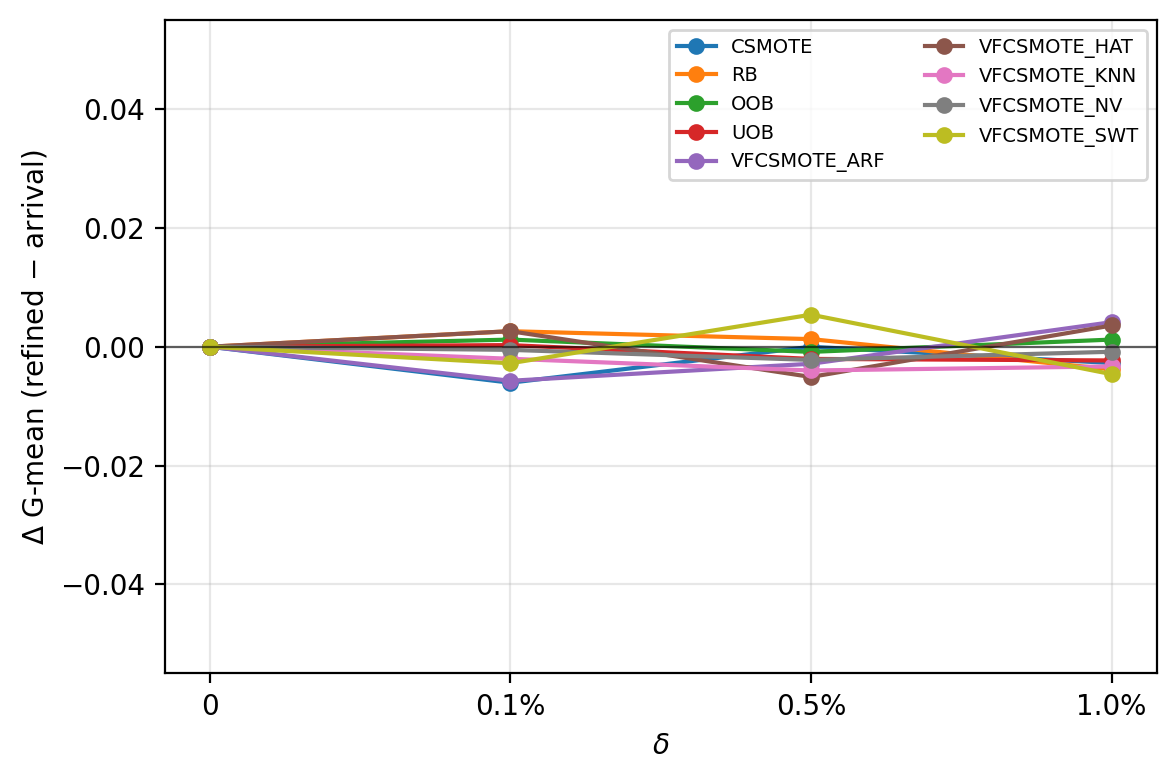}
    \caption{PAKDD}
    \label{fig:gmean-packdd:delta}
  \end{subfigure}

  \caption{$\Delta$ G-mean versus delay  $\delta$ for all oversampling
           methods across the six real data streams.}
  \label{fig:real-gmean:delta}
\end{figure}

Figure~\ref{fig:real-gmean:delta} shows $\Delta$ G-mean (refined $-$ arrival) across the six real streams. The synthetic finding under RQ2 was that re-evaluation produces a positive gain on every method, every stream, and every delay regime. The real data picture refines that conclusion: re-evaluation recovers what delay degraded when there is a recoverable degradation, and not otherwise.

On the drift-free streams, PAKDD and CreditCard, re-evaluation produces no systematic recovery. PAKDD is the clean case: $\Delta$ G-mean fluctuates within $\pm 0.01$ of zero for every method at every delay (Figure~\ref{fig:gmean-packdd:delta}). CreditCard is mostly near zero but exhibits a small number of method-specific excursions, \texttt{VFCSMOTE\_NV} reaches $\Delta \approx +0.055$ at $\delta=0.5\%$ and \texttt{VFCSMOTE\_KNN} swings between $+0.022$ and slightly negative across $\delta$, with no consistent direction or scaling with delay. Under the severe class imbalance of CreditCard, G-mean is dominated by a small number of minority-class calls, so a handful of correct re-predictions by an individual method can produce a visible $\Delta$ without indicating genuine delay-driven recovery. The insects dataset reproduces the PAKDD-style null result, $\Delta$ within $\pm 0.015$ across all methods and delays despite its high detected drift count, reinforcing the count-versus-impact reading from RQ1: drifts that do not move the boundary at arrival also do not create a recoverable loss for re-evaluation.

On the drift-active streams with substantial arrival-time loss, Covertype and KDD Cup, re-evaluation produces a clear positive recovery, and the magnitude scales with the magnitude of the arrival-time loss. Covertype, which suffered the most severe arrival degradation in Figure~\ref{fig:gmean-covertype}, shows the largest recoveries, with several methods exceeding $\Delta = +0.15$ at $\delta=0.5\%$ and \texttt{VFCSMOTE\_KNN} reaching $+0.247$. KDD Cup's recovery grows with $\delta$ as the arrival loss it is recovering from also grows. We read this as evidence that the synthetic finding, re-evaluation as a near-free recovery mechanism, is correct in mechanism but conditional in scope: it operates on losses attributable to delayed model updates, which dominate on drift-active streams and are absent on stationary ones.

Electricity is the partial exception. It produced substantial arrival-time loss (Figure~\ref{fig:gmean-electricity}), but the corresponding recoveries are modest in magnitude and uneven across methods. Most methods sit in the $+0.005$ to $+0.025$ band across delays, with one outlier climbing to $\Delta \approx +0.044$ at $\delta=1.0\%$ and \texttt{RB} exhibiting a meaningfully negative excursion ($\Delta \approx -0.033$ at $\delta=0.1\%$) before recovering at larger delays. \texttt{RB} pairs SWT as its base learner, and SWT, the temporally-augmented classifier of Bifet et~al.~\cite{bifet2013pitfalls}, consumes recent true class labels as input features; on Electricity's autocorrelated cyclical structure, a small delay decorrelates those appended labels enough to produce a worse re-prediction than the arrival prediction, which is consistent with the \texttt{RB} excursion at $\delta=0.1\%$. Despite Electricity having a larger detected drift count than Covertype or KDD Cup, its re-evaluation gains are smaller than Covertype's at every delay and smaller than KDD Cup's at $\delta \geq 0.5\%$; the arrival-time loss is real but is recovered far less efficiently. We read this as evidence that the mechanism behind a stream's arrival loss matters as much as its magnitude, detected drift count alone does not predict how much of the loss re-evaluation can claw back.

Finally, the appearance of a negative $\Delta$ G-mean at several stream$\times$method points (Figures~\ref{fig:gmean-creditcard:delta},~\ref{fig:gmean-electricity:delta}, and~\ref{fig:gmean-packdd:delta}) is itself a real-data finding distinct from the synthetic result. Re-evaluation in the real-data regime can occasionally shift a correct arrival prediction to an incorrect one without a compensating benefit from later training; the strict universality of the synthetic finding does not carry over to real streams.

\begin{figure}[htb!]
  \centering
  \begin{subfigure}[b]{0.32\textwidth}
    \includegraphics[width=\linewidth]{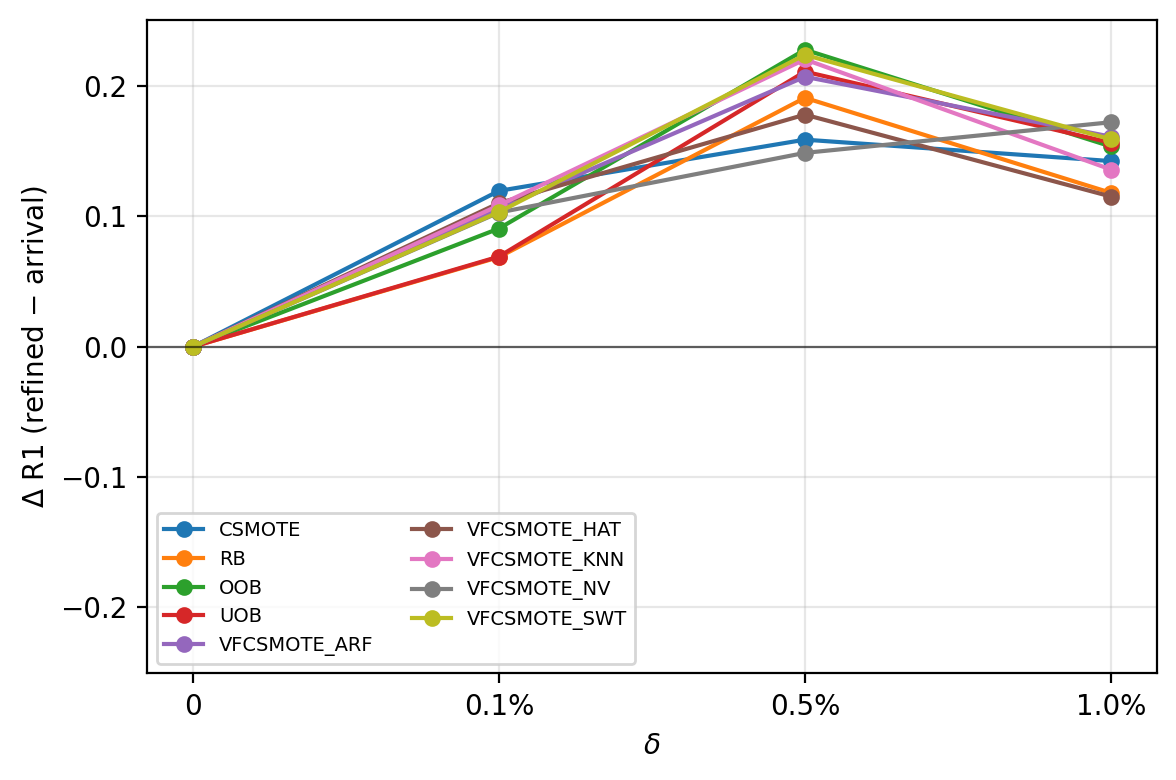}
    \caption{Cover type}
    \label{fig:r1-covertype:delta}
  \end{subfigure}
  \hfill
  \begin{subfigure}[b]{0.32\textwidth}
    \includegraphics[width=\linewidth]{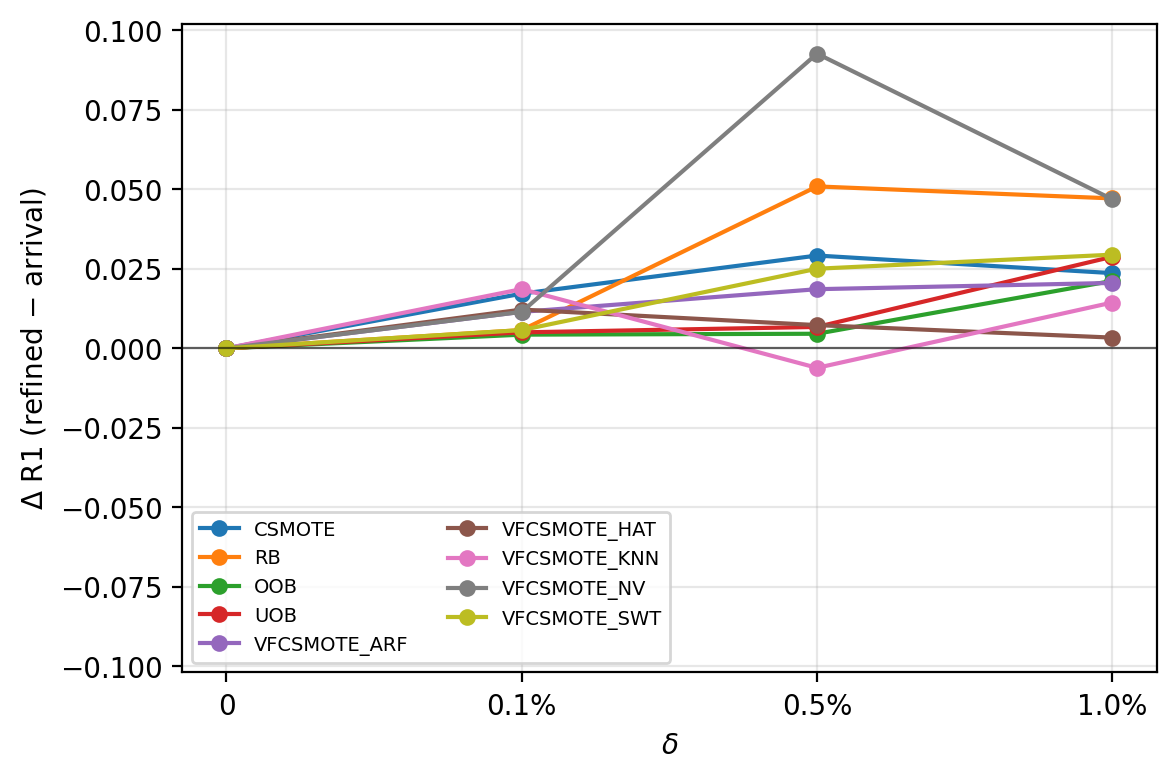}
    \caption{credit card}
    \label{fig:r1-creditcard:delta}
  \end{subfigure}
  \hfill
  \begin{subfigure}[b]{0.32\textwidth}
    \includegraphics[width=\linewidth]{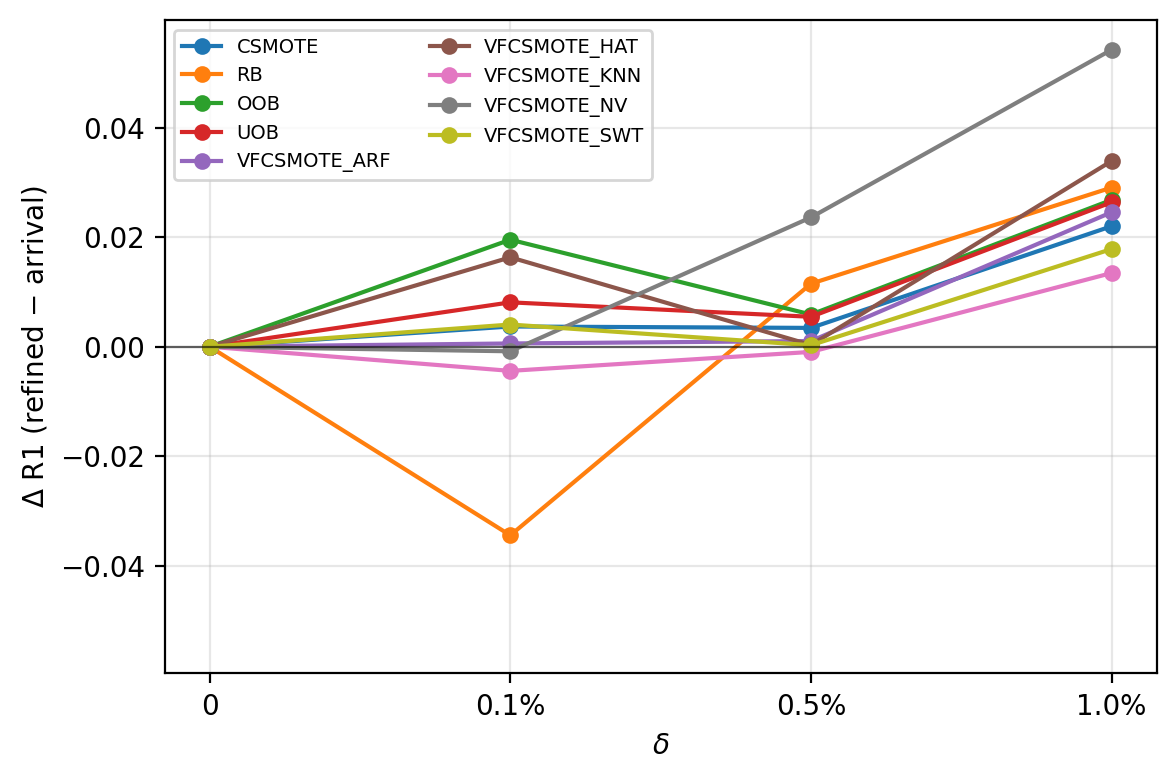}
    \caption{Electricity}
    \label{fig:r1-electricity:delta}
  \end{subfigure}

  \medskip
  \begin{subfigure}[b]{0.32\textwidth}
    \includegraphics[width=\linewidth]{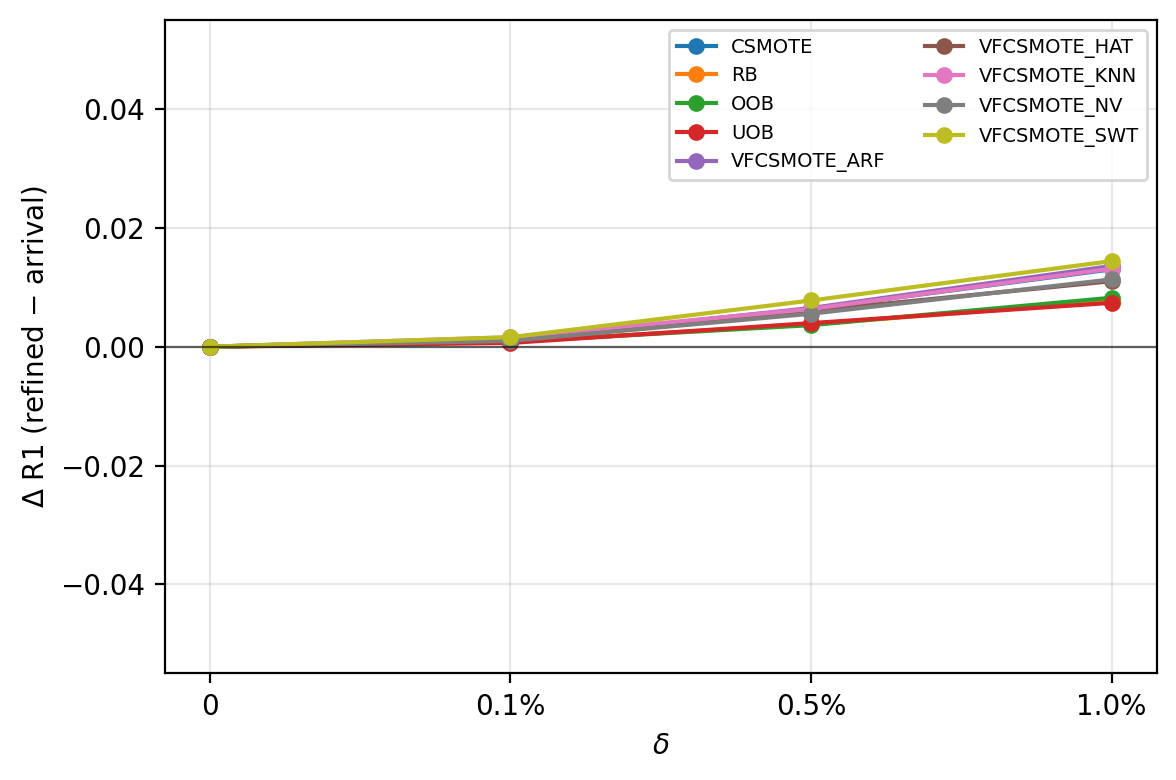}
    \caption{Insects}
    \label{fig:r1-insects:delta}
  \end{subfigure}
  \hfill
  \begin{subfigure}[b]{0.32\textwidth}
    \includegraphics[width=\linewidth]{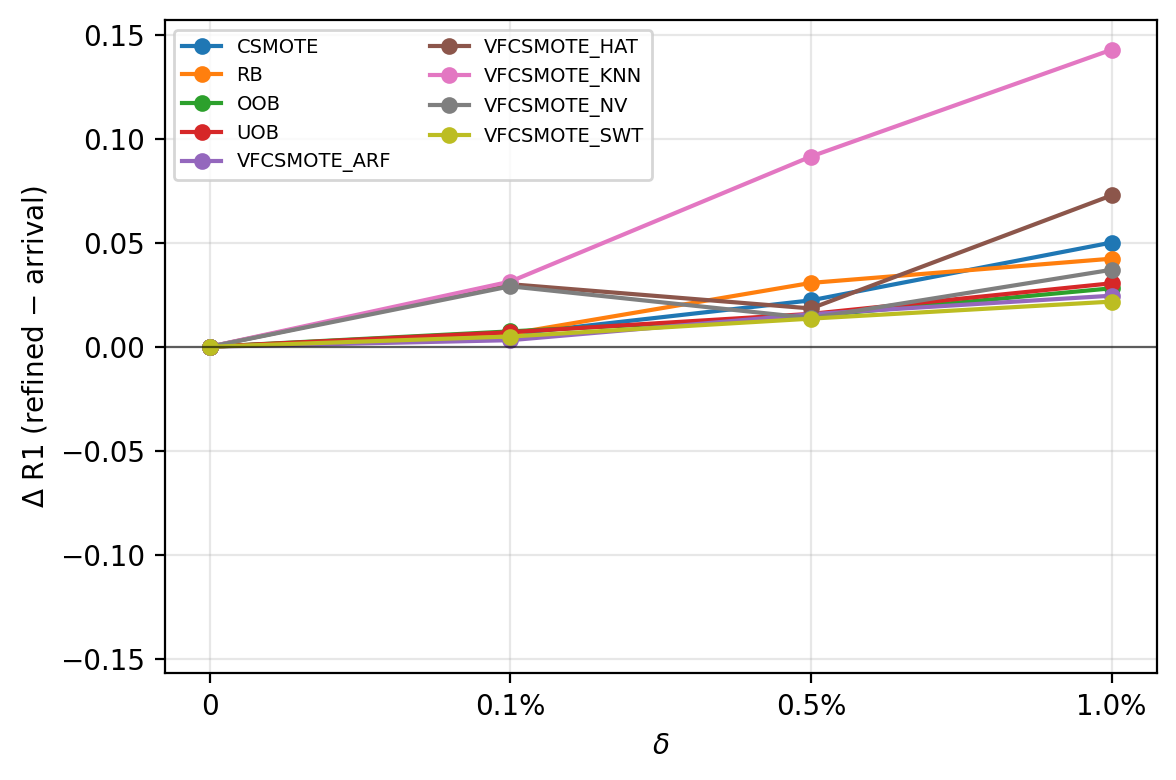}
    \caption{KDD Cup}
    \label{fig:r1-KDDCup:delta}
  \end{subfigure}
  \hfill
  \begin{subfigure}[b]{0.32\textwidth}
    \includegraphics[width=\linewidth]{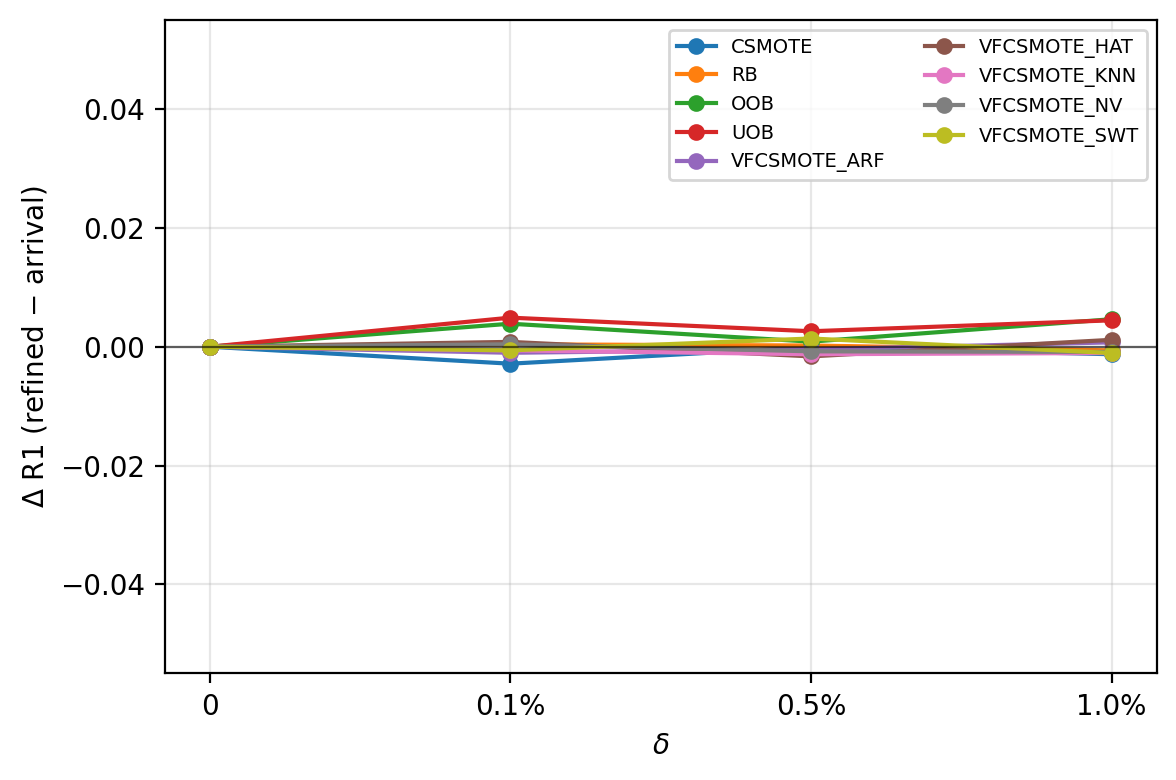}
    \caption{PAKDD}
    \label{fig:r1-packdd:delta}
  \end{subfigure}

  \caption{$\Delta$ $R1$ versus delay  $\delta$ for all oversampling
           methods across the six real data streams.}
  \label{fig:real-r1:delta}
\end{figure}

\clearpage

\paragraph{RQ3}

Following the conditional rule stated in Section~\ref{sec:exp:rqs}, we apply the DiD test only to configurations in which re-evaluation produced a positive average gain across the six real datasets ($\overline{\Delta_G^{\text{oversampler}}} > 0$). For each such configuration, we run the paired Wilcoxon on
$\Delta_G^{\text{oversampler}} - \Delta_G^{\text{bare}}$ across the six datasets, with significance assessed at the $n=6$ floor of $p=0.031$.

\begin{table}[ht]
\centering
\caption{Wilcoxon paired-test results for oversampler vs.\ bare-learner DiD on
the real datasets. Cells are pooled per pair over the $6 \times 3 = 18$ (dataset,
$\delta$) combinations; a cell is eligible iff $\Delta_G^{\mathrm{oversampler}} > 0$.
$+/-/=$ counts eligible cells with positive, negative, and zero DiD. The only
significant pair ($\alpha=0.05$) is shown in bold.}
\label{tab:DID:real}
\small
\setlength{\tabcolsep}{4pt}
\begin{tabular}{llcllcc}
\toprule
Oversampler & Bare learner & $n$ & Excluded cells ($\Delta_{\text{over}} \le 0$) & $+/-/=$ & mean DiD & $p$ \\
\midrule
VFCSMOTE & ARF    & 16 & PAKDD@\{0.1, 0.5\}                       & 8/8/0  & $+0.0013$        & 0.918           \\
CSMOTE        & ARF    & 15 & PAKDD@\{0.1, 0.5, 1.0\}                  & 8/7/0  & $+0.0022$        & 0.629           \\
\midrule
VFCSMOTE & HAT    & 17 & PAKDD@\{0.5\}                            & 4/13/0 & $-0.0176$        & 0.207           \\
\midrule
VFCSMOTE  & NV     & 15 & PAKDD@\{0.1, 0.5, 1.0\}                  & 15/0/0 & $\mathbf{+0.0419}$ & $\mathbf{0.0001}$ \\
\midrule
VFCSMOTE & KNN    & 14 & \makecell[l]{Elec@\{0.5\}, \\ PAKDD@\{0.1, 0.5, 1.0\}}    & 4/10/0 & $-0.0163$        & 0.135           \\
\midrule
VFCSMOTE & SWT    & 16 & PAKDD@\{0.1, 1.0\}                       & 9/7/0  & $+0.0062$        & 0.214           \\
RB            & SWT    & 16 & \makecell[l]{Elec@\{0.1\}, \\ PAKDD@\{1.0\}}              & 7/9/0  & $-0.0110$        & 0.376           \\
\midrule
OOB           & OzaBag & 17 & PAKDD@\{0.5\}                            & 10/5/2 & $+0.0072$        & 0.118           \\
UOB           & OzaBag & 16 & PAKDD@\{0.5, 1.0\}                       & 10/5/1 & $+0.0014$        & 0.173           \\
\bottomrule
\end{tabular}
\end{table}
The Wilcoxon paired test on the eligible DiD set (Table~\ref{tab:DID:real}) returns a single significant result:  VFCSMOTE\_NV vs.\ NV, with positive DiD on every one of its $15$ eligible cells ($p = 0.0001$, mean DiD $+0.042$). The remaining eight pairs are non-significant. Two trend negative (VFCSMOTE\_HAT vs.\ HAT, VFCSMOTE\_KNN vs.\ KNN; $p \in [0.13, 0.21]$), three
trend mildly positive (OOB, UOB, VFCSMOTE\_SWT; $p \in [0.12, 0.22]$), and
the rest are null. 

The VFCSMOTE\_NV result is mechanistically interpretable along three axes. \textit{Minority-class learnability}:  bare NV's posterior is dominated by the class prior when the minority is rare or weakly separable, starving its class-conditional estimates; synthetic minority instances directly enrich them. \textit{Drift}: NV has no built-in drift adaptation, so bare NV's slow incremental updates leave
substantial recovery headroom that the oversampler fills. \textit{Delay}: as $\delta$ increases, each re-evaluation tick releases more synthetic minority training signal, so the gap between VFCSMOTE\_NV's and bare NV's recovery gain widens monotonically with delay (median DiD rises from $+0.006$ at $\delta{=}0.1\%$ to $+0.040$ at $\delta{=}0.5\%$ to $+0.048$ at $\delta{=}1.0\%$). All three axes align in the oversampler's favour, and the unanimous 15/0 sign agreement reflects that alignment rather than any exceptional absolute performance of the configuration.

The negative trends for VFCSMOTE\_HAT and VFCSMOTE\_KNN follow the same three axes operating in reverse. Bare HAT and KNN are already minority-aware: KNN's local decision rule is intrinsically responsive to minority clusters, and HAT's Hoeffding splits track conditional class probabilities directly. Both are also drift-reactive on their own — HAT through internal ADWIN-driven splits and KNN through its sliding window. When the bare learner exploits delay-released labels efficiently, its $\Delta_G$ becomes large, and the oversampler's smaller $\Delta_G$ produces a negative DiD by arithmetic, even when the oversampler itself improves with delay in absolute terms. We refer to this as a \emph{re-evaluation headroom collapse}: oversampling raises the arrival-time level, leaving less room for re-evaluation to recover, so the bare learner accrues the larger refinement gain. The same mechanism explains why the VFCSMOTE variants over already-strong bases (ARF, SWT) produce near-null DiD: when the bare learner is both minority-aware and drift-reactive, oversampling adds neither at  arrival nor at refinement.

Taken together, RQ3 identifies a narrow regime in which oversampling demonstrably amplifies re-evaluation: the bare learner must be structurally weak on minority-class structure \emph{and} lack intrinsic drift adaptation, so that synthetic minority instances that accumulate over re-evaluation cycles can fill the recovery headroom. This regime is occupied uniquely by NV within our suite. Where the bare learner satisfies even one of these conditions on its own, the marginal contribution of oversampling falls below detection; where it satisfies both, the contribution becomes negative through headroom collapse, in which oversampling and re-evaluation compete for the same delay-induced recovery budget.

\paragraph{RQ4}

The synthetic Pareto analysis (Section~\ref{sec:results:synthetic}) aggregated
the $n{=}6$ streams within each delay-to-drift ratio $\rho$. Those six
streams span two drift types (SEA abrupt and SEA$_g$ gradual) and several
imbalance levels, but they share the same $\rho$ regime by construction and are stratified by design across the drift-type and imbalance dimensions.

The six real datasets cannot be aggregated this way. They differ in drift count by two orders of magnitude (see the discussion under \textbf{RQ1} earlier in this section), in drift impact, in minority frequency, and in absolute recovery dynamics, and their heterogeneity is uncontrolled rather than stratified. The delay-to-drift regime $\rho$ that defined the synthetic aggregation unit is not even well-defined uniformly across the real datasets, since the implied drift cadence ranges from zero (PAKDD, CreditCard) to dense (Elec at 3.7 drifts per 1000 instances). Aggregating them into a single Pareto would conflate fundamentally different streaming regimes and produce a front that is optimal on no individual dataset. We therefore present the Pareto analysis per dataset, allowing the front to reflect the drift profile of each stream.

\begin{table}[ht]
\centering


\caption{Per-dataset Pareto status (refined G-mean vs.\ overhead replayed):
Pareto-optimal learners plus the best dominated one. Blocks are $\delta$-invariant except Elec and Insects$@\delta{=}1.0\%$. Full means in the repository.}
\label{tab:pareto:real}
\small
\begin{tabular}{lrcc}
\toprule
Learner & Overhead & G-mean & Pareto \\
\midrule
\multicolumn{4}{c}{Covertype (all $\delta$; Pareto-optimal: 2/15)} \\
\midrule
SWT           &          0 & 0.893 & Y \\
OOB           &  5{,}884{,}187 & 0.916 & Y \\
ARF           &          0 & 0.890 & N \\
\midrule
\multicolumn{4}{c}{CreditCard (all $\delta$; 2/15)} \\
\midrule
NV            &          0 & 0.897 & Y \\
UOB           &     71{,}974 & 0.902 & Y \\
OOB           &  2{,}883{,}332 & 0.868 & N \\
\midrule
\multicolumn{4}{c}{Elec, $\delta=0.1\%$ (1/15)} \\
\midrule
SWT           &          0 & 0.804 & Y \\
ARF           &          0 & 0.801 & N \\
\midrule
\multicolumn{4}{c}{Elec, $\delta\ge0.5\%$ (1/15)} \\
\midrule
ARF           &          0 & 0.786 & Y \\
SWT           &          0 & 0.785 & N \\
\midrule
\multicolumn{4}{c}{Insects (all $\delta$; 2/15, SWT joins at $\delta=1.0\%$)} \\
\midrule
ARF           &          0 & 0.875 & Y \\
VFCSMOTE\_ARF &    468{,}609 & 0.879 & Y \\
SWT$^{\ast}$  &          0 & 0.875 & Y \\
CSMOTE        &    490{,}871 & 0.878 & N \\
\midrule
\multicolumn{4}{c}{KDDCup (all $\delta$; 3/15)} \\
\midrule
ARF           &          0 & 0.996 & Y \\
OzaBag        &          0 & 0.996 & Y \\
SWT           &          0 & 0.996 & Y \\
CSMOTE        &      1{,}255 & 0.996 & N \\
\midrule
\multicolumn{4}{c}{PAKDD (all $\delta$; 2/15)} \\
\midrule
NV            &          0 & 0.575 & Y \\
OOB           &    790{,}180 & 0.608 & Y \\
VFCSMOTE\_NV  &     60{,}138 & 0.524 & N \\
\bottomrule
\multicolumn{4}{l}{\footnotesize $^{\ast}$ SWT is Pareto-optimal only at $\delta=1.0\%$.} \\
\end{tabular}
\end{table}

\textbf{RQ4} on real data inherits the marginal-versus-absolute distinction from the synthetic analysis, but the clearest illustration runs in reverse. The exemplar on synthetic data was \texttt{CSMOTE}: null $\mathrm{DiD}$ (\textbf{RQ3}) yet Pareto-optimal at every $\rho$ (\textbf{RQ4}). On real data the inverse holds: \texttt{VFCSMOTE\_NV} shows a significant positive $\mathrm{DiD}$ against \texttt{NV} (Table~\ref{tab:DID:real}) yet is absent from every Pareto front, dominated by bare \texttt{NV} on the two stationary streams (PAKDD, CreditCard) and by stronger learners elsewhere. Significant marginal contribution does not translate to Pareto-optimal deployment when the oversampler's gain over the bare base does not lift it above the absolute level of competing learners.

Table~\ref{tab:pareto:real} reveals a partition aligned with the drift, delay, and minority axes. In regimes where drift density is
high and bare-learner drift adaptation is strong (Elec under
\texttt{SWT}/\texttt{ARF}), the bare-learner recovery channel absorbs most of the delay-induced signal at zero overhead, and no oversampler can improve refined G-mean enough to justify positive replay cost. Symmetrically, in regimes where refined G-mean is near-saturated regardless of oversampling (KDDCup at $\approx 0.996$ across multiple bare learners), there is no headroom for an oversampler to climb. In both regimes, the deployment recommendation collapses to a bare learner at zero overhead. In intermediate regimes, moderate drift density, non-saturated refined G-mean, and non-trivial minority imbalance, exactly one oversampler joins one bare learner on the front, but the cost-effectiveness of the oversampler varies by two orders of magnitude across these cells. The Pareto-optimal oversamplers in this table span $0.25\times$ to $15.8\times$ the original stream size in synthesis overhead, in exchange for refined G-mean gains of $+0.004$ to $+0.032$ over the best bare learner. Whether a given oversampler is worth deploying, therefore, depends on the deployment context's tolerance 
for replay overhead, not on the existence of a positive marginal effect 
alone.

The three candidates carried forward from the synthetic analysis
(\texttt{UOB},\\ \texttt{VFCSMOTE\_ARF}, \texttt{VFCSMOTE\_KNN}) deliver
markedly different outcomes. \texttt{UOB}, the synthetic top-line and
Pareto-optimal at every $\rho$, enters the real-data Pareto only on
CreditCard, and is dominated on the other five datasets. \texttt{VFCSMOTE\_ARF}, the most consistent synthesis-based candidate on synthetic, enters the real-data Pareto only on Insects, and by a $+0.004$ margin over bare \texttt{ARF}. \texttt{VFCSMOTE\_KNN}, carried forward as the strongest synthetic case for synthesis-based oversampling, is dominated on every real dataset; the synthetic sweet spot does not survive.

Two opposite-direction reversals are worth flagging. \texttt{OOB}, which was dominated by \texttt{UOB} on every synthetic $\rho$, enters the real-data Pareto on two datasets with qualitatively different profiles: one with high-impact drift (Covertype), and one with low minority-class learnability despite moderate imbalance (PAKDD), a dataset where only aggressive resampling-based methods such as \texttt{OOB} and \texttt{UOB} reach appreciable minority recall (see figures~\ref{fig:r1-packdd} and~\ref{fig:r1-covertype}), while standard bare learners default to the majority prior. The base-bagging machinery that was structurally inferior on controlled synthetic streams becomes a viable oversampling choice either when drift is severe enough to widen the recovery budget or when the minority is intrinsically hard to learn from labels alone. This pattern is corroborated by the per-pair Wilcoxon DiD in Table~\ref{tab:DID:real}: both \texttt{OOB} vs.\ \texttt{OzaBag} and \texttt{UOB} vs.\ \texttt{OzaBag} trend positive on refined G-mean ($+0.0072$ and $+0.0014$; 10/5 positive splits each; $p>0.1$). The trend is non-significant because pooling spans datasets on which bare learners already capture the available minority signal, but the directional consistency, under imbalance regimes that hold R0\footnote{Figures for $R0$, $F1_0$, $F1_1$, and accuracy can be found in our GitHub repository} near unity, attributes the lift to the synthetic minority-coverage channel rather than to any interaction with re-evaluation. \texttt{CSMOTE}, the lightest synthetic Pareto winner and the cleanest synthetic illustration of the marginal-versus-absolute distinction, is dominated on every real dataset; its synthetic level effect does not survive against drift-reactive bare learners on real streams. \texttt{VFCSMOTE\_HAT} and \texttt{VFCSMOTE\_SWT} are also dominated everywhere, consistent with the headroom-collapse mechanism described in Section~\ref{sec:discussion}: when the bare learner is both minority-aware and drift-reactive, oversampling adds neither at arrival nor at refinement.


\clearpage
\section{Discussion}
\label{sec:discussion}

The DiD significance of oversampling under re-evaluation depends on the relative width of two competing recovery channels: the bare learner's per-label drift- and minority-correction channel, and the oversampler's synthetic minority-coverage channel. The former widens with the bare learner's intrinsic drift adaptation (NV $<$ OzaBag $<$ HAT, KNN, SWT, ARF) and with the stream's drift density; the latter widens with the difficulty the bare learner has in learning the minority class from labels alone, which jointly reflects how rare the minority is and how well the features separate it. A significant positive DiD requires the oversampler channel to dominate; a null or negative DiD occurs when the bare channel is sufficiently wide to absorb most of the recovery signal on its own.

On the synthetic SEA streams, drift density is fixed by the generator at a moderate level, minority imbalance varies between the streams within each $\rho$ level, and $\rho$ itself, the ratio of delay to inter-drift interval, varies across $\{0.25, 1.0, 2.0\}$. Out of 27 (pair, $\rho$) cells, only six cleared significance: OOB vs.\ OzaBag at every $\rho$, and VFCSMOTE\_ARF, VFCSMOTE\_KNN, and UOB at $\rho{=}2.0$ alone . The pattern reflects the same channel competition: OOB's aggressive minority resampling produces an oversampler contribution wide enough to dominate at any $\rho$, while the other three pairs require $\rho{=}2.0$, the regime where the bare learner falls furthest behind drift between re-evaluation ticks, before the oversampler channel exceeds their bare learners' intrinsic drift adaptation. 

On the real-data eligible set, drift density is one to two orders of magnitude higher on the drift-active datasets (Elec, Insects, KDDCup, Covertype) and minority-class learnability is heterogeneous; the bare-learner channel for drift-reactive learners (HAT, KNN, ARF, SWT) is correspondingly wider, and the synthetic significances do not reproduce. VFCSMOTE\_NV emerges as the only significant pair on real data because bare NV has neither drift adaptation nor a fast minority update mechanism, leaving the entire recovery budget for the oversampler to exploit regardless of the (drift density, minority-class learnability) combination on a given cell.

\begin{figure}[ht!]
\centering
\begin{tikzpicture}[
  font=\small, node distance=10mm and 18mm,
  start/.style={rounded corners, draw, fill=gray!10, align=center, inner sep=2mm},
  decision/.style={diamond, aspect=2.4, draw, align=center, inner sep=0.5mm},
  action/.style={rectangle, draw, align=center, inner sep=2mm, text width=36mm},
  arr/.style={-{Latex}, thick}
]
\node[start] (start) {Data stream};
\node[decision, below=of start] (learn) {Minority-class\\learnability\\binding?};
\node[action, right=of learn] (resample) {Aggressive class-balanced resampling (\texttt{OOB}, \texttt{UOB})\\{\footnotesize coverage channel binds}};
\node[decision, below=of learn] (drift) {High drift\\density?};
\node[action, right=of drift] (adaptive) {Drift-adaptive bare learner (\texttt{HAT}, \texttt{KNN}, \texttt{SWT}, \texttt{ARF})\\{\footnotesize recovery channel wide}};
\node[action, below=of drift] (nv) {Lightweight bare learner (\texttt{NV})\\{\footnotesize neither channel active}};
\draw[arr] (start) -- (learn);
\draw[arr] (learn) -- node[above]{yes} (resample);
\draw[arr] (learn) -- node[right]{no} (drift);
\draw[arr] (drift) -- node[above]{yes} (adaptive);
\draw[arr] (drift) -- node[right]{no} (nv);
\end{tikzpicture}

\caption{Decision rule for method selection by stream profile, at the evaluated settings ($K=100$, label delay $\delta \leq 1.0\%$).}
\label{fig:decision}
\end{figure}
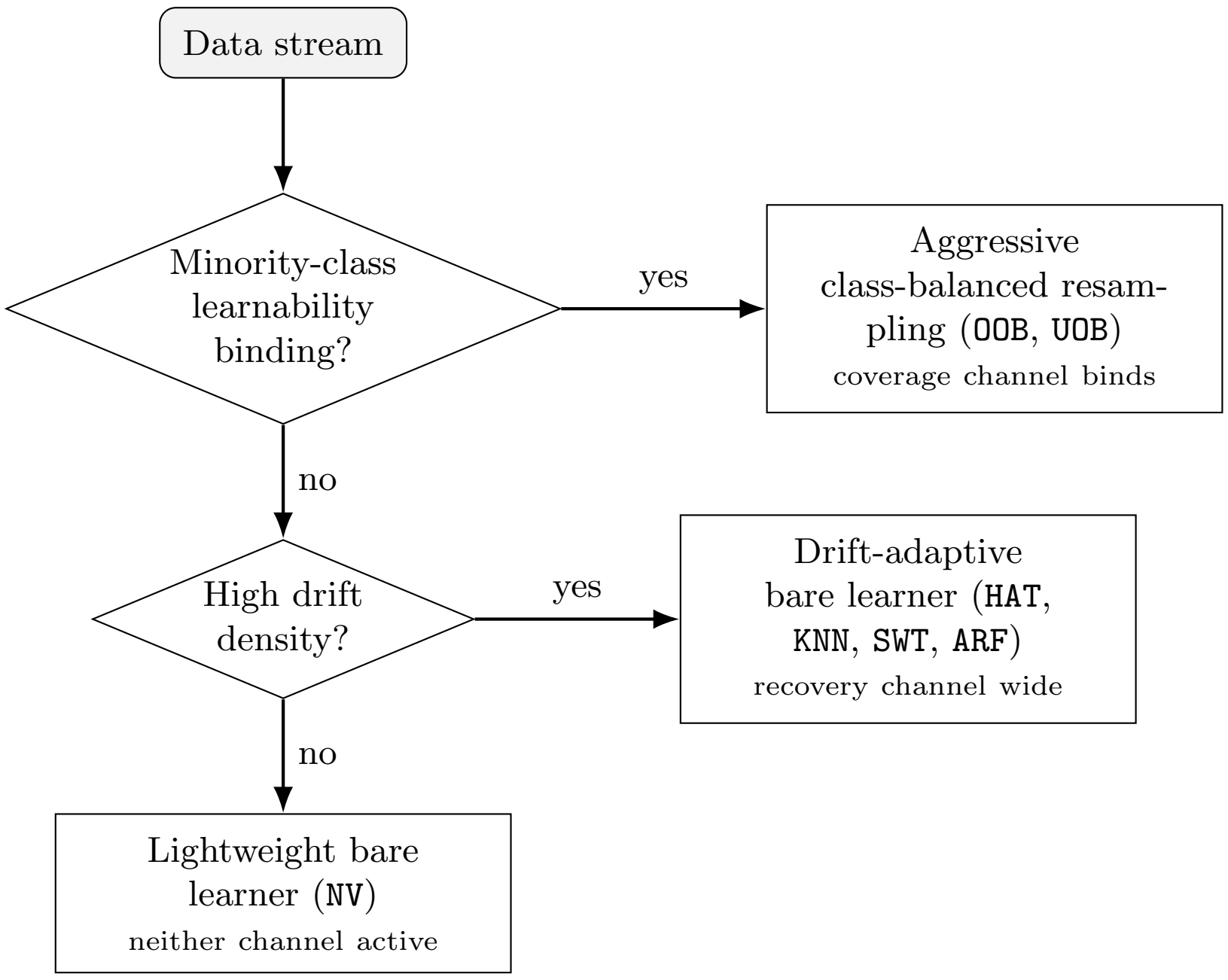

\paragraph{Findings about the methods studied}
The three-axis framework partitions the evaluated methods into operating regimes rather than ranking them on a single scale. Bare learners with intrinsic drift adaptation (\texttt{HAT}, \texttt{KNN}, \texttt{SWT}, \texttt{ARF}) own the high-drift regime: their per-label recovery channel is already wide enough to absorb most of the delay-induced signal, so synthesis-based oversamplers built on these bases (\texttt{VFCSMOTE\_HAT}, \texttt{VFCSMOTE\_KNN}, \texttt{VFCSMOTE\_ARF}, \texttt{VFCSMOTE\_SWT}) cannot widen the oversampler channel enough to enter the Pareto front under real-stream drift densities. Aggressive class-balanced resampling (\texttt{OOB}, \texttt{UOB}) occupies the regime where minority-class learnability is the dominant constraint, either through severe imbalance or through weak feature--minority separation at moderate imbalance, and can reach minority recall that standard methods cannot; this regime is independent of whether drift is present, which is why these methods appear on Pareto fronts spanning stationary and drift-active datasets alike. The drift-blind, slow-update base (\texttt{NV}) becomes Pareto-optimal precisely when neither the drift-adaptation channel nor the minority-coverage channel is active in the stream, the stationary, low-imbalance regime, where its lightweight statistical update suffices, and aggressive alternatives offer no compensating benefit. Synthesis-based oversamplers as a class do not own any real-data regime cleanly: the level effects that placed \texttt{CSMOTE} and \texttt{VFCSMOTE\_KNN} on synthetic Pareto fronts arise from controlled-stream conditions that the three real-data dimensions do not reproduce in combination, and the only synthesis variant retaining Pareto status on real data (\texttt{VFCSMOTE\_ARF}) does so by a margin smaller than the variation between deltas on its single qualifying dataset. The framework thus identifies oversampling as useful only where the bare learner's recovery channel is narrowed along at least one of the three axes; outside that condition, the practitioner should expect oversampling to be either redundant or actively harmful to the cost--quality trade. These regimes form a selection rule rather than a post-hoc account: a stream's position on the three axes determines which channel binds, and hence which method to deploy and whether oversampling is worth its cost. Figure~\ref{fig:decision} states the rule. The third axis, delay, does not appear as a branch: within the evaluated range, its effect on arrival performance is recovered by re-evaluation (RQ2) and refined performance is invariant to it (RQ4), so it conditions the operating regime rather than the method choice.

\paragraph{Practitioner recommendations}
Because stream characteristics are non-stationary and not fully observable under label delay, method selection should not be treated as a one-shot pre-deployment decision. Instead, we recommend periodically re-estimating the stream's minority-class learnability and drift density using recent labelled data, and revisiting Figure~\ref{fig:decision} to determine whether the indicated branch, and hence the deployed pipeline, should change. Replay overhead should be treated as a first-class deployment constraint rather than a secondary metric, since the synthetic-to-original ratios of Pareto-optimal oversamplers on real data span $0.25\times$ to $15.8\times$ and only the lower end of that range is plausibly deployable in latency-sensitive settings. Pareto status on a single dataset should not be generalised across streams without verifying that the minority-learnability and drift axes align, and, in particular, synthetic-data Pareto membership should not be used to select methods for real deployment.

\section{Conclusion}
\label{sec:conclusion}

This paper studied how oversampling and re-evaluation interact under label delay on imbalanced streams, contributing two findings of equal weight. The first is a decision framework: three axes---drift density, minority-class learnability, and delay relative to the refinement window---resolve, through a two-channel competition between the bare learner's per-label recovery channel and the oversampler's minority-coverage channel, which intervention to deploy on a given stream. The second is empirical: the cost--performance Pareto recommendations identified on controlled synthetic streams do not transfer to real data. \texttt{VFCSMOTE\_KNN}, the synthetic ``sweet spot'', is dominated on every real dataset, and synthesis-based oversamplers as a class are absent from real-data Pareto fronts. Real-data fronts instead partition by drift regime, placing bare learners or aggressive resampling at the front depending on which axis dominates the stream. Method selection by synthetic Pareto status is, accordingly, an unsafe default; selection should be conditioned on the stream's profile and revisited as that profile evolves.

The results are scoped to the evaluated settings ($K{=}100$, $\delta \leq 1.0\%$), within which re-evaluation recovers the delay-induced loss and refined performance is invariant to delay; behaviour beyond this boundary is left to follow-on work. The drift characterisation uses a single Hoeffding Tree and ADWIN probe on the $0/1$ loss stream, sufficient to establish the partition by drift profile but not to separate minority-specific drift from aggregate drift. The evaluation covers binary imbalanced classification on six real datasets spanning the drift profiles relevant to the framework; multi-class settings remain for future evaluation.

Three directions follow. First, characterise the $K$--$\delta$ boundary at which the refinement window stops absorbing delay, extending the framework beyond the settings tested here. Second, develop a lightweight proxy for minority-class learnability that guides selection without running the full evaluation pipeline. Third, investigate adaptive oversampling triggers that synthesise minority instances only when the bare learner's recovery channel is observably narrowing, reducing the high cost ratios observed for currently Pareto-optimal oversamplers.


\section*{Acknowledgements}
The authors have no acknowledgements to declare.

\section*{Declarations}

\paragraph{Ethical Approval} Not applicable\\

\paragraph{Funding} Not applicable\\

\paragraph{Availability of data and materials} All codes and used data can be found in this GitHub repo, \url{https://github.com/AhmedAwad/streaming-oversampling-under-delay-drift}.
.

\section*{Funding}
The authors received no specific funding for this work.

\section*{Declaration of Competing Interest}
The authors declare that they have no known competing financial interests 
or personal relationships that could have appeared to influence the work 
reported in this paper.












\end{document}